# On the Convergence Rate of Training Recurrent Neural Networks

Zeyuan Allen-Zhu
zeyuan@csail.mit.edu
Microsoft Research AI

Yuanzhi Li
yuanzhil@stanford.edu
Stanford University
Princeton University

Zhao Song
zhaos@utexas.edu
UT-Austin
University of Washington
Harvard University

October 28, 2018

(version 4)[*]

**Abstract**

How can local-search methods such as stochastic gradient descent (SGD) avoid bad local minima in training multi-layer neural networks? Why can they fit random labels even given non-convex and non-smooth architectures? Most existing theory only covers networks with one hidden layer, so can we go deeper?

In this paper, we focus on recurrent neural networks (RNNs) which are multi-layer networks widely used in natural language processing. They are harder to analyze than feedforward neural networks, because the *same* recurrent unit is repeatedly applied across the entire time horizon of length $L$, which is analogous to feedforward networks of depth $L$. We show when the number of neurons is sufficiently large, meaning polynomial in the training data size and in $L$, then SGD is capable of minimizing the regression loss in the linear convergence rate. This gives theoretical evidence of how RNNs can memorize data.

More importantly, in this paper we build general toolkits to analyze multi-layer networks with ReLU activations. For instance, we prove why ReLU activations can prevent exponential gradient explosion or vanishing, and build a perturbation theory to analyze first-order approximation of multi-layer networks.

---
[*]V1 appears on this date and no new result is added since then. V2/V3/V4 polish writing. Work was done when Yuanzhi Li and Zhao Song were 2018 summer interns at Microsoft Research, Redmond. When this work was performed, Yuanzhi Li was also affiliated with Princeton, and Zhao Song was also affiliated with UW and Harvard. We would like to specially thank Greg Yang, Sebastien Bubeck and Ofer Dekel for useful conversations.

# 1 Introduction

Neural networks have been one of the most powerful tools in machine learning over the past a few decades [4, 22, 29, 34, 39, 57, 58]. The multi-layer structure of neural network gives it supreme power in expressibility and learning performance. However, it raises complexity concerns: the training objective is generally *non-convex* and *non-smooth*. In practice, local-search algorithms such as stochastic gradient descent (SGD) are capable of finding global optima, at least on the training data [21, 69]. How SGD avoids local minima for such objectives remains an open theoretical question since Goodfellow et al. [21].

In recent years, there have been a number of theoretical results aiming at a better understanding of this phenomenon. Many of them focus on two-layer (thus one-hidden-layer) neural networks and assume that the inputs are random Gaussian or sufficiently close to Gaussian [8, 16, 20, 38, 44, 59, 64, 70, 71]. Some study deep neural networks but assuming the activation function is linear [6, 7, 25]. Some study the convex task of training essentially only the last layer of the network [14]. On the technique side, some of these results try to understand the gradient dynamics [8, 9, 14, 16, 38, 44, 59, 64, 65, 70, 71], while others focus on the geometry properties of the training objective [19, 20, 25, 50, 72].

More recently, Safran and Shamir [50] provided evidence that, even when inputs are standard Gaussians, two-layer neural networks can indeed have spurious local minima, and suggested that *over-parameterization* (i.e., increasing the number of neurons) may be the key in avoiding spurious local minima. Li and Liang [37] showed that, for two-layer networks with the cross-entropy loss, in the over-parametrization regime, gradient descent (GD) is capable of finding nearly-global optimal solutions on the training data. This result was later extended to the $\ell_2$ loss by Du et al. [17].

In this paper, we show GD and SGD are capable of training multi-layer neural networks (with ReLU activation) to global minima on any non-degenerate training data set. Furthermore, the running time is polynomial in the number of layers and the number of data points. Since there are many different types of multi-layer networks (convolutional, feedforward, recurrent, etc.), in this present paper, we focus on recurrent neural networks (RNN) as our choice of multi-layer networks, and feedforward networks are only its "special case" (see for instance a follow-up work [3]).

**Recurrent Neural Networks.** Among different architectures of neural networks, one of the *least* theoretically-understood structure is the recurrent one [18]. A recurrent neural network recurrently applies the same network unit to a sequence of input tokens, such as a sequence of words in a language sentence. RNN is particularly useful when there are long-term, non-linear interactions between input tokens in the same sequence. These networks are widely used in practice for natural language processing, language generation, machine translation, speech recognition, video and music processing, and many other tasks [12, 13, 32, 42, 43, 51, 61, 63]. On the theory side, while there are some attempts to show that an RNN is more expressive than a feedforward neural network [33], when and how an RNN can be efficiently learned has little theoretical explanation.

In practice, RNN is usually trained by simple local-search algorithms such as SGD. However, unlike shallow networks, the training process of RNN often runs into the trouble of vanishing or exploding gradient [62]. That is, the value of the gradient becomes exponentially small or large in the time horizon, even when the training objective is still constant. [1] In practice, one of the popular ways to resolve this is by the long short term memory (LSTM) structure [30]. However, one can also use rectified linear units (ReLUs) as activation functions to avoid vanishing or exploding gradient [52]. In fact, one of the earliest adoptions of ReLUs was on applications of RNNs for this purpose twenty years ago [24, 53]. For a detailed survey on RNN, we refer the readers to Salehinejad et al. [52].

---
[1]Intuitively, an RNN recurrently applies the same network unit for $L$ times if the input sequence is of length $L$. When this unit has "operator norm" larger than one or smaller than one, the final output can possibly exponentially explode or vanish in $L$. More importantly, when one back propagates through time—which intuitively corresponds

## 1.1 Our Question

In this paper, we study the following general question

- *Can ReLU provably stabilize the training process and avoid vanishing/exploding gradient?*
- *Can RNN be trained close to zero training error* efficiently *under mild assumptions?*

*Remark* 1.1. When there is no activation function, RNN is known as *linear dynamical system*. Hardt, Ma, and Recht [26] first proved the convergence of finding global minima for such linear dynamical systems. Followups in this line of research include [27, 28].

**Motivations.** One may also want to study whether RNN can be trained close to zero test error. However, unlike feedforward networks, the training error, or the ability to memorize examples, may actually be desirable for RNN. After all, many tasks involving RNN are related to memories, and certain RNN units are even referred to memory cells. Since RNN applies the same network unit to all input tokens in a sequence, the following question can possibly of its own interest:

- *How does RNN learn mappings (say from token 3 to token 7) without destroying others?*

Another motivation is the following. An RNN can be viewed as a space constraint, differentiable Turing machine, except that the input is only allowed to be read in a fixed order. It was shown in Siegelmann and Sontag [56] that all Turing machines can be simulated by fully-connected recurrent networks built of neurons with non-linear activations. In practice, RNN is also used as a tool to build neural Turing machines [23], equipped with a grand goal of automatically learning an algorithm based on the observation of the inputs and outputs. To this extent, we believe the task of understanding the *trainability* as a first step towards understanding RNN can be meaningful on its own.

**Our Result.** To present the simplest result, we focus on the classical Elman network with ReLU activation:

$$h_\ell = \phi(W \cdot h_{\ell-1} + A x_\ell) \in \mathbb{R}^m \qquad \text{where } W \in \mathbb{R}^{m \times m}, A \in \mathbb{R}^{m \times d_x}$$
$$y_\ell = B \cdot h_\ell \in \mathbb{R}^d \qquad \text{where } B \in \mathbb{R}^{d \times m}$$

We denote by $\phi$ the ReLU activation function: $\phi(x) = \max(x, 0)$. We note that (fully-connected) feedforward networks are only "special cases" to this by replacing $W$ with $W_\ell$ for each layer.[2]

We consider a regression task where each sequence of inputs consists of vectors $x_1, \ldots, x_L \in \mathbb{R}^{d_x}$ and we perform least-square regression with respect to $y_1^*, \ldots, y_L^* \in \mathbb{R}^d$. We assume there are $n$ training sequences, each of length $L$. We assume the training sequences are $\delta$-separable (say vectors $x_1$ are different by relative distance $\delta > 0$ for every pairs of training sequences). Our main theorem can be stated as follows

---
to applying the reverse unit multiple times— the gradient can also vanish or explode. Controlling the operator norm of a non-linear operator can be quite challenging.

[2]Most of the technical lemmas of this paper remain to hold (and become much simpler) once $W$ is replaced with $W_\ell$. This is carefully treated by [3].

**Theorem.** *If the number of neurons $m \geq \mathsf{poly}(n, d, L, \delta^{-1}, \log \varepsilon^{-1})$ is polynomially large, we can find weight matrices $W, A, B$ where the RNN gives $\varepsilon$ training error*

- *if gradient descent (GD) is applied for $T = \Omega\bigl(\frac{\mathsf{poly}(n,d,L)}{\delta^2} \log \frac{1}{\varepsilon}\bigr)$ iterations, starting from random Gaussian initializations; or*
- *if (mini-batch or regular) stochastic gradient descent (SGD) is applied for $T = \Omega\bigl(\frac{\mathsf{poly}(n,d,L)}{\delta^2} \log \frac{1}{\varepsilon}\bigr)$ iterations, starting from random Gaussian initializations.[3]*

(To present the simplest possible result, we have not tried to tighten the polynomial dependency with respect to $n, d$ and $L$. We only tightened the dependency with respect to $\delta$ and $\varepsilon$.)

**Our Contribution.** We summarize our contributions as follows.

- We believe this is the first proof of convergence of GD/SGD for training the *hidden layers* of recurrent neural networks (or even for any multi-layer networks of more than two layers) when activation functions are present.[4]
- Our results provide arguably the first theoretical evidence towards the empirical finding of Goodfellow et al. [21] on multi-layer networks, regarding the ability of SGD to avoid (spurious) local minima. Our theorem does not exclude the existence of bad local minima
- We build new technical toolkits to analyze multi-layer networks with ReLU activation, which have now found many applications [1–3, 10]. For instance, combining this paper with new techniques, one can derive guarantees on testing error for RNN in the PAC-learning language [1].

EXTENSION: DEEP RNN. Elman RNN is also referred to as three-layer RNN, and one may also study the convergence of RNNs with more hidden layers. This is referred to as deep RNN [52]. Our theorem also applies to deep RNNs (by combining this paper together with [3]).

EXTENSION: LOSS FUNCTIONS. For simplicity, in this paper we have adopted the $\ell_2$ regression loss. Our results generalize to other Lipschitz smooth (but possibly nonconvex) loss functions, by combining with the techniques of [3].

## 1.2 Other Related Works

Another relevant work is Brutzkus et al. [9] where the authors studied over-paramterization in the case of two-layer neural network under a linear-separable assumption.

Instead of using randomly initialized weights like this paper, there is a line of work proposing algorithms using weights generated from some "tensor initialization" process [5, 31, 55, 64, 71].

There is huge literature on using the mean-field theory to study neural networks [11, 15, 36, 41, 45–47, 49, 54, 66–68]. At a high level, they study the network dynamics at random initialization when the number of hidden neurons grow to infinity, and use such initialization theory to predict performance after training. However, they do not provide theoretical convergence rate for the training process (at least when the number of neurons is finite).

---

[3]At a first glance, one may question how it is possible for SGD to enjoy a logarithmic time dependency in $\varepsilon^{-1}$; after all, even when minimizing strongly-convex and Lipschitz-smooth functions, the typical convergence rate of SGD is $T \propto 1/\varepsilon$ as opposed to $T \propto \log(1/\varepsilon)$. We quickly point out there is no contradiction here if the stochastic pieces of the objective enjoy a *common* global minimizer. In math terms, suppose we want to minimize some function $f(x) = \frac{1}{n}\sum_{i=1}^{n} f_i(x)$, and suppose $x^*$ is the global minimizer of convex functions $f_1(x), \ldots, f_n(x)$. Then, if $f(x)$ is $\sigma$-strongly convex, and each each $f_i(x)$ is $L$-Lipschitz smooth, then SGD—moving in negative direction of $\nabla f_i(x)$ for a random $i \in [n]$ per step— can find $\varepsilon$-minimizer of this function in $O\bigl(\frac{L^2}{\sigma^2} \log \frac{1}{\varepsilon}\bigr)$ iterations.

[4]Our theorem holds even when $A, B$ are at random initialization and only the hidden weight matrix $W$ is trained. This is much more difficult to analyze than the convex task of training only the last layer $B$ [14]. Training only the last layer can significantly reduce the learning power of (recurrent or not) neural networks in practice.

# 2 Notations and Preliminaries

We denote by $\|\cdot\|_2$ (or sometimes $\|\cdot\|$) the Euclidean norm of vectors, and by $\|\cdot\|_2$ the spectral norm of matrices. We denote by $\|\cdot\|_\infty$ the infinite norm of vectors, $\|\cdot\|_0$ the sparsity of vectors or diagonal matrices, and $\|\cdot\|_F$ the Frobenius norm of matrices. Given matrix $W$, we denote by $W_k$ or $w_k$ the $k$-th row vector of $W$. We denote the row $\ell_p$ norm for $W \in \mathbb{R}^{m \times d}$ as

$$\|W\|_{2,p} \stackrel{\text{def}}{=} \left(\sum_{i \in [m]} \|w_i\|_2^p\right)^{1/p}. \tag{2.1}$$

By definition, $\|W\|_{2,2} = \|W\|_F$.

We use $\mathcal{N}(\mu, \sigma)$ to denote Gaussian distribution with mean $\mu$ and variance $\sigma$; or $\mathcal{N}(\mu, \Sigma)$ to denote Gaussian vector with mean $\mu$ and covariance $\Sigma$. We use $\mathbb{1}_{event}$ to denote the indicator function of whether *event* is true. We denote by $\mathbf{e}_k$ the $k$-th standard basis vector. We use $\phi(\cdot)$ to denote the ReLU function, namely $\phi(x) = \max\{x, 0\} = \mathbb{1}_{x \geq 0} \cdot x$. Given univariate function $f \colon \mathbb{R} \to \mathbb{R}$, we also use $f$ to denote the same function over vectors: $f(x) = (f(x_1), \ldots, f(x_m))$ if $x \in \mathbb{R}^m$.

Given vectors $v_1, \ldots, v_n \in \mathbb{R}^m$, we define $U = \mathsf{GS}(v_1, \ldots, v_n)$ as their Gram-Schmidt orthonormalization. Namely, $U = [\widehat{v}_1, \ldots, \widehat{v}_n] \in \mathbb{R}^{m \times n}$ where

$$\widehat{v}_1 = \frac{v_1}{\|v_1\|} \quad \text{and} \quad \text{for } i \geq 2: \quad \widehat{v}_i = \frac{\prod_{j=1}^{i-1}(I - \widehat{v}_j \widehat{v}_j^\top) v_i}{\|\prod_{j=1}^{i-1}(I - \widehat{v}_j \widehat{v}_j^\top) v_i\|}.$$

Note that in the occasion that $\prod_{j=1}^{i-1}(I - \widehat{v}_j \widehat{v}_j^\top) v_i$ is the zero vector, we let $\widehat{v}_i$ be an arbitrary unit vector that is orthogonal to $\widehat{v}_1, \ldots, \widehat{v}_{i-1}$.

## 2.1 Elman Recurrent Neural Network

We assume $n$ training inputs are given: $(x_{i,1}, x_{i,2}, \ldots, x_{i,L}) \in \left(\mathbb{R}^{d_x}\right)^L$ for each input $i \in [n]$. We assume $n$ training labels are given: $(y_{i,1}^*, y_{i,2}^*, \ldots, y_{i,L}^*) \in \left(\mathbb{R}^d\right)^L$ for each input $i \in [n]$. Without loss of generality, we assume $\|x_{i,\ell}\| \leq 1$ for every $i \in [n]$ and $\ell \in [L]$. Also without loss of generality, we assume $\|x_{i,1}\| = 1$ and its last coordinate $[x_{i,1}]_{d_x} = \frac{1}{\sqrt{2}}$ for every $i \in [n]$.[5]

We make the following assumption on the input data (see Footnote 10 for how to relax it):

**Assumption 2.1.** $\|x_{i,1} - x_{j,1}\| \geq \delta$ *for some parameter $\delta \in (0, 1]$ and every pair of $i \neq j \in [n]$.*

Given weight matrices $W \in \mathbb{R}^{m \times m}$, $A \in \mathbb{R}^{m \times d_x}$, $B \in \mathbb{R}^{d \times m}$, we introduce the following notations to describe the evaluation of RNN on the input sequences. For each $i \in [n]$ and $j \in [L]$:

$$h_{i,0} = 0 \in \mathbb{R}^m \qquad\qquad g_{i,\ell} = W \cdot h_{i,\ell-1} + A x_{i,\ell} \in \mathbb{R}^m$$
$$y_{i,\ell} = B \cdot h_{i,\ell} \in \mathbb{R}^d \qquad\qquad h_{i,\ell} = \phi(W \cdot h_{i,\ell-1} + A x_{i,\ell}) \in \mathbb{R}^m$$

A very important notion that this entire paper relies on is the following:

**Definition 2.2.** *For each $i \in [n]$ and $\ell \in [L]$, let $D_{i,\ell} \in \mathbb{R}^{m \times m}$ be the diagonal matrix where*

$$(D_{i,\ell})_{k,k} = \mathbf{1}_{(W \cdot h_{i,\ell-1} + A x_{i,\ell})_k \geq 0} = \mathbf{1}_{(g_{i,\ell})_k \geq 0} \ .$$

*As a result, we can write $h_{i,\ell} = D_{i,\ell} W h_{i,\ell-1}$.*

We consider the following random initialization distributions for $W$, $A$ and $B$.

---

[5] If it only satisfies $\|x_{i,1}\| \leq 1$ one can pad it with an additional coordinate to make $\|x_{i,1}\| = 1$ hold. As for the assumption $[x_{i,1}]_{d_x} = \frac{1}{\sqrt{2}}$, this is equivalent to adding a bias term $\mathcal{N}(0, \frac{1}{m})$ for the first layer.

**Definition 2.3.** *We say that $W, A, B$ are at random initialization, if the entries of $W$ and $A$ are i.i.d. generated from $\mathcal{N}(0, \frac{2}{m})$, and the entries of $B_{i,j}$ are i.i.d. generated from $\mathcal{N}(0, \frac{1}{d})$.*

Throughout this paper, for notational simplicity, we refer to index $\ell$ as the $\ell$-th *layer* of RNN, and $h_{i,\ell}, x_{i,\ell}, y_{i,\ell}$ respectively as the hidden neurons, input, output on the $\ell$-th layer. We acknowledge that in certain literatures, one may regard Elman network as a three-layer RNN.

**Assumption 2.4.** *We assume $m \geq \mathsf{poly}(n, d, L, \frac{1}{\delta}, \log \frac{1}{\varepsilon})$ for some sufficiently large polynomial.*

Without loss of generality, we assume $\delta \leq \frac{1}{CL^2 \log^3 m}$ for some sufficiently large constant $C$ (if this is not satisfied one can decrease $\delta$). Throughout the paper except the detailed appendix, we use $\widetilde{O}, \widetilde{\Omega}$ and $\widetilde{\Theta}$ notions to hide polylogarithmic dependency in $m$. To simplify notations, we denote by
$$\rho \stackrel{\text{def}}{=} nLd\log m \quad \text{and} \quad \varrho \stackrel{\text{def}}{=} nLd\delta^{-1}\log(m/\varepsilon) \ .$$

### 2.2 Objective and Gradient

For simplicity, we only optimize over the weight matrix $W \in \mathbb{R}^{m \times m}$ and let $A$ and $B$ be at random initialization. As a result, our $\ell_2$-regression objective is a function over $W$:[6]

$$f(W) \stackrel{\text{def}}{=} \sum_{i=1}^n f_i(W) \quad \text{and} \quad f_i(W) \stackrel{\text{def}}{=} \frac{1}{2} \sum_{\ell=2}^L \|\mathsf{loss}_{i,\ell}\|_2^2 \quad \text{where} \quad \mathsf{loss}_{i,\ell} \stackrel{\text{def}}{=} Bh_{i,\ell} - y_{i,\ell}^* \ .$$

Using chain rule, one can write down a closed form of the (sub-)gradient:

**Fact 2.5.** *For $k \in [m]$, the gradient with respect to $W_k$ (denoted by $\nabla_k$) and the full gradient are*

$$\nabla_k f(W) = \sum_{i=1}^n \sum_{a=2}^L \sum_{\ell=1}^{a-1} (\mathsf{Back}_{i,\ell+1\to a}^\top \cdot \mathsf{loss}_{i,a})_k \cdot h_{i,\ell} \cdot \mathbf{1}_{\langle W_k, h_{i,\ell}\rangle + \langle A_k, x_{i,\ell+1}\rangle \geq 0}$$

$$\nabla f(W) = \sum_{i=1}^n \sum_{a=2}^L \sum_{\ell=1}^{a-1} D_{i,\ell+1}\big(\mathsf{Back}_{i,\ell+1\to a}^\top \cdot \mathsf{loss}_{i,a}\big) \cdot h_{i,\ell}^\top$$

*where for every $i \in [n]$, $\ell \in [L]$, and $a = \ell + 1, \ell + 2, \ldots, L$:*

$$\mathsf{Back}_{i,\ell\to\ell} \stackrel{\text{def}}{=} B \in \mathbb{R}^{d\times m} \quad \text{and} \quad \mathsf{Back}_{i,\ell\to a} \stackrel{\text{def}}{=} BD_{i,a}W \cdots D_{i,\ell+1}W \in \mathbb{R}^{d\times m} \ .$$

## 3 Our Results

Our main results can be formally stated as follows.

**Theorem 1** (GD). *Suppose $\eta = \widetilde{\Theta}\big(\frac{\delta}{m}\mathsf{poly}(n,d,L)\big)$ and $m \geq \mathsf{poly}(n,d,L,\delta^{-1},\log \varepsilon^{-1})$. Let $W^{(0)}, A, B$ be at random initialization. With high probability over the randomness of $W^{(0)}, A, B$, if we apply gradient descent for $T$ steps $W^{(t+1)} = W^{(t)} - \eta \nabla f(W^{(t)})$, then it satisfies*

$$f(W^{(T)}) \leq \varepsilon \quad \text{for} \quad T = \widetilde{\Omega}\Big(\frac{\mathsf{poly}(n,d,L)}{\delta^2}\log\frac{1}{\varepsilon}\Big).$$

**Theorem 2** (SGD). *Suppose $\eta = \widetilde{\Theta}\big(\frac{\delta}{m}\mathsf{poly}(n,d,L)\big)$ and $m \geq \mathsf{poly}(n,d,L,\delta^{-1},\log \varepsilon^{-1})$. Let $W^{(0)}, A, B$ be at random initialization. If we apply stochastic gradient descent for $T$ steps $W^{(t+1)} =$* $W^{(t)} - \eta \nabla f_i(W^{(t)})$ *for a random index* $i \in [n]$ *per step, then with high probability (over* $W^{(0)}, A, B$ *and the randomness of SGD), it satisfies*

$$f(W^{(T)}) \leq \varepsilon \quad for \quad T = \widetilde{\Omega}\Big(\frac{\mathsf{poly}(n,d,L)}{\delta^2} \log \frac{1}{\varepsilon}\Big).$$

---

[6] The index $\ell$ starts from 2, because $Bh_{i,1} = B\phi(Ax_{i,1})$ remains constant if we are not optimizing over $A$ and $B$.

In both cases, we essentially have *linear convergence rates*.[7] Notably, our results show that the dependency of the number of layers $L$, is *polynomial*. Thus, even when RNN is applied to sequences of long input data, it does not suffer from exponential gradient explosion or vanishing (e.g., $2^{\Omega(L)}$ or $2^{-\Omega(L)}$) through the entire training process.

**Main Technical Theorems.** Our main Theorem 1 and Theorem 2 are in fact natural consequences of the following two technical theorems. They both talk about the first-order behavior of RNNs when the weight matrix $W$ is sufficiently close to some random initialization.

The first theorem is similar to the classical Polyak-Łojasiewicz condition [40, 48], and says that $\|\nabla f(W)\|_F^2$ is at least as large as the objective value.

**Theorem 3.** *With high probability over random initialization* $\widetilde{W}, A, B$, *it satisfies*

$$\forall W \in \mathbb{R}^{m \times m} \text{ with } \|W - \widetilde{W}\|_2 \leq \frac{\mathsf{poly}(\varrho)}{\sqrt{m}}: \qquad \|\nabla f(W)\|_F^2 \geq \frac{\delta}{\mathsf{poly}(\rho)} \times m \times f(W) \ ,$$

$$\|\nabla f(W)\|_F^2, \|\nabla f_i(W)\|_F^2 \leq \mathsf{poly}(\rho) \times m \times f(W) \ .$$

The second theorem shows a special "semi-smoothness" property of the objective.

**Theorem 4.** *With high probability over random initialization* $\widetilde{W}, A, B$, *it satisfies for every* $\breve{W} \in \mathbb{R}^{m \times m}$ *with* $\|\breve{W} - \widetilde{W}\| \leq \frac{\mathsf{poly}(\varrho)}{\sqrt{m}}$, *and for every* $W' \in \mathbb{R}^{m \times m}$ *with* $\|W'\| \leq \frac{\tau_0}{\sqrt{m}}$,

$$f(\breve{W} + W') \leq f(\breve{W}) + \langle \nabla f(\breve{W}), W' \rangle + \mathsf{poly}(\varrho) m^{1/3} \cdot \sqrt{f(\breve{W})} \cdot \|W'\|_2 + \mathsf{poly}(\rho) m \|W'\|_2^2 \ .$$

At a high level, the convergence of GD and SGD are careful applications of the two technical theorems above: indeed, Theorem 3 shows that as long as the objective value is high, the gradient is large; and Theorem 4 shows that if one moves in the (negative) gradient direction, then the objective value can be sufficiently decreased. These two technical theorems together ensure that GD/SGD does not hit any saddle point or (bad) local minima along its training trajectory. This was practically observed by Goodfellow et al. [21] and a theoretical justification was open since then.

**An Open Question.** We did not try to tighten the polynomial dependencies of $(n, d, L)$ in the proofs. When $m$ is sufficiently large, we make use of the randomness at initialization to argue that, *for all the points* within a certain radius from initialization, for instance Theorem 3 holds. In practice, however, the SGD can create additional randomness as time goes; also, in practice, it suffices for those points on the SGD trajectory to satisfy Theorem 3. Unfortunately, such randomness can— in principle— be correlated with the SGD trajectory, so we do not know how to use that in the proofs. Analyzing such correlated randomness is certainly beyond the scope of this paper, but can possibly explain why in practice, the size of $m$ needed is not that large.

### 3.1 Conclusion

Overall, we provide the first proof of convergence of GD/SGD for non-linear neural networks that have more two layers. We show with overparameterization GD/SGD can avoid hitting any (bad) local minima along its training trajectory. This was practically observed by Goodfellow et al. [21] and a theoretical justification was open since then. We present our result using recurrent neural networks (as opposed to the simpler feedforward networks [3]) in this very first paper, because memorization in RNN could be of independent interest. Also, our result proves that RNN can learn mappings from different input tokens to different output tokens *simultaneously* using the same recurrent unit.

---

[7]We remark here that the $\widetilde{O}$ notation may hide additional polynomial dependency in $\log \log \varepsilon^{-1}$. This is not necessary, at the expense of slightly complicating the proofs, as shown by follow up [3].

Last but not least, we build new tools to analyze multi-layer networks with ReLU activations that could facilitate many new research on deep learning. For instance, our techniques in Section 4 provide a general theory for why ReLU activations avoid exponential exploding (see e.g. (4.1), (4.4)) or exponential vanishing (see e.g. (4.1), (4.3)); and our techniques in Section 5 give a general theory for the stability of multi-layer networks against adversarial weight perturbations, which is at the heart of showing the semi-smoothness Theorem 4, and used by all the follow-up works [1–3, 10].

# PROOF SKETCH

The main difficulty of this paper is to prove Theorem 3 and 4, and we shall sketch the proof ideas in Section 4 through 7. In such high-level discussions, we shall put our emphasize on

- how to avoid exponential blow up in $L$, and
- how to deal with the issue of randomness dependence across layers.

We genuinely hope that this high-level sketch can (1) give readers a clear overview of the proof without the necessity of going to the appendix, and (2) appreciate our proof and understand why it is necessarily long.[8]

## 4 Basic Properties at Random Initialization

In this section we derive basic properties of the RNN when the weight matrices $W, A, B$ are all *at random initialization*. The corresponding precise statements and proofs are in Appendix B.

The first one says that the forward propagation neither explodes or vanishes, that is,

$$\frac{1}{2} \leq \|h_{i,\ell}\|_2, \|g_{i,\ell}\|_2 \leq O(L) \ . \tag{4.1}$$

Intuitively, (4.1) very reasonable. Since the weight matrix $W$ is randomly initialized with entries i.i.d. from $\mathcal{N}\left(0, \frac{2}{m}\right)$, the norm $\|Wz\|_2$ is around $\sqrt{2}$ for any fixed vector $z$. Equipped with ReLU activation, it "shuts down" roughly half of the coordinates of $Wz$ and reduces the norm $\|\phi(Wz)\|$ to one. Since in each layer $\ell$, there is an additional unit-norm signal $x_{i,\ell}$ coming in, we should expect the final norm of hidden neurons to be at most $O(L)$.

Unfortunately, the above argument cannot be directly applied since the weight matrix $W$ is reused for $L$ times so there is no fresh new randomness across layers. Let us explain how we deal with this issue carefully, because it is at the heart of *all* of our proofs in this paper. Recall, each time $W$ is applied to some vector $h_{i,\ell}$, it only uses "one column of randomness" of $W$. Mathematically, letting $U_\ell \in \mathbb{R}^{m \times n\ell}$ denote the column orthonormal matrix using Gram-Schmidt

$$U_\ell \stackrel{\text{def}}{=} \mathsf{GS}\,(h_{1,1}, \ldots, h_{n,1}, h_{1,2}, \ldots, h_{n,2}, \ldots, h_{1,\ell}, \ldots, h_{n,\ell})\ ,$$

we have $W h_{i,\ell} = W U_{\ell-1} U_{\ell-1}^\top h_{i,\ell} + W(I - U_{\ell-1} U_{\ell-1}^\top) h_{i,\ell}$.

---

[8]For instance, proving gradient norm lower bound in Theorem 3 for a single neuron $k \in [m]$ is easy, but how to apply concentration across neurons? Crucially, due to the recurrent structure these quantities are never independent, so we have to build necessary probabilistic tools to tackle this. If one is willing to ignore such subtleties, then our sketched proof is sufficiently short and gives a good overview.

- The second term $W(I - U_{\ell-1} U_{\ell-1}^\top) h_{i,\ell}$ has new randomness independent of the previous layers.[9]

- The first term $W U_{\ell-1} U_{\ell-1}^\top h_{i,\ell}$ relies on the randomness of $W$ in the directions of $h_{i,a}$ for $a < \ell$ of the previous layers. We cannot rely on the randomness of this term, because when applying inductive argument till layer $\ell$, the randomness of $W U_{\ell-1}$ is already used.

  Fortunately, $W U_{\ell-1} \in \mathbb{R}^{m \times n(\ell-1)}$ is a rectangular matrix with $m \gg n(\ell-1)$ (thanks to overparameterization!) so one can bound its spectral norm by roughly $\sqrt{2}$. This ensures that no matter how $h_{i,\ell}$ behaves (even arbitrarily correlated with $W U_{\ell-1}$), the norm of the first term cannot be too large. It is crucial here that $W U_{\ell-1}$ is a *rectangular* matrix, because for a square random matrix such as $W$, its spectral norm is 2 and using that, the forward propagation bound will exponentially blow up.

This summarizes the main idea for proving $\|h_{i,\ell}\| \leq O(L)$ in (4.1); the lower bound $\frac{1}{2}$ is similar. Our next property says in each layer, the amount of "fresh new randomness" is non-negligible:

$$\|(I - U_{\ell-1} U_{\ell-1}^\top) h_{i,\ell}\|_2 \geq \widetilde{\Omega}(\frac{1}{L^2})\ . \tag{4.2}$$

This relies on a more involved inductive argument than (4.1). At high level, one needs to show that in each layer, the amount of "fresh new randomness" reduces only by a factor at most $1 - \frac{1}{10L}$.

Using (4.1) and (4.2), we obtain the following property about the data separability:

$$(I - U_\ell U_\ell^\top) h_{i,\ell+1} \text{ and } (I - U_\ell U_\ell^\top) h_{j,\ell+1} \text{ are } (\delta/2)\text{-separable}, \forall i,j \in [n] \text{ with } i \neq j \tag{4.3}$$

Here, we say two vectors $x$ and $y$ are $\delta$-separable if $\|(I - yy^\top/\|y\|_2^2)x\| \geq \delta$ and vice versa. Property (4.3) shows that the separability information (say on input token 1) *does not diminish* by more than a polynomial factor even if the information is propagated for $L$ layers.

We prove (4.3) by induction. In the first layer $\ell = 1$ we have $h_{i,1}$ and $h_{j,1}$ are $\delta$-separable which is a consequence of Assumption 2.1. If having fresh new randomness, given two $\delta$ separable vectors $x, y$, one can show that $\phi(Wx)$ and $\phi(Wy)$ are also $\delta(1 - o(\frac{1}{L}))$-separable. Again, in RNN, we do not have fresh new randomness, so we rely on (4.2) to give us reasonably large fresh new randomness. Applying a careful induction helps us to derive that (4.3) holds for all layers.[10]

**Intermediate Layers and Backward Propagation.** Training neural network is not only about forward propagation. We also have to bound intermediate layers and backward propagation.

The first two results we derive are the following. For every $\ell_1 \geq \ell_2$ and diagonal matrices $D'$ of sparsity $s \in [\rho^2, m^{0.49}]$:

$$\|W D_{i,\ell_1} \cdots D_{i,\ell_2} W\|_2 \leq O(L^3) \tag{4.4}$$

$$\|D' W D_{i,\ell_1} \cdots D_{i,\ell_2} W D'\|_2 \leq \widetilde{O}(\sqrt{s}/\sqrt{m}) \tag{4.5}$$

---

[9]More precisely, letting $v = (I - U_{\ell-1} U_{\ell-1}^\top) h_{i,\ell}$, we have $W(I - U_{\ell-1} U_{\ell-1}^\top) h_{i,\ell} = \left(W \frac{v}{\|v\|}\right) \|v\|$. Here, $W \frac{v}{\|v\|}$ is a random Gaussian vector in $\mathcal{N}(0, \frac{2}{m} I)$ and is *independent* of all $\{h_{i,a} \mid i \in [n], a < \ell\}$.

[10]This is the only place that we rely on Assumption 2.1. This assumption is somewhat necessary in the following sense. If $x_{i,\ell} = x_{j,\ell}$ for some pair $i \neq j$ for all the first ten layers $\ell = 1, 2, \ldots, 10$, and if $y_{i,\ell}^* \neq y_{j,\ell}^*$ for even just one of these layers, then there is no hope in having the training objective decrease to zero. Of course, one can make more relaxed assumption on the input data, involving both $x_{i,\ell}$ and $y_{i,\ell}^*$. While this is possible, it complicates the statements so we do not present such results in this paper.

Intuitively, one cannot use spectral bound argument to derive (4.4) or (4.5): the spectral norm of $W$ is 2, and even if ReLU activations cancel half of its mass, the spectral norm $\|DW\|_2$ remains to be $\sqrt{2}$. When stacked together, this grows exponential in $L$.

Instead, we use an analogous argument to (4.1) to show that, for each fixed vector $z$, the norm of $\|WD_{i,\ell_1} \cdots D_{i,\ell_2} Wz\|_2$ is at most $O(1)$ with extremely high probability $1 - e^{-\Omega(m/L^2)}$. By standard $\varepsilon$-net argument, $\|WD_{i,\ell_1} \cdots D_{i,\ell_2} Wz\|_2$ is at most $O(1)$ for all $\frac{m}{L^3}$-sparse vectors $z$. Finally, for a possible dense vector $z$, we can divide it into $L^3$ chunks each of sparsity $\frac{m}{L^3}$. Finally, we apply the upper bound for $L^3$ times. This proves (4.4). One can use similar argument to prove (4.5).

*Remark* 4.1. We did not try to tighten the polynomial factor here in $L$. We conjecture that proving an $O(1)$ bound may be possible, but that question itself may be a sufficiently interesting random matrix theory problem on its own.

The next result is for back propagation. For every $\ell_1 \geq \ell_2$ and diagonal matrices $D'$ of sparsity $s \in [\rho^2, m^{0.49}]$:

$$\left\| BD_{i,\ell_1} \cdots D_{i,\ell_2} WD' \right\|_2 \leq \widetilde{O}(\sqrt{s}) \tag{4.6}$$

Its proof is in the same spirit as (4.5), with the only difference being the spectral norm of $B$ is around $\sqrt{m/d}$ as opposed to $O(1)$.

## 5 Stability After Adversarial Perturbation

In this section we study the behavior of RNN after adversarial perturbation. The corresponding precise statements and proofs are in Appendix C.

Letting $\widetilde{W}, A, B$ be at random initialization, we consider some matrix $W = \widetilde{W} + W'$ for $\|W'\|_2 \leq \frac{\mathsf{poly}(\varrho)}{\sqrt{m}}$. Here, $W'$ may depend on the randomness of $\widetilde{W}, A$ and $B$, so we say it can be *adversarially* chosen. The results of this section will later be applied essentially twice:

- Once for those updates generated by GD or SGD, where $W'$ is how much the algorithm has moved away from the random initialization.

- The other time (see Section 6.3) for a technique that we call "randomness decomposition" where we decompose the true random initialization $W$ into $W = \widetilde{W} + W'$, where $\widetilde{W}$ is a "fake" random initialization but identically distributed as $W$. Such technique at least traces back to smooth analysis [60].

To illustrate our high-level idea, from this section on (so in Section 5, 6 and 7)

> we ignore the polynomial dependency in $\varrho$ and *hide it in the big-O notion.*

We denote by $\widetilde{D}_{i,\ell}, \widetilde{g}_{i,\ell}, \widetilde{h}_{i,\ell}$ respectively the values of $D_{i,\ell}, g_{i,\ell}$ and $h_{i,\ell}$ determined by $\widetilde{W}$ and $A$ at random initialization; and by $D_{i,\ell} = \widetilde{D}_{i,\ell} + D'_{i,\ell}$, $g_{i,\ell} = \widetilde{g}_{i,\ell} + g'_{i,\ell}$ and $h_{i,\ell} = \widetilde{h}_{i,\ell} + h'_{i,\ell}$ respectively those determined by $W = \widetilde{W} + W'$ after the adversarial perturbation.

**Forward Stability.** Our first, and most technical result is the following:

$$\|g'_{i,\ell}\|_2, \|h'_{i,\ell}\|_2 \leq O(m^{-1/2}) \ , \quad \|D'_{i,\ell}\|_0 \leq O(m^{2/3}) \quad \text{and} \quad \|D'_{i,\ell} g_{i,\ell}\|_2 \leq O(m^{-1/2}) \ . \tag{5.1}$$

Intuitively, one may hope to prove (5.1) by induction, because we have (ignoring subscripts in $i$)

$$g'_{\ell'} = \underbrace{W' D_{\ell'-1} g_{\ell'-1}}_{①} + \underbrace{\widetilde{W} D'_{\ell'-1} g_{\ell'-1}}_{②} + \underbrace{\widetilde{W} \widetilde{D}_{\ell'-1} g'_{\ell'-1}}_{③} \ .$$

The main issue here is that, the spectral norm of $\widetilde{W}\widetilde{D}_{\ell'-1}$ in ③ is greater than 1, so we cannot apply naive induction due to exponential blow up in $L$. Neither can we apply techniques from Section 4, because the changes such as $g_{\ell'-1}$ can be *adversarial*.

In our actual proof of (5.1), instead of applying induction on ③, we recursively expand ③ by the above formula. This results in a total of $L$ terms of ① type and $L$ terms of ② type. The main difficulty is to bound a term of ② type, that is:
$$\big\|\widetilde{W}\widetilde{D}_{\ell_1}\cdots\widetilde{D}_{\ell_2+1}\widetilde{W}D'_{\ell_2}g_{\ell_2}\big\|_2$$

Our argument consists of two conceptual steps.

(1) Suppose $g_{\ell_2} = \widetilde{g}_{\ell_2} + g'_{\ell_2} = \widetilde{g}_{\ell_2} + g'_{\ell_2,1} + g'_{\ell_2,2}$ where $\|g'_{\ell_2,1}\|_2 \leq m^{-1/2}$ and $\|g'_{\ell_2,2}\|_\infty \leq m^{-1}$, then we argue that $\|D'_{\ell_2}g_{\ell_2}\|_2 \leq O(m^{-1/2})$ and $\|D'_{\ell_2}g_{\ell_2}\|_0 \leq O(m^{2/3})$.

(2) Suppose $x \in \mathbb{R}^m$ with $\|x\|_2 \leq m^{-1/2}$ and $\|x\|_0 \leq m^{2/3}$, then we show that $y = \widetilde{W}\widetilde{D}_{\ell_1}\cdots\widetilde{D}_{\ell_2+1}\widetilde{W}x$ can be written as $y = y_1 + y_2$ with $\|y_1\|_2 \leq O(m^{-2/3})$ and $\|y_2\|_\infty \leq O(m^{-1})$.

The two steps above enable us to perform induction without exponential blow up. Indeed, they together enable us to go through the following logic chain:

$$\left.\begin{array}{l}\|\cdot\|_2 \leq m^{-1/2} \text{ and } \|\cdot\|_\infty \leq m^{-1} \\ \|\cdot\|_2 \leq m^{-2/3} \text{ and } \|\cdot\|_\infty \leq m^{-1}\end{array}\right\} \begin{array}{c}(1) \\ \Longrightarrow \\ \Longleftarrow \\ (2)\end{array} \|\cdot\|_2 \leq m^{-1/2} \text{ and } \|\cdot\|_0 \leq m^{2/3}$$

Since there is a gap between $m^{-1/2}$ and $m^{-2/3}$, we can make sure that all blow-up factors are absorbed into this gap, using the property that $m$ is polynomially large. This enables us to perform induction to prove (5.1) without exponential blow-up.

**Intermediate Layers and Backward Stability.** Using (5.1), and especially using the sparsity $\|D'\|_0 \leq m^{2/3}$ from (5.1), one can apply the results in Section 4 to derive the following stability bounds for intermediate layers and backward propagation:

$$\big\|D_{i,\ell_1}W\cdots D_{i,\ell_2}W - \widetilde{D}_{i,\ell_1}\widetilde{W}\cdots \widetilde{D}_{i,\ell_2}\widetilde{W}\big\|_2 \leq O(L^7) \tag{5.2}$$

$$\big\|BD_{i,\ell_1}W\cdots D_{i,\ell_2}W - B\widetilde{D}_{i,\ell_1}\widetilde{W}\cdots \widetilde{D}_{i,\ell_2}\widetilde{W}\big\|_2 \leq O(m^{1/3}) \ . \tag{5.3}$$

**Special Rank-1 Perturbation.** For technical reasons, we also need two bounds in the special case of $W' = yz^\top$ for some unit vector $z$ and sparse $y$ with $\|y\|_0 \leq \mathsf{poly}(\varrho)$. We prove that, for this type of rank-one adversarial perturbation, it satisfies for every $k \in [m]$:

$$|((\widetilde{W}+W')h'_{i,\ell})_k| \leq O(m^{-2/3}) \tag{5.4}$$

$$\big\|BD_{i,\ell_1}W\cdots D_{i,\ell_2}W\mathbf{e}_k - B\widetilde{D}_{i,\ell_1}\widetilde{W}\cdots \widetilde{D}_{i,\ell_2}\widetilde{W}\mathbf{e}_k\big\|_2 \leq O(m^{-1/6}) \tag{5.5}$$

# 6  Proof Sketch of Theorem 3: Polyak-Łojasiewicz Condition

The upper bound in Theorem 3 is easy to prove (based on Section 4 and 5), but the lower bound (a.k.a. the Polyak-Łojasiewicz condition) is the most technically involved result to prove in this paper. We introduce the notion of "fake gradient". Given *fixed* vectors $\{\mathsf{loss}_{i,a}\}_{i\in[n],a\in\{2,\ldots,L\}}$, we define

$$\widehat{\nabla}_k f(W) \stackrel{\mathrm{def}}{=} \sum_{i=1}^n \sum_{\ell=1}^{L-1} (u_{i,\ell})_k \cdot h_{i,\ell} \cdot \mathbf{1}_{(g_{i,\ell+1})_k \geq 0} \tag{6.1}$$

where $u_{i,\ell} \stackrel{\text{def}}{=} \sum_{a=\ell+1}^{L} \mathsf{Back}_{i,\ell+1 \to a}^{\top} \cdot \mathsf{loss}_{i,a}$. Note that if $\mathsf{loss}_{i,a} = Bh_{i,a} - y_{i,a}^*$ is the *true* loss vector, then $\widehat{\nabla}_k f(W)$ will be identical to $\nabla_k f(W)$ by Fact 2.5. Our main technical theorem is the following:

**Theorem 5.** *For every fixed vectors* $\{\mathsf{loss}_{i,a}\}_{i \in [n], a \in \{2,\ldots,L\}}$, *if* $W, A, B$ *are at* random initialization, *then with high probability*

$$\|\widehat{\nabla} f(W)\|_F^2 \geq \widetilde{\Omega}\left(\frac{\delta m}{\mathsf{poly}(\rho)}\right) \times \max_{i,\ell}\{\|\mathsf{loss}_{i,\ell}\|^2\} \ .$$

There are only two conceptually simple steps from Theorem 5 to Theorem 3 (see Appendix F).

- First, one can use the stability lemmas in Section 5 to show that, the fake gradient $\|\widehat{\nabla} f(W + W')\|_F$ after adversarial perturbation $W'$ (with $\|W'\|_2 \leq \frac{1}{\sqrt{m}}$) is also large.

- Second, one can apply $\varepsilon$-net and union bound to turn "fixed $\mathsf{loss}$" into "for all $\mathsf{loss}$". This allows us to turn the lower bound on the fake gradient into a lower bound on the true gradient $\|\nabla_k f(W + W')\|_F$.

Therefore, in the rest of this section, we only sketch the ideas behind proving Theorem 5.

Let $(i^*, \ell^*) = \arg\max_{i,\ell}\{\|\mathsf{loss}_{i,\ell}\|_2\}$ be the sample and layer corresponding to the largest loss. Recall $W, A, B$ are at random initialization.

## 6.1 Indicator and Backward Coordinate Bounds

There are three factors in the notion of fake gradient (6.1): $(u_{i,\ell})_k \in \mathbb{R}$ the backward coordinate, $h_{i,\ell} \in \mathbb{R}^m$ the forward vector, and $\mathbf{1}_{(g_{i,\ell+1})_k \geq 0} \in \{0, 1\}$ the indicator coordinate. We already know very well how the forward vector $h_{i,\ell}$ behaves from the previous sections. Let us provide bounds on the other two factors at the random initialization. (Details in Appendix D.)

Our "backward coordinate bound" controls the value of $(u_{i,\ell})_k$: at random initialization,

$$\left|(u_{i^*,\ell^*})_k\right| \geq \frac{\|\mathsf{loss}_{i^*,\ell^*}\|_2}{\rho} \quad \text{for at least } 1 - o(1) \text{ fraction of } k \in [m] \tag{6.2}$$

The main idea behind proving (6.2) is to use the randomness of $B$. For a fixed $k \in [m]$, it is in fact not hard to show that $\left|(u_{i^*,\ell^*})_k\right|$ is large with high probability. Unfortunately, the randomness of $B$ are shared for different coordinates $k$. We need to also bound the correlation between pairs of coordinates $k_1, k_2 \in [m]$, and resort to MiDiarmid inequality to provide a high concentration bound with respect to all the coordinates.

Our indicator coordinate bound controls the value $(g_{i,\ell+1})_k$ inside the indicator functions $\mathbf{1}_{(g_{i,\ell+1})_k \geq 0}$. It says, letting $\beta_+ \stackrel{\text{def}}{=} \frac{\delta}{\rho^2}$ and $\beta_- \stackrel{\text{def}}{=} \frac{\delta}{\rho^{10}}$, then at random initialization, for at least $\frac{\delta}{\mathsf{poly}(\rho)}$ *fraction of* the coordinates $k \in [m]$,

$$|(g_{i^*,\ell^*+1})_k| \leq \frac{\beta_-}{\sqrt{m}} \quad \text{and} \quad |(g_{i,\ell+1})_k| \geq \frac{\beta_+}{\sqrt{m}} \quad \begin{array}{l} \text{for } i \neq i^* \text{ and } \ell = \ell^*, \text{ or} \\ \text{for } i \in [n] \text{ and } \ell > \ell^*. \end{array} \tag{6.3}$$

This should be quite intuitive to prove, in the following two steps.

- First, there are $\frac{\delta}{\mathsf{poly}(\rho)} m$ coordinates $k \in [m]$ with $|(g_{i^*,\ell^*+1})_k| \leq \frac{\beta_-}{\sqrt{m}}$.

  To show this, we write $(g_{i^*,\ell^*+1})_k = (WU_{\ell^*}U_{\ell^*}h_{i^*,\ell^*} + Ax_{i^*,\ell^*+1})_k$, and prove that for every $z \in \mathbb{R}^{n\ell^*}$ with bounded norms (by $\varepsilon$-net), there are at least $\frac{\delta}{\mathsf{poly}(\rho)} m$ coordinates $k \in [m]$ with $|(WU_{\ell^*}z + Ax_{i^*,\ell^*+1})_k| \leq \frac{\beta_-}{\sqrt{m}}$. This is possible using the independence between $WU_{\ell^*}$ and $A$.

- Then, *conditioning* on the first event happens, we look at $|(g_{i,\ell+1})_k|$ for (1) each $i \neq i^*$ and $\ell = \ell^*$, or (2) each $i \in [n]$ and $\ell > \ell^*$. In both cases, even though the value of $g_{i^*,\ell^*+1}$ is fixed, we still have sufficient fresh new randomness (by invoking (4.3) for case (1) and (4.2) for case (2)). Such additional randomness can make sure that, with high probability (over the fresh new randomness), the value of $(g_{i,\ell+1})_k$ is larger than $\frac{\beta_+}{\sqrt{m}}$.

## 6.2 Thought Experiment: Adding Small Rank-One Perturbation

We now focus on a fixed coordinate $k \in [m]$ satisfying (6.3) and (6.2) in the Section 6.1, and denote by $v_{i^*,\ell^*} \stackrel{\text{def}}{=} \frac{(I-U_{\ell^*-1}U_{\ell^*-1}^\top)h_{i^*,\ell^*}}{\|(I-U_{\ell^*-1}U_{\ell^*-1}^\top)h_{i^*,\ell^*}\|}$.

For analysis purpose, imagine that we apply a small *random* perturbation $W'_k$ to the already-randomly initialized matrix $W$, in the rank-one direction $\mathbf{e}_k v_{i^*,\ell^*}^\top$. Namely, we set $W'_k = g \cdot \mathbf{e}_k v_{i^*,\ell^*}^\top$ where $g \sim \mathcal{N}(0, \theta^2)$ and $\theta$ is a parameter satisfying $\frac{\beta_-}{\sqrt{m}} \ll \theta \ll \frac{\beta_+}{\sqrt{m}}$. Using the fact that $k$ satisfies (6.3), one can show that[11]

(a) $\mathbf{1}_{(g_{i,\ell+1})_k \geq 0}$ stays the same with respect to perturbation $W'_k$, except for $i = i^*$ and $\ell = \ell^*$; and

(b) $\mathbf{1}_{(g_{i^*,\ell^*+1})_k \geq 0}$ can be 0 or 1 each with at least constant probability over $W'_k$.

At the same time, using the fact that $k$ satisfies (6.2), one can show that

(c) $h_{i,\ell}$ and $(u_{i,\ell})_k$ does not change by much (owing to Section 5); and

(d) $|(u_{i^*,\ell^*+1})_k| \geq \frac{\|\mathsf{loss}_{i^*,\ell^*}\|_2}{\rho}$ is large (owing to (6.2)).

Putting (a), (b), (c), and (d) together, we know for such specially chosen $k$, at least with constant probability over the random perturbation of $W'_k$,

$$\|\widehat{\nabla}_k f(W + W'_k)\|_F \geq \Omega\Big(\frac{\|\mathsf{loss}_{i^*,\ell^*}\|_2}{\rho}\Big) \ . \tag{6.4}$$

## 6.3 Real Proof: Randomness Decomposition and McDiarmid's Inequality

There are only two main differences between (6.4) and our desired Theorem 5. First, (6.4) gives a gradient lower bound at $W + W'_k$, while in Theorem 5 we need a gradient lower bound at random initialization $W$. Second, (6.4) gives a lower bound on $\widehat{\nabla}_k f(\cdot)$ with *constant probability* for a small fraction of good coordinates $k$, but in Theorem 5 we need a lower bound for the entire $\widehat{\nabla} f(\cdot)$.

**Randomness Decomposition.** To fix the first issue, we resort to a randomness decomposition technique at least tracing back to the smooth analysis of Spielman and Teng [60]:

**Proposition 6.1.** *Given small constant $\theta \in (0,1)$ and $m$-dimensional random $g \sim \mathcal{N}(0, \frac{1}{m}I)$, we can rewrite $g = g_1 + g_3$ where $g_1$ follows from $\mathcal{N}(0, \frac{1}{m}I)$ and $g_3$ is very close to $\mathcal{N}(0, \frac{\theta^2}{m}I)$.* (Note that there is no contradiction here because $g_1$ and $g_3$ shall be correlated.)

---
[11]Specifically,

- For every $\ell < \ell^*$, we have $(g_{i,\ell+1})_k$ is unchanged with respect to $W'_k$. This is because $g_{i,\ell+1}$ only depends on the randomness of $WU_\ell$, but $v_{i^*,\ell^*}$ is orthogonal to the columns of $U_\ell$.
- For $i = i^*$ and $i = \ell^*$, we have $(g_{i,\ell+1})_k$ will change sign with constant probability with respect to $W'_k$. This is because $\|W'_k\|$ is above $\frac{\beta_-}{\sqrt{m}}$— the original $(g_{i,\ell+1})_k$ before perturbation.
- For any other $i$ and $\ell$, we have $(g_{i,\ell+1})_k$ will not change sign with high probability with respect to $W'_k$. This is because $\|W'_k\|$ is below $\frac{\beta_+}{\sqrt{m}}$— the original $(g_{i,\ell+1})_k$ before perturbation.

*Proof.* Let $g_1, g_2 \in \mathbb{R}^m$ be two independent random vectors sampled from $\mathcal{N}(0, \frac{1}{m}I)$. We can couple them and make sure $g = \sqrt{1-\theta^2}g_1 + \theta g_2$. We now choose $g_3 = \theta g_2 - (1-\sqrt{1-\theta^2})g_1$. Since $(1-\sqrt{1-\theta^2}) \leq O(\theta^2)$, when $\theta$ is sufficiently small, we know that $g_3$ is close to being generated from distribution $\mathcal{N}(0, \frac{\theta^2}{m}I)$. □

Using Proposition 6.1, for each good coordinate $k$, instead of "adding" perturbation $W'_k$ to $W$, we can instead decompose $W$ into $W = W_0 + W'_k$, where $W_0$ is distributed in the same way as $W$. In other words, $W_0$ is also at random initialization. If this idea is carefully implemented, one can immediately turn (6.4) into

$$\|\widehat{\nabla}_k f(W)\|_F = \|\widehat{\nabla}_k f(W_0 + W'_k)\|_F \geq \Omega\Big(\frac{\|\mathsf{loss}_{i^*,\ell^*}\|_2}{\rho}\Big) \ . \tag{6.5}$$

**Extended McDiarmid's Inequality.** To fix the second issue, one may wish to consider all the indices $k \in [m]$ satisfying (6.3) and (6.2). Since there are at least $\frac{\delta}{\mathsf{poly}(\rho)}$ fraction of such coordinates, if all of them satisfied (6.5), then we would have already proved Theorem 5. Unfortunately, neither can we apply Chernoff bound (because the events with different $k \in [m]$ are correlated), nor can we apply union bound (because the event occurs only with constant probability).

Our technique is to resort to (an extended probabilistic variant of) McDiarmid's inequality (see Appendix A.6) in a very non-trivial way to boost the confidence.

Give any fixed subset $N$ of cardinality $|N| = (\mathsf{poly}(\rho)/\delta)^2$, one can show that there again exists $\frac{\delta}{\mathsf{poly}(\rho)}$ fraction of coordinate $k \in N$ satisfying (6.3) and (6.2).[12] Now, instead of decomposing $W = W_0 + W'_k$, we decompose it as $W = W_1 + W'_N$ for $W'_N = u_N v^\top_{i^*,\ell^*}$ where $u_N$ is only supported on coordinates $k \in N$. In other words, we *simultaneously* perturb in the directions of $W'_k$ for all $k \in N$. Since this perturbation is small enough—i.e., $\|W'_N\|_2 \leq \frac{\mathsf{poly}(\varrho)}{\sqrt{m}}$— one can show that (6.5) remains true, that is, for a large fraction of $k \in N$, with at least constant probability over $W'_N$:

$$\|\widehat{\nabla}_k f(W_1 + W'_N)\|_F \geq \Omega\Big(\frac{\|\mathsf{loss}_{i^*,\ell^*}\|_2}{\rho}\Big) \ . \tag{6.6}$$

In order to apply McDiarmid's, we next need to bound on the difference between $\|\widehat{\nabla}_k f(W_1 + W'_N)\|_F$ and $\|\widehat{\nabla}_k f(W_1 + W'_N + W''_j)\|_F$ for an arbitrary (but small) perturbation $W''_j$ (in the direction of $\mathbf{e}_j v^\top_{i^*,\ell^*}$). We show that,

$$\sum_{k \in [N]} \Big|\|\widehat{\nabla}_k(W_1 + W'_N)\|_2^2 - \|\widehat{\nabla}_k(W_1 + W'_N + W''_j)\|_2^2\Big| \leq O\Big(\rho^8 + \frac{|N|}{m^{1/6}}\Big) \ . \tag{6.7}$$

In other words, although there are $|N|$ difference terms, their total summation only grows in rate $\frac{|N|}{m^{1/6}}$ according to (6.7). After applying a variant of McDiarmid's inequality, we derive that with *high probability* over $W'_N$, it satisfies

$$\sum_{k \in N} \|\widehat{\nabla}_k f(W)\|_F^2 = \sum_{k \in N} \|\widehat{\nabla}_k f(W_1 + W'_N)\|_F^2 \geq \Omega\Big(\frac{\|\mathsf{loss}_{i^*,\ell^*}\|_2^2}{\mathsf{poly}(\rho)}|N|\Big) \ . \tag{6.8}$$

Finally, by sampling sufficiently many random sets $N$ to cover the entire space $[m]$, we can show

$$\|\widehat{\nabla} f(W)\|_F^2 = \sum_{k \in [m]} \|\widehat{\nabla}_k f(W)\|_F^2 \geq \Omega\Big(\frac{\|\mathsf{loss}_{i^*,\ell^*}\|_2^2}{\mathsf{poly}(\rho)}m\Big) \ .$$

and therefore Theorem 5 holds.

---
[12]We use $(\mathsf{poly}(\rho))^2$ to emphasize that the first polynomial needs to be bigger than the second $\mathsf{poly}(\rho)$.

# 7 Proof Sketch of Theorem 4: Objective Semi-Smoothness

The objective semi-smoothness Theorem 4 turns out to be much simpler to prove than Theorem 3. It only relies on Section 4 and 5, and does not need randomness decomposition or McDiarmid's inequality. (Details in Appendix G.)

Recall that in Theorem 4, $\widetilde{W}, A, B$ are at random initialization. $\breve{W}$ is an adversarially chosen matrix with $\|\breve{W} - \widetilde{W}\| \leq \frac{\mathsf{poly}(\varrho)}{\sqrt{m}}$, and $W'$ is some other adversarial perturbation on top of $\breve{W}$, satisfying $\|W'\| \leq \frac{\tau_0}{\sqrt{m}}$. We denote by

- $\widetilde{D}_{i,\ell}, \widetilde{g}_{i,\ell}, \widetilde{h}_{i,\ell}$ respectively the values of $D_{i,\ell}, g_{i,\ell}, h_{i,\ell}$ determined by weight matrix $\widetilde{W}$;
- $\breve{D}_{i,\ell}, \breve{g}_{i,\ell}, \breve{h}_{i,\ell}, \breve{\mathsf{loss}}_{i,\ell}$ respectively those of $D_{i,\ell}, g_{i,\ell}, h_{i,\ell}$ and $\mathsf{loss}_{i,\ell}$ at weight matrix $\breve{W}$; and
- $D_{i,\ell}, g_{i,\ell}, h_{i,\ell}$ respectively the values of $D_{i,\ell}, g_{i,\ell}, h_{i,\ell}$ at weight matrix $W = \breve{W} + W'$.

Our main tool is to derive the following strong formula for $h_{i,\ell} - \breve{h}_{i,\ell}$: there exist diagonal matrices $D''_{i,\ell} \in \mathbb{R}^{m \times m}$ with entries in $[-1, 1]$ and sparsity $\|D''_{i,\ell}\|_0 \leq O(m^{2/3})$ such that,

$$h_{i,\ell} - \breve{h}_{i,\ell} = \sum_{a=1}^{\ell-1} (\breve{D}_{i,\ell} + D''_{i,\ell}) \breve{W} \cdots \breve{W} (\breve{D}_{i,a+1} + D''_{i,a+1}) W' h_{i,a} \tag{7.1}$$

In particular, (7.1) implies $\|h_{i,\ell} - \breve{h}_{i,\ell}\| \leq O(L^9) \|W'\|_2$ after careful linear-algebraic manipulations (esp. using (4.5)). The main take-away message from (7.1) is that, this difference $h_{i,\ell} - \breve{h}_{i,\ell}$ is proportional to the norm of the perturbation, $\|W'\|_2$, no matter how small it is. In contrast, in (4.1), we only derived a weak upper bound of the form $\|h_{i,\ell} - \widetilde{h}_{i,\ell}\| \leq O(m^{-1/2})$. Nevertheless, the proof of (7.1) relies on (4.1), so we are not duplicating proofs.

Finally, we carefully derive that

$$f(\breve{W} + W') - f(W) - \langle \nabla f(W), W' \rangle$$
$$= \sum_{i=1}^{n} \sum_{\ell=2}^{L} \breve{\mathsf{loss}}_{i,\ell}^\top B \left( (h_{i,\ell} - \breve{h}_{i,\ell}) - \sum_{a=1}^{\ell-1} \breve{D}_{i,\ell} \breve{W} \cdots \breve{W} \breve{D}_{i,a+1} W' \breve{h}_{i,a} \right) + \frac{1}{2} \|B(h_{i,\ell} - \breve{h}_{i,\ell})\|^2 \tag{7.2}$$

and plug (7.1) into (7.2) to derive our final Theorem 4.

# 8 Appendix Roadmap

- Appendix A recalls some old lemmas and derives some new lemmas in probability theory.
- Appendix B serves for Section 4, the basic properties at random initialization.
- Appendix C serves for Section 5, the stability after adversarial perturbation.
- Appendix D, E and F together serve for Section 6 and prove Theorem 3, the Polyak-Łojasiewicz condition and gradient upper bound. In particular:
  - Appendix D serves for Section 6.1 (the indicator and backward coordinate bounds).
  - Appendix E serves for Section 6.3 (the randomness decomposition and McDiarmid's inequality) and proves Theorem 5.
  - Appendix F shows how to go from Theorem 5 to Theorem 3.
- Appendix G serves for Section 7, the proof of Theorem 4, the objective semi-smoothness.
- Appendix H gives the final proof for Theorem 1, the GD convergence theorem.

- Appendix I gives the final proof for Theorem 2, the SGD convergence theorem.

**Parameters.** We also summarize a few parameters we shall use in the proofs.

- In Definition D.1, we shall introduce two parameters
$$\beta_+ \stackrel{\text{def}}{=} \frac{\delta}{\rho^2} \text{ and } \beta_- \stackrel{\text{def}}{=} \frac{\delta}{\rho^{10}}$$
to control the thresholds of indicator functions (recall Section 6.1).

- In Definition E.3, we shall introduce parameter
$$\theta \in \left[\rho^4 \cdot \beta_-, \rho^{-3} \cdot \beta_+\right]$$
to describe how much randomness we want to decompose out of $W$ (recall Section 6.2).

- In (E.13) of Appendix E.4, we shall choose
$$N = \frac{\rho^{22}}{\beta_-^2}$$
which controls the size of the set $N$ where we apply McDiarmid's inequality (recall Section 6.3).

# Appendix

## Table of Contents

# A Preliminaries on Probability Theory

The goal of this section is to present a list of probability tools.
- In Section A.1, we recall how to swap randomness.
- In Section A.2, we recall concentration bounds for the chi-square distribution.
- In Section A.3, we proved a concentration bound of sum of squares of ReLU of Gaussians.
- In Section A.4 and Section A.5, we show some properties for random Gaussian vectors.
- In Section A.6, we recall the classical McDiarmid's inequality and then prove a general version of it.

## A.1 Swapping Randomness

**Fact A.1** (probability splitting). *If $f(X,Y)$ holds with probability at least $1 - \varepsilon$, then*
- *with probability at least $1 - \sqrt{\varepsilon}$ (over randomness of $X$), the following event holds,*
  - *$f(X,Y)$ holds with probability at least $1 - \sqrt{\varepsilon}$ (over randomness of $Y$).*

*In other words,*
$$\mathbf{Pr}_X\left[\mathbf{Pr}_Y[f(X,Y)] \geq 1 - \sqrt{\varepsilon}\right] \geq 1 - \sqrt{\varepsilon}.$$

**Fact A.2** (swapping probability and expectation). *If $\mathbf{Pr}_{X,Y}[f(X,Y) \geq a] \geq \varepsilon$, then*
$$\Pr_X\left[\mathbb{E}_Y[f(X,Y)] \geq a\varepsilon/2\right] \geq \varepsilon/2.$$

*Proof.* We prove it by making a contradiction,

Suppose
$$\Pr_X\left[\mathbb{E}_Y[f(X,Y)] \leq a\varepsilon/2\right] \geq 1 - \varepsilon/2,$$

which implies that
$$\Pr_{X,Y}[f(X,Y) \geq a|X] \leq \varepsilon.$$

□

## A.2 Concentration of Chi-Square Distribution

**Lemma A.3** (Lemma 1 on page 1325 of Laurent and Massart [35]). *Let $X \sim \mathcal{X}_k^2$ be a chi-squared distributed random variable with $k$ degrees of freedom. Each one has zero mean and $\sigma^2$ variance. Then*
$$\mathbf{Pr}[X - k\sigma^2 \geq (2\sqrt{kt} + 2t)\sigma^2] \leq \exp(-t)$$
$$\mathbf{Pr}[k\sigma^2 - X \geq 2\sqrt{kt}\sigma^2] \leq \exp(-t)$$

One straightforward application is

**Lemma A.4.** *Let $x_1, x_2, \cdots, x_n$ denote i.i.d. samples from $\mathcal{N}(0, \sigma^2)$. For any $b \geq 1$, we have*
$$\mathbf{Pr}\left[|\|x\|_2^2 - n\sigma^2| \geq \frac{n}{b}\sigma^2\right] \leq 2\exp(-n/(8b^2)).$$

*Proof.* We choose $t = k/(8b^2)$ in Lemma A.3,
$$\mathbf{Pr}\left[|\|x\|_2^2 - n\sigma^2| \geq (\frac{2n}{\sqrt{8b}} + \frac{2n}{8b^2})\sigma^2\right] \leq 2\exp(-n/(8b^2)).$$

Since $\frac{2n}{\sqrt{8b}} + \frac{2n}{8b^2} \leq \frac{2n}{\sqrt{8b}} + \frac{2n}{8b} \leq n/b$. Thus,

$$\Pr\left[\left|\|x\|_2^2 - n\sigma^2\right| \geq \frac{n}{b}\sigma^2\right] \leq 2\exp(-n/(8b^2)),$$

which completes the proof. □

**Lemma A.5.** *Let $x_1, x_2, \cdots, x_m$ denote i.i.d. samples from $\mathcal{N}(0, 1)$, and $y_i = \max\{x_i^2 - \log m, 0\}$. We have*

$$\Pr\left[\sum_{i=1}^m y_i \geq 2\sqrt{m}\right] \leq e^{-\Omega(\sqrt{m})}.$$

*Proof.* First of all, letting $g = \sqrt{\log m}$,

$$\mathbb{E}[y_i] = \int_g^\infty \frac{\exp\left(-\frac{x^2}{2}\right)(x-g)^2}{\sqrt{2\pi}}\,dx = \frac{1}{2}\left(g^2+1\right)\mathrm{erfc}\left(\frac{g}{\sqrt{2}}\right) - \frac{e^{-\frac{g^2}{2}}g}{\sqrt{2\pi}} \leq \frac{1}{e^{\frac{g^2}{2}}} = \frac{1}{\sqrt{m}}.$$

On the other hand, each random variable $y_i$ is $O(1)$-subgaussian. By subgaussian concentration,

$$\Pr\left[\sum_{i=1}^m y_i \geq 2\sqrt{m}\right] \leq e^{-\Omega(\sqrt{m})}$$

□

## A.3 Concentration of Sum of Squares of ReLU of Gaussians

**Lemma A.6** (Upper bound). *Given $n$ i.i.d. Gaussian random variables $x_1, x_2, \cdots, x_n \sim \mathcal{N}(0, \sigma^2)$, we have*

$$\Pr\left[\left(\sum_{i=1}^n \max(x_i, 0)^2\right)^{1/2} < (1+\varepsilon)\sqrt{n/2}\sigma\right] \geq 1 - \exp(-\varepsilon^2 n/100).$$

*Proof.* Using Chernoff bound, we know that with probability $1 - \exp(-\varepsilon^2 n/6)$, $\sum_{i=1}^n \max(x_i, 0)^2$ is a at most degree-$(1+\varepsilon)\frac{n}{2}$ Chi-square random variable. Let us say this is the first event.

Using Lemma A.3, we have

$$\Pr[X \geq k\sigma^2 + (2\sqrt{kt} + 2t)\sigma^2] \leq \exp(-t)$$
$$\implies \Pr[X \geq k\sigma^2 + (2\varepsilon k + 2\varepsilon^2 k)\sigma^2] \leq \exp(-\varepsilon^2 k) \quad \text{by choosing } t = \varepsilon^2 k$$
$$\implies \Pr[X \geq k(1 + 4\varepsilon)\sigma^2] \leq \exp(-\varepsilon^2 k)$$
$$\implies \Pr\left[X \geq (1+\varepsilon)(1+4\varepsilon)\frac{n}{2}\sigma^2\right] \leq \exp(-\varepsilon^2(1+\varepsilon)n/2) \quad \text{by } k = (1+\varepsilon)n/2$$

Thus, we have with probability at least $1 - \exp(-\varepsilon^2 n/2)$,

$$X \leq (1+4\varepsilon)^2 \frac{n}{2}\sigma^2.$$

Let the above event denote the second event.

By taking the union bound of two events, we have with probability $1 - 2\exp(-\varepsilon^2 n/6)$

$$\left(\sum_{i=1}^n \max(x_i, 0)^2\right)^{1/2} \leq X \leq (1+4\varepsilon)\sqrt{\frac{n}{2}}\sigma$$

Then rescaling the $\varepsilon$, we get the desired result. □

**Lemma A.7** (Lower bound). *Given $n$ i.i.d. Gaussian random variables $x_1, x_2, \cdots, x_n \sim \mathcal{N}(0, \sigma^2)$, we have*

$$\mathbf{Pr}\left[\left(\sum_{i=1}^n \max(x_i, 0)^2\right)^{1/2} > (1-\varepsilon)\sqrt{n/2}\sigma\right] \geq 1 - \exp(\varepsilon^2 n/100)$$

*Proof.* Using Chernoff bound, we know that with probability $1 - \exp(-\varepsilon^2 n/6)$, $\sum_{i=1}^n \max(x_i, 0)^2$ is a at most degree-$(1-\varepsilon)\frac{n}{2}$ Chi-square random variable. Let us say this is the first event.

Using Lemma A.3, we have

$$\mathbf{Pr}[X \leq k\sigma^2 - 2\sqrt{kt}\sigma^2] \leq \exp(-t)$$
$$\implies \mathbf{Pr}[X \leq k\sigma^2 - 2\varepsilon k\sigma^2] \leq \exp(-\varepsilon^2 k) \qquad \text{by choosing } t = \varepsilon^2 k$$
$$\implies \mathbf{Pr}[X \leq k\sigma^2(1 - 2\varepsilon)] \leq \exp(-\varepsilon^2 k)$$
$$\implies \mathbf{Pr}\left[X \leq (1-\varepsilon)(1-2\varepsilon)\frac{n}{2}\sigma^2\right] \leq \exp(-\varepsilon^2(1-\varepsilon)n/2) \qquad \text{by } k = (1-\varepsilon)n/2$$

Thus, we have with probability at least $1 - \exp(-\varepsilon^2 n/4)$,

$$X \geq (1-2\varepsilon)^2 \frac{n}{2}\sigma^2.$$

Let the above event denote the second event.

By taking the union bound of two events, we have with probability at least $1 - 2\exp(-\varepsilon^2 n/6)$,

$$\left(\sum_{i=1}^n \max(x_i, 0)^2\right)^{1/2} \geq X \geq (1-2\varepsilon)\frac{n}{2}\sigma.$$

Then rescaling the $\varepsilon$, we get the desired result. □

Combining Lemma A.7 and Lemma A.6, we have

**Lemma A.8** (Two sides bound). *Given $n$ i.i.d. Gaussian random variables $x_1, x_2, \cdots, x_n \sim \mathcal{N}(0, \sigma^2)$, let $\phi(a) = \max(a, 0)^2$. We have*

$$\mathbf{Pr}_x\left[\|\phi(x)\|_2 \in ((1-\varepsilon)\sqrt{n/2}\sigma, (1+\varepsilon)\sqrt{n/2}\sigma)\right] \geq 1 - 2\exp(-\varepsilon^2 n/100)$$

**Corollary A.9** (Two sides bound for single matrix). *Let $x \in \mathbb{R}^n$ denote a fixed vector. Given a random Gaussian matrix $A \in \mathbb{R}^{m \times n}$ where each entry is i.i.d. sampled from $\mathcal{N}(0, 2\sigma^2/m)$.*

$$\mathbf{Pr}_A[\|\phi(Ax)\|_2 \in ((1-\varepsilon)\|x\|_2\sigma, (1+\varepsilon)\|x\|_2\sigma)] \geq 1 - 2\exp(-\varepsilon^2 m/100).$$

*Proof.* For each $i \in [m]$, let $y_i = (Ax)_i$. Then $y_i \sim \mathcal{N}(0, \widetilde{\sigma}^2)$, where $\widetilde{\sigma}^2 = 2\sigma^2/m \cdot \|x\|_2^2$. Using Corollary A.8, we have

$$\mathbf{Pr}_y\left[\|\phi(y)\|_2 \in ((1-\varepsilon)\sqrt{m/2}\widetilde{\sigma}, (1+\varepsilon)\sqrt{m/2}\widetilde{\sigma})\right] \geq 1 - 2\exp(-\varepsilon m/100)$$

which implies

$$\mathbf{Pr}_y[\|\phi(y)\|_2 \in ((1-\varepsilon)\|x\|_2\sigma, (1+\varepsilon)\|x\|_2\sigma)] \geq 1 - 2\exp(-\varepsilon^2 m/100).$$

Since $y = Ax$, thus we complete the proof. □

**Corollary A.10** (Two sides bound for multiple matrices). *Let $x_1, x_2, \cdots, x_k$ denote $k$ fixed vectors where $x_i \in \mathbb{R}^{n_i}$. Let $x = [x_1^\top \ x_2^\top \ \cdots \ x_k^\top]^\top \in \mathbb{R}^n$ where $n = \sum_{i=1}^k n_i$. Let $A_1, A_2, \cdots A_k$ denote* $k$ independent random Gaussian matrices where each entry of $A_i \in \mathbb{R}^{m \times n_i}$ is i.i.d. sampled from $\mathcal{N}(0, 2\sigma_i^2/m)$ for each $i \in [k]$. Let $A = [A_1 \ A_2 \ \cdots \ A_k]$. We have

$$\Pr_A \left[ \|\phi(Ax)\|_2 \in \left( (1-\varepsilon)\left(\sum_{i=1}^k \|x_i\|_2^2 \sigma_i^2\right)^{1/2}, (1+\varepsilon)\left(\sum_{i=1}^k \|x_i\|_2^2 \sigma_i^2\right)^{1/2} \right) \right] \geq 1 - 2\exp(-\varepsilon^2 m/100).$$

*Proof.* It is similar as Corollary A.9. □

**Fact A.11** (see e.g. [3])**.** *Let $h, q \in \mathbb{R}^p$ be fixed vectors and $h \neq 0$, $W \in \mathbb{R}^{m \times p}$ be random matrix with i.i.d. entries $W_{i,j} \sim \mathcal{N}(0, \frac{2}{m})$, and vector $v \in \mathbb{R}^m$ defined as $v_i = \mathbf{1}_{\langle W_i, h+q \rangle \geq 0} \langle W_i, h \rangle$. Then,*

- *$|v_i|$ follows i.i.d. from the following distribution: with half probability $|v_i| = 0$, and with the other half probability $|v_i|$ follows from folded Gaussian distributions $|\mathcal{N}(0, \frac{2\|h\|^2}{m})|$.*
- *$\frac{m\|v\|^2}{2\|h\|^2}$ is in distribution identical to $\chi_\omega^2$ (chi-square distribution of order $\omega$) where $\omega$ follows from binomial distribution $\mathcal{B}(m, 1/2)$ ($m$ trials each with success rate $1/2$).*

### A.4 Gaussian Vector Percentile: Center

**Fact A.12.** *Suppose $x \sim \mathcal{N}(0, \sigma^2)$ is a Gaussian random variable. For any $t \in (0, \sigma]$ we have*

$$\Pr[x \geq t] \in \left[ \frac{1}{2}\left(1 - \frac{4}{5}\frac{t}{\sigma}\right), \frac{1}{2}\left(1 - \frac{2}{3}\frac{t}{\sigma}\right) \right].$$

*Similarly, if $x \sim \mathcal{N}(\mu, \sigma^2)$, for any $t \in (0, \sigma]$, we have*

$$\Pr[|x| \geq t] \in \left[ 1 - \frac{4}{5}\frac{t}{\sigma}, 1 - \frac{2}{3}\frac{t}{\sigma} \right].$$

**Lemma A.13.** *Let $x \sim \mathcal{N}(0, \sigma^2 I)$. For any $\alpha \in (0, 1/2)$, we have with probability at least $1 - \exp(-\alpha^2 m/100)$,*

- *there exists at least $\frac{1}{2}(1-\alpha)$ fraction of $i$ such that $x_i \geq 5\alpha\sigma/16$, and*
- *there exists at least $\frac{1}{2}(1-\alpha)$ fraction of $i$ such that $x_i \leq -5\alpha\sigma/16$.*

*Proof.* Let $c_1 = 4/5$. For each $i \in [m]$, we define random variable $y_i$ as

$$y_i = \begin{cases} 1, & \text{if } x_i \geq \alpha\sigma/(4c_1); \\ 0, & \text{otherwise.} \end{cases}$$

Let $p = \Pr[y_i = 1]$. Using Fact A.12 with $t = \alpha\sigma/(4c_1)$, we know that $p \geq \frac{1}{2}(1 - \alpha/4)$. Letting $Y = \sum_{i=1}^m y_i$, we have $\mu = \mathbb{E}[Y] = mp \geq \frac{1}{2}(1 - \alpha/4)m$. Using Chernoff bound, we have

$$\Pr[Y \leq (1-\delta)\mu] \leq \exp(-\delta^2 \mu/2).$$

Choosing $\delta = \frac{1}{4}\alpha$, we have

$$\Pr[Y \leq (1 - \alpha/4)\mu] \leq \exp(-\alpha^2 \mu/32)$$

Further we have

$$\begin{aligned}
\mathbf{Pr}[Y \leq \frac{1}{2}(1-\alpha)m] &\leq \mathbf{Pr}[Y \leq (1-\alpha/4)\frac{1}{2}(1-\alpha/4)m] && \text{by } (1-\alpha) \leq (1-\alpha/4)^2 \\
&\leq \mathbf{Pr}[Y \leq (1-\alpha/4)\mu] && \text{by } \mu \geq \frac{1}{2}(1-\alpha/4)m \\
&= \mathbf{Pr}[Y \leq (1-\delta)\mu] && \text{by } \delta = \alpha/4 \\
&\leq \exp(-\alpha^2 \mu/32) \\
&\leq \exp(-\alpha^2 \frac{1}{2}(1-\alpha/4)m/32) && \text{by } \mu \geq \frac{1}{2}(1-\alpha/4)m \\
&= \exp(-\alpha^2(1-\alpha/4)m/64) \ . && \square
\end{aligned}$$

We provide the definition of $(\alpha, \sigma)$-good. Note that this definition will be used often in the later proof.

**Definition A.14** (($(\alpha, \sigma)$-good)**.** *Given $w \in \mathbb{R}^m$, we say $w$ is $(\alpha, \sigma)$-good if the following two conditions holding:*

- *there are at least $\frac{1}{2}(1-\alpha)$ fraction coordinates satisfy that $w_i \geq \alpha\sigma$; and*
- *there are at least $\frac{1}{2}(1-\alpha)$ fraction coordinates satisfy that $w_i \leq -\alpha\sigma$.*

Lemma A.13 gives the following immediate corollary:

**Corollary A.15** (random Gaussian is $(\alpha, \sigma/4)$-good)**.** *Let $x \sim \mathcal{N}(0, \sigma^2 I)$. For any $\alpha \in (0, 1/2)$, we have with probability at least $1 - \exp(-\alpha^2 m/100)$ that $x$ is $(\alpha, \sigma/4)$-good.*

**Corollary A.16.** *Let $x_1, x_2, \cdots, x_k$ be $k$ fixed vectors where $x_i \in \mathbb{R}^{n_i}$, and $A_1, A_2, \cdots, A_k$ be $k$ independent random Gaussian matrices where each entry of $A_i \in \mathbb{R}^{m \times n_i}$ is i.i.d. sampled from $\mathcal{N}(0, \sigma_i^2)$ for each $i \in [k]$. Denote by $x = (x_1, \ldots, x_k) \in \mathbb{R}^n$ for $n = \sum_{i=1}^k n_i$, $A = [A_1, A_2, \cdots, A_k] \in \mathbb{R}^{m \times n}$, and $\sigma = \left(\sum_{i=1}^k \sigma_i^2 \|x_i\|_2^2\right)^{1/2}$. For any fixed parameter $\alpha \in (0, 1/2)$, we have*

$$Ax = A_1 x_1 + \cdots A_k x_k \text{ is } (\alpha, \sigma/4)\text{-good with probability at least } 1 - \exp(-\alpha^2 m/100).$$

*Proof.* It is clear that $Ax$ follows from a Gaussian distribution $\mathcal{N}\big(0, \big(\sum_{i=1}^k \sigma_i^2 \|x_i\|_2^2\big)I\big)$, so we can directly apply Corollary A.15. $\square$

### A.5 Gaussian Vector Percentile: Tail

**Lemma A.17.** *Suppose $W \in \mathbb{R}^{m \times n}$ is a random matrix with entries drawn i.i.d. from $\mathcal{N}\big(0, \frac{2}{m}\big)$. Given $d \leq s \leq m$, with probability at least $1 - e^{-\Omega(s \log^2 m)}$, for all $x \in \mathbb{R}^n$, letting $y = Wx$, we can write $y = y_1 + y_2$ with*

$$\|y_1\| \leq \frac{\sqrt{s} \log^2 m}{\sqrt{m}} \cdot \|x\| \quad \text{and} \quad \|y_2\|_\infty \leq \frac{\log m}{\sqrt{m}} \cdot \|x\| \ .$$

*Proof of Lemma A.17.* Without loss of generality we only prove the result for $\|x\| = 1$.

Fixing any such $x$ and letting $\beta = \frac{\log m}{2\sqrt{m}}$, we have $y_i \sim \mathcal{N}\big(0, \frac{2}{m}\big)$ so for every $p \geq 1$, by Gaussian tail bound

$$\mathbf{Pr}[|y_i| \geq \beta p] \leq e^{-\Omega(\beta^2 p^2 m)} \ .$$

Since $\beta^2 p^2 m \geq \beta^2 m \gg \Omega(\log m)$, we know that if $|y_i| \geq \beta p$ occurs for $q/p^2$ indices $i$ out of $[m]$, this cannot happen with probability more than

$$\binom{m}{q/p^2} \times \left(e^{-\Omega(\beta^2 p^2 m)}\right)^{q/p^2} \leq e^{\frac{q}{p^2}\left(O(\log m) - \Omega(\beta^2 p^2 m)\right)} \leq e^{-\Omega(\beta^2 qm)} \ .$$

In other words,

$$\mathbf{Pr}\left[|\{i \in [m]\colon |y_i| \geq \beta p\}| > q/p^2\right] \leq e^{-\Omega(\beta^2 qm)} \ .$$

Finally, by applying union bound over $p = 1, 2, 4, 8, 16, \ldots$ we have with probability $\geq 1 - e^{-\Omega(\beta^2 qm)} \cdot \log q$,

$$\sum_{i \colon |y_i| \geq \beta} y_i^2 \leq \sum_{k=0}^{\lceil \log q \rceil} (2^{k+1}\beta)^2 \left|\{i \in [m]\colon |y_i| \geq 2^k \beta\}\right| \leq \sum_{k=0}^{\lceil \log q \rceil} (2^{k+1}\beta)^2 \cdot \frac{q}{2^{2k}} \leq 4q\beta^2 \log q \ . \quad (A.1)$$

In other words, vector $y$ can be written as $y = y_1 + y_2$ where $\|y_2\|_\infty \leq \beta$ and $\|y_1\|^2 \leq 4q\beta^2 \log q$.

At this point, we can choose $q = \frac{s \log^2 m}{m\beta^2} = 4s$ so the above event happens with probability at least $1 - e^{-\Omega(\beta^2 qm)} \geq 1 - e^{-\Omega(s \log^2 m)}$. Finally, applying standard $\varepsilon$-net argument over all unit vectors $x \in \mathbb{R}^n$, we have for each such $x \in \mathbb{R}^n$, we can decompose $y = Wx$ into $y = y_1 + y_2$ where

$$\|y_2\|_\infty \leq 2\beta = \frac{\log m}{\sqrt{m}} \text{ and } \|y_1\|^2 \leq \frac{8s \log^3 m}{m} \ . \qquad \square$$

## A.6 McDiarmid's Inequality and An Extension

We state the standard McDiarmid's inequality,

**Lemma A.18** (McDiarmid's inequality). *Consider independent random variables $x_1, \cdots, x_n \in \mathcal{X}$ and a mapping $f\colon \mathcal{X}^n \to \mathbb{R}$. If for all $i \in [n]$ and for all $y_1, \cdots, y_n, y_i' \in \mathcal{X}$, the function $f$ satisfies*

$$|f(y_1, \cdots, y_{i-1}, y_i, y_{i+1}, \cdots, y_n) - f(y_1, \cdots, y_{i-1}, y_i', y_{i+1}, \cdots, y_n)| \leq c_i.$$

*Then*

$$\mathbf{Pr}[f(x_1, \cdots, x_n) - \mathbb{E} f \geq t] \geq \exp(\frac{-2t^2}{\sum_{i=1}^n c_i^2}),$$

$$\mathbf{Pr}[f(x_1, \cdots, x_n) - \mathbb{E} f \leq -t] \geq \exp(\frac{2t^2}{\sum_{i=1}^n c_i^2}).$$

We prove a more general version of McDiarmid's inequality,

**Lemma A.19** (McDiarmid extension). *Let $w_1, \ldots, w_N$ be independent random variables and $f\colon (w_1, \ldots, w_N) \mapsto [0, 1]$. Suppose it satisfies:*

- $\mathbb{E}_{w_1, \ldots, w_N}[f(w_1, \ldots, w_N)] \geq \mu$, *and*
- *With probability at least $1 - p$ over $w_1, \ldots, w_N$, it satisfies*

$$\forall k \in [N], \forall w_k''\colon \ \left|f(w_{-k}, w_k) - f(w_{-k}, w_k'')\right| \leq c \qquad (A.2)$$

*Then, $\mathbf{Pr}[f(w_1, \ldots, w_N) \geq \mu/2] \geq 1 - N^2 \sqrt{p} - e^{\Omega(\frac{-\mu^2}{N(c^2+p)})} \ .$*

*Proof of Lemma A.19.* For each $t \in [N]$, we have with probability at least $1 - \sqrt{p}$ over $w_1, \ldots, w_t$, it satisfies

$$\mathbf{Pr}_{w_{t+1}, \ldots, w_N}\left[\forall k \in [N], \forall w_k''\colon \ \left|f(w_{-k}, w_k) - f(w_{-k}, w_k'')\right| \leq c\right] \geq 1 - \sqrt{p} \ .$$

Define those $(w_1, \ldots, w_t)$ satisfying the above event to be $K_{\leq t}$.

Define random variable $X_t$ (which depends only on $w_1, \ldots, w_t$) as

$$X_t := \mathop{\mathbb{E}}_{w_{>t}} [f(\vec{w}) \mid w_{\leq t}] \mathbb{1}_{(w_{\leq 1},\ldots,w_{\leq t}) \in K_{\leq 1} \times \cdots \times K_{\leq t}} + N(1 - \mathbb{1}_{(w_{\leq 1},\ldots,w_{\leq t}) \in K_{\leq 1} \times \cdots \times K_{\leq t}})$$

For every $t$ and fixed $w_1, \ldots, w_{t-1}$.

- If $(w_{\leq 1}, \ldots, w_{<t}) \notin K_{\leq 1} \times \cdots \times K_{<t}$, then $X_t = X_{t-1} = N$.
- If $(w_{\leq 1}, \ldots, w_{<t}) \in K_{\leq 1} \times \cdots \times K_{<t}$,
  - If $w_{\leq t} \notin K_{\leq t}$, then $X_t - X_{t-1} = N - \cdots \geq 0$.
  - If $w_{\leq t} \in K_{\leq t}$, then
  
  $$X_t - X_{t-1} = \mathop{\mathbb{E}}_{w_{>t}} [f(w_{<t}, w_t, w_{>t}) \mid w_{\leq t}] - \mathop{\mathbb{E}}_{w_{\geq t}} [f(w_{<t}, w_t, w_{>t}) \mid w_{<t}]$$
  
  Recall from our assumption that, with probability at least $1 - \sqrt{p}$ over $w_t$ and $w_{>t}$, it satisfies
  
  $$\forall w_t'' : \left| f(w_{<t}, w_t'', w_{>t}) - f(w_{<t}, w_t, w_{>t}) \right| \leq c$$
  
  Taking expectation over $w_t$ and $w_{>t}$, we have
  
  $$\forall w_t'' : \mathop{\mathbb{E}}_{w_{>t}} \left[ f(w_{<t}, w_t'', w_{>t}) \right] - \mathop{\mathbb{E}}_{w_{\geq t}} [f(w_{<t}, w_t, w_{>t})] \geq -(c + \sqrt{p})$$
  
  This precisely means $X_t - X_{t-1} \geq c + \sqrt{p}$.

In sum, we have just shown that $X_t - X_{t-1} \geq -(c + \sqrt{p})$ always holds. By applying martingale concentration (with one-sided bound),

$$\mathbf{Pr}[X_N - X_0 \leq -t] \leq \exp\left(\frac{-t^2}{N(c+\sqrt{p})^2}\right)$$

Notice that $X_0 = \mu$ so if we choose $t = \mu/2$, we have

$$\mathbf{Pr}[X_N \geq \mu/2] \geq 1 - \exp\left(\frac{-\mu^2}{N(c+\sqrt{p})^2}\right)$$

Recalling

$$X_N := f(\vec{w})\mathbb{1}_{(w_{\leq 1},\ldots,w_{\leq t}) \in K_{\leq 1} \times \cdots \times K_{\leq t}} + N(1 - \mathbb{1}_{(w_{\leq 1},\ldots,w_{\leq t}) \in K_{\leq 1} \times \cdots \times K_{\leq t}})$$

and we have $X_N = f(w_1, \ldots, w_N)$ with probability at least $1 - N\sqrt{p}$ (and $X_N = N$ with the remaining probabilities). Together, we have the desired theorem.

□

# B Basic Properties at Random Initialization

Recall that the recursive update equation of RNN can be described as follows

$$h_{i,0} = 0 \qquad \forall i \in [n]$$
$$h_{i,\ell} = \phi(W \cdot h_{i,\ell-1} + Ax_{i,\ell}) \qquad \forall i \in [n], \forall \ell \in [L]$$
$$y_{i,\ell} = B \cdot h_{i,\ell} \qquad \forall i \in [n], \forall \ell \in [L]$$

Throughout this section, we assume that matrices $W \in \mathbb{R}^{m \times m}$, $A \in \mathbb{R}^{m \times d_x}$, $B \in \mathbb{R}^{d \times m}$ are at their random initialization position: each entry of $W$ and $A$ is sampled i.i.d. from $\mathcal{N}(0, \frac{2}{m})$ and each entry of $B$ is sampled i.i.d. from $\mathcal{N}(0, \frac{1}{d})$. We recall

**Definition 2.2.** *For each $i \in [n]$ and $\ell \in [L]$, let $D_{i,\ell} \in \mathbb{R}^{m \times m}$ be the diagonal matrix where*

$$(D_{i,\ell})_{k,k} = \mathbf{1}_{(W \cdot h_{i,\ell-1} + Ax_{i,\ell})_k \geq 0} = \mathbf{1}_{(g_{i,\ell})_k \geq 0} \ .$$

*As a result, we can write $h_{i,\ell} = D_{i,\ell} W h_{i,\ell-1}$.*

We introduce two notations that shall repeatedly appear in our proofs.

**Definition B.1** ($U_\ell$). *Let $U_\ell \in \mathbb{R}^{m \times n\ell}$ denote the column orthonormal matrix using Gram-Schmidt*

$$U_\ell \stackrel{\text{def}}{=} \mathsf{GS}(h_{1,1}, \ldots, h_{n,1}, h_{1,2}, \ldots, h_{n,2}, \ldots, h_{1,\ell}, \ldots, h_{n,\ell})$$

**Definition B.2** ($v_{i,\ell}$). *For each $i \in [n]$, $\ell \in [L]$, we define vector $v_{i,\ell} \in \mathbb{R}^m$ as*

$$v_{i,\ell} = \frac{(I - U_{\ell-1} U_{\ell-1}^\top) h_{i,\ell}}{\|(I - U_{\ell-1} U_{\ell-1}^\top) h_{i,\ell}\|_2}$$

**Roadmap.**

- Section B.1 proves that the forward propagation $\|h_{i,\ell}\|_2$ neither vanishes nor explodes.
- Section B.2 gives a lower bound on the projected forward propagation $\|(I - U_\ell U_\ell^\top) h_{i,\ell+1}\|_2$.
- Section B.3 proves that for two data points $i, j \in [n]$, the projected forward propagation $(I - U_\ell U_\ell^\top) h_{i,\ell+1}$ and $(I - U_\ell U_\ell^\top) h_{j,\ell+1}$ are separable from either other.
- Section B.4 and Section B.5 prove that the consecutive intermediate layers, in terms of spectral norm, do not explode (for full and sparse vectors respectively).
- Section B.6 proves that the backward propagation does not explode.

## B.1 Forward Propagation

Our first result of this section is on showing upper and lower bounds on the forward propagation.

**Lemma B.3** (c.f. (4.1)). *With probability at least $1 - \exp(-\Omega(m/L^2))$ over the random initialization $W, A$ (see Definition 2.3), it satisfies*

$$\forall i \in [n], \quad \forall \ell \in \{0, 1, \ldots, L-1\} \quad : \quad (1 - 1/(4L))^\ell \leq \|h_{i,\ell+1}\|_2 \leq 2\ell + 4.$$
$$\|g_{i,\ell+1}\|_2 \leq 4\ell + 8.$$

We prove Lemma B.3 by induction on $\ell \in [L]$. We only prove the $h_{i,\ell+1}$ part and the $g_{i,\ell+1}$ part is completely analogous.

For the base case $\ell = 0$, we have $\|x_{i,1}\|_2 = 1$ and $h_{i,1} = \phi(Ax_{i,1})$. Using Corollary A.10 we have $1 - 1/4L \leq \|h_{i,1}\|_2 \leq 1 + 1/4L$ with probability at least $1 - \exp(-\Omega(-m/L^2))$. We proceed the proof for the case of $\ell \geq 1$, assuming that Lemma B.3 already holds for $0, 1, \ldots, \ell - 1$. We first show

**Claim B.4.** *With probability at least $1 - \exp(-\Omega(m/L^2))$ over $W$ and $A$,*
$$\forall i \in [n]\colon \|h_{i,\ell+1}\|_2 \leq (1 + 1/L)(\|h_{i,\ell}\|_2 + 1)$$

Note if Claim B.4 holds for all $\ell = 0, 1, \ldots, \ell - 1$, then we must have $\|h_{i,\ell}\|_2 \leq (1+1/L)^\ell \|h_{i,0}\|_2 + \sum_{i=1}^{\ell}(1+1/L)^i = (1+1/L)^\ell + \sum_{i=1}^{\ell}(1+1/L)^i \leq 2\ell + 4$.

*Proof of Claim B.4.* Recall the definition of $h_{i,\ell} \in \mathbb{R}^m$, we have
$$\|h_{i,\ell+1}\|_2 = \|\phi(W \cdot h_{i,\ell} + A x_{i,\ell+1})\|_2$$

We can rewrite vector $W h_{i,\ell} \in \mathbb{R}^m$ as follows
$$W h_{i,\ell} = W U_{\ell-1} U_{\ell-1}^\top h_{i,\ell} + W(I - U_{\ell-1} U_{\ell-1}^\top) h_{i,\ell}$$
$$= W U_{\ell-1} U_{\ell-1}^\top h_{i,\ell} + W \frac{(I - U_{\ell-1} U_{\ell-1}^\top) h_{i,\ell}}{\|(I - U_{\ell-1} U_{\ell-1}^\top) h_{i,\ell}\|_2} \cdot \|(I - U_{\ell-1} U_{\ell-1}^\top) h_{i,\ell}\|_2$$
$$= W U_{\ell-1} U_{\ell-1}^\top h_{i,\ell} + W v_{i,\ell} \cdot \|(I - U_{\ell-1} U_{\ell-1}^\top) h_{i,\ell}\|,$$

where the last step follows by definition of $v_{i,\ell}$ (See Definition B.2). Define $z_1 \in \mathbb{R}^{n(\ell-1)}, z_2 \in \mathbb{R}, z_3 \in \mathbb{R}^d$ as follows
$$z_1 = U_{\ell-1}^\top h_{i,\ell}, \quad z_2 = \|(I - U_{\ell-1} U_{\ell-1}^\top) h_{i,\ell}\|_2, \quad z_3 = x_{i,\ell+1}. \tag{B.1}$$

Then
$$\|z_1\|_2^2 + z_2^2 + \|z_3\|_2^2 = \|U_{\ell-1}^\top h_{i,\ell}\|_2^2 + \|(I - U_{\ell-1} U_{\ell-1}^\top) h_{i,\ell}\|_2^2 + \|x_{i,\ell+1}\|_2^2$$
$$= \|h_{i,\ell}\|_2^2 + \|x_{i,\ell+1}\|_2^2 \leq \|h_{i,\ell}\|_2^2 + 1 \;. \tag{B.2}$$

We can thus rewrite
$$W h_{i,\ell} + A x_{i,\ell+1} = W U_{\ell-1} z_1 + W v_{i,\ell} z_2 + A z_3$$
$$= \begin{bmatrix} M_1 & M_2 & M_3 \end{bmatrix} \cdot \begin{bmatrix} z_1 \\ z_2 \\ z_3 \end{bmatrix}$$
$$= M \cdot z,$$

where the last step follows by defining $M \in \mathbb{R}^{m \times (n(\ell-1)+1+d)}$ as follows,
$$M = \begin{bmatrix} M_1 & M_2 & M_3 \end{bmatrix} = \begin{bmatrix} W U_{\ell-1} & W v_{i,\ell} & A \end{bmatrix} \tag{B.3}$$

We stress here that the entries of $M_1, M_2, M_3$ are i.i.d. from $\mathcal{N}(0, \frac{2}{m})$.[13] We have $\|h_{i,\ell+1}\| = \|\phi(W h_{i,\ell} + A x_{i,\ell+1})\| = \|\phi(M \cdot z)\|$.

Next, applying Corollary A.10, we know that if $z_1, z_2, z_3$ are fixed (instead of defined as in (B.1)), then, letting Choosing $\varepsilon = 1/2L$, we have
$$\Pr_M \left[ \|h_{i,\ell+1}\| \leq (1 + \varepsilon) \cdot \sqrt{\|z_1\|_2^2 + z_2^2 + \|z_3\|_2^2} \right] \geq 1 - \exp(-\Omega(\varepsilon^2 m)) \;.$$

---
[13] Indeed, after Gram-Schmidt we can write $U_{\ell-1} = [\widehat{h}_1, \ldots, \widehat{h}_{n(\ell-1)}]$, where each $\widehat{h}_j$ only depends on the randomness of $A$ and $W[\widehat{h}_1, \ldots, \widehat{h}_{j-1}]$. In other words, conditioning on any choice of $A$ and $W[\widehat{h}_1, \ldots, \widehat{h}_{j-1}]$, we still have $W\widehat{h}_j$ is an independent Gaussian vector from $\mathcal{N}(0, \frac{2}{m} I)$. Similarly, $v_{i,\ell}$ may depend on the randomness of $A$ and $W U_{\ell-1}$, but conditioning on any choice of $A$ and $W U_{\ell-1}$, we still have $W v_{i,\ell}$ follows from $\mathcal{N}(0, \frac{2}{m} I)$. This proves that the entries of $M_1, M_2, M_3$ are independent.

To move from fixed choices of $z_1$ and $z_2$ to *all* choices of $z_1$ and $z_2$, we perform a standard $\varepsilon$-net argument. Since the dimension of $z_1$ and $z_2$ are respectively $n(\ell-1)$ and 1, the size of $\varepsilon$-net for $z_1$ and $z_2$ is at most $e^{O(nL \log L)}$. Thus, with probability at least $1 - e^{O(nL \log L + \log n)} \cdot \exp(-\Omega(\varepsilon^2 m)) \geq 1 - \exp(-\Omega(\varepsilon^2 m))$ we have: for all $z_1 \in \mathbb{R}^{n(\ell-1)}$ and $z_2 \in \mathbb{R}$ satisfying $\|z_1\| \leq 2L+4$ and $0 \leq z_2 \leq 2L+4$, and fixed $z_3 = x_{i,\ell+1}$,

$$\|h_{i,\ell+1}\| \leq (1 + 2\varepsilon) \cdot \sqrt{\|z_1\|_2^2 + z_2^2 + \|z_3\|_2^2}.$$

In particular, since we have "for all" quantifies on $z_1$ and $z_2$ above, we can substitute the choice of $z_1$ and $z_2$ in (B.1) (which may depend on the randomness of $W$ and $A$). This, together with (B.2), gives

$$\mathbf{Pr}\left[\|h_{\ell+1}\|_2 \leq (1+\varepsilon) \cdot \sqrt{\|h_{i,\ell}\|_2^2 + 1}\right] \geq 1 - \exp(-\Omega(\varepsilon^2 m)) \ . \qquad \square$$

Similarly, we can prove a lower bound

**Claim B.5.** *With probability at least $1 - \exp(-\Omega(m/L^2))$ over $W$ and $A$,*

$$\|h_{i,\ell+1}\|_2 \geq (1 - \frac{1}{4L})\|h_{i,\ell}\|_2.$$

*Proof.* We can define $z_1, z_2, z_3$ in the same way as (B.1). This time, we show a lower bound

$$\|z_1\|_2^2 + z_2^2 + \|z_3\|_2^2 \geq \|z_1\|_2^2 + z_2^2 = \|h_{i,\ell}\|_2^2. \tag{B.4}$$

Applying Corollary A.10, we know if $z_1, z_2, z_3$ are fixed (instead of defined as in (B.1)), then choosing $\varepsilon = 1/8L$,

$$\mathbf{Pr}\left[\|h_{i,\ell+1}\| \geq (1-\varepsilon) \cdot \sqrt{\|z_1\|_2^2 + z_2^2 + \|z_3\|_2^2}\right] \geq 1 - \exp(-\Omega(\varepsilon^2 m)) \ ..$$

Again, after applying $\varepsilon$-net, we know with probability at least $1 - e^{O(nL \log L + \log n)} \cdot \exp(-\Omega(\varepsilon^2 m)) \geq 1 - \exp(-\Omega(\varepsilon^2 m))$, for all $z_1 \in \mathbb{R}^{n(\ell-1)}$ and $z_2 \in \mathbb{R}$ satisfying $\|z_1\| \leq 2L+4$ and $0 \leq z_2 \leq 2L+4$, and fixed $z_3 = x_{i,\ell+1}$,

$$\|h_{i,\ell+1}\|_2 \geq (1-\varepsilon) \cdot \sqrt{\|z_1\|_2^2 + z_2^2 + \|z_3\|_2^2}.$$

Substituting the choice of $z_1, z_2, z_3$ in (B.1), and the lower bound (B.4), we have

$$\mathbf{Pr}\left[\|h_{i,\ell+1}\| \geq (1-\varepsilon) \cdot \|h_{i,\ell}\|_2\right] \geq 1 - \exp(-\Omega(\varepsilon^2 m)).$$

Choosing $\varepsilon = 1/(4L)$ gives the desired statement. $\square$

Finally, recursively applying Claim B.4 and Claim B.5 for all $\ell = 0, 1, \ldots, L-1$, we have with probability at least $1 - L^2 \exp(-\Omega(m/L^2)) \geq 1 - \exp(-\Omega(m/L^2))$, we have

$$\|h_{i,\ell}\|_2 \geq (1 - \frac{1}{4L})^\ell, \quad \|h_{i,\ell}\|_2 \leq 2\ell + 4 \ .$$

This finishes the proof of Lemma B.3. ∎

## B.2 Forward Correlation

This subsection proves the following lemma which, as discussed in Section 4, bounds how much "fresh new randomness" is left after propagating to layer $\ell$.

**Lemma B.6** (c.f. (4.2)). *With probability at least $1 - e^{-\Omega(\sqrt{m})}$ over the random initialization $W, A$ in Definition 2.3, letting $U_\ell$ be defined in Definition B.1, we have*

$$\forall i \in [n], \forall \ell \in \{0, 1, \cdots, L-1\} \quad : \quad \|(I - U_\ell U_\ell^\top) h_{i,\ell+1}\|_2 \geq \frac{1}{2 \cdot 10^6 L^2 \log^3 m} \quad .$$

To prove Lemma B.6, we inductively (with the increasing order of $\ell$) show for each $i \in [n]$, for each $\ell \in \{0, 1, \cdots, L-1\}$, we have

$$\|(I - U_\ell U_\ell^\top) h_{i,\ell+1}\|_2 \geq \xi_\ell \stackrel{\text{def}}{=} \frac{1}{10^6 L^2 \log^3 m} (1 - 2\alpha - \frac{1}{4L})^\ell \quad \text{where} \quad \alpha \stackrel{\text{def}}{=} \frac{1}{2 \cdot 10^4 L^2 \log^2 m} \quad . \quad (B.5)$$

We first show (B.5) in the base case $\ell = 0$. Since $U_0$ is an empty matrix, we have $(I - U_0 U_0^\top) h_{i,1} = h_{i,1}$ so according to Lemma B.3 we have $\|h_{i,1}\|_2 \geq (1 - 1/4L)$.

The remainder of the proof assumes (B.5) already holds for $\ell - 1$. We can write

$$h_{i,\ell+1} = \phi(W h_{i,\ell} + A x_{i,\ell+1}) = \phi(M_1 z_1 + M_2 z_2 + M_3 z_3).$$

where $M_1 \in \mathbb{R}^{m \times n(\ell-1)}$, $M_2 \in \mathbb{R}^{m \times 1}$, $M_3 \in \mathbb{R}^{m \times d}$ and $z_1 \in \mathbb{R}^{n(\ell-1)}$, $z_2 \in \mathbb{R}$, $z_3 \in \mathbb{R}^d$ are defined as

$$\begin{aligned} M_1 &= W U_{\ell-1} & M_2 &= W v_{i,\ell} & M_3 &= A \\ z_1 &= U_{\ell-1}^\top h_{i,\ell} & z_2 &= \|(I - U_{\ell-1} U_{\ell-1}^\top) h_{i,\ell}\|_2 & z_3 &= x_{i,\ell+1} \quad . \end{aligned} \quad (B.6)$$

in the same way as (B.1) and (B.3) as in the proof of Lemma B.3. We again have the entries of $M_1, M_2, M_3$ are i.i.d. from $\mathcal{N}(0, \frac{2}{m})$ (recall Footnote 13). For the quantity $M_2 z_2$, we can further decompose it as follows

$$M z_2 = \nu \cdot (z_2 - c_5 \alpha)_+ + \nu' z_2' \quad (B.7)$$

where $c_5 = \frac{1}{16 \log m}$ is some fixed parameter, $\nu$ and $\nu'$ denote two vectors that are independently generated from $\mathcal{N}(0, \frac{2I}{m})$, and

$$(z_2 - c_5 \alpha)_+ = \begin{cases} 0, & \text{if } z_2 < c_5 \alpha; \\ \sqrt{z_2^2 - c_5^2 \alpha^2}, & \text{if } z_2 \geq c_5 \alpha. \end{cases} \qquad z_2' = \begin{cases} z_2, & \text{if } z_2 < c_5 \alpha; \\ c_5 \alpha, & \text{if } z_2 \geq c_5 \alpha. \end{cases}$$

It is clear that the two sides of (B.7) are identical in distribution (because $M_2 \sim \mathcal{N}(0, \frac{2I}{m})$ and $(z_2 - c_5\alpha)_+^2 + (z_2')^2 = z_2^2$).

Next, suppose $z_1$ and $z_2$ are fixed (instead of depending on the randomness of $W$ and $A$) and satisfies[14]

$$\frac{1}{\sqrt{2}} \leq \|z_1\| \leq 2L + 6 \text{ and } |z_2| \leq 2L + 6. \quad (B.8)$$

We can apply Corollary A.16 to obtain the following statement: with probability at least $1 - \exp(-\Omega(\alpha^2 m))$,

$$w \stackrel{\text{def}}{=} M_1 z_1 + \nu(z_2 - c_5 \alpha)_+ + M_3 z_3$$

is $(\alpha, \sigma/4)$-good where $\sigma = \left(\frac{2}{m}\|z_1\|_2^2 + \frac{2}{m}((z_2 - c_5\alpha)_+)^2 + \frac{2}{m}\|z_3\|_2^2\right)^{1/2}$. Using $\|z_1\|_2^2 \geq 1/2$, we can lower bound $\sigma^2$ as

$$\sigma^2 \geq \frac{2}{m}(\|z_1\|_2^2 + z_2^2 - c_5^2 \alpha^2) \geq \frac{2}{m}(\frac{1}{2} - c_5^2 \alpha^2) \geq \frac{1}{2m} \quad .$$

In other words, $w$ is $(\alpha, \frac{1}{4\sqrt{2}\sqrt{m}})$-good. Next, applying standard $\varepsilon$-net argument, we have with probability at least $1 - \exp(-\Omega(\alpha^2 m))$, for all vectors $z_1 \in \mathbb{R}^{n(\ell-1)}$ and $z_2 \in \mathbb{R}$ satisfying (B.8), it satisfies $w$ is $(\alpha, \frac{1}{8\sqrt{m}})$-good. This allows us to plug in the random choice of $z_1$ and $z_2$ in (B.6).

---
[14] Note that if $z_1$ and $z_2$ are random, then they satisfy such constraints by Lemma B.3.

We next apply Lemma B.7 with
$$w = M_1 z_1 + \nu(z_2 - c_5\alpha)_+ + M_3 z_3, \quad r = z_2', \quad v = \nu, \quad U = U_\ell.$$

(We can do so because the randomness of $v$ is independent of the randomness of $U_\ell$ and $w$.)
Lemma B.7 tells us that, with probability at least $1 - \exp(-\Omega(\sqrt{m}))$ over the randomness of $v$,
$$\|(I - U_\ell U_\ell^\top) \cdot \phi(w + rv)\| \geq r(1 - 2\alpha) - \frac{\alpha^{1.5}}{4} \ .$$

By induction hypothesis, we know
$$r = z_2' \geq \min\{z_2, c_5\alpha\} \geq \min\{\xi_{\ell-1}, c_5\alpha\} \geq \xi_{\ell-1} \ ,$$

where the last step follows by $\xi_{\ell-1} \leq c_5\alpha$. Thus, we have
$$\|(I - U_\ell U_\ell^\top) \cdot \phi(w + rv)\| \geq r(1 - 2\alpha) - \frac{\alpha^{1.5}}{4} \geq \xi_{\ell-1}(1 - 2\alpha) - \frac{\alpha^{1.5}}{4} \geq \xi_{\ell-1}(1 - 2\alpha - \frac{1}{4L}) = \xi_\ell \ ,$$

where the last inequality uses $\alpha^{1.5} \leq \frac{\xi_{\ell-1}}{L}$.

■

### B.2.1 Tools

**Lemma B.7.** *Suppose $\alpha \in [\frac{1}{100L^4}, 1/2]$, $m \geq 4\widetilde{d}/\alpha$ and $r \in (0, \frac{\alpha}{16 \log m}]$. Suppose $w$ is a fixed vector that is $(\alpha, \frac{1}{8\sqrt{m}})$-good (see Definition A.14), and $U \in \mathbb{R}^{m \times \widetilde{d}}$ is a fixed column orthonormal matrix. Then, if $v \sim \mathcal{N}(0, \frac{2}{m}I)$, with probability at least $1 - \exp(-\Omega(\sqrt{m}))$, it satisfies*
$$\|(I - UU^\top) \cdot \phi(w + rv)\| \geq r(1 - 2\alpha) - \frac{\alpha^{1.5}}{4} \ .$$

*Proof.* We split $w \in \mathbb{R}^m$ into three pieces $w = (w_1, w_2, w_3) \in \mathbb{R}^m$ with disjoint support such that:

- $w_1$ corresponds to the coordinates that are $\geq \alpha/8\sqrt{m}$,
- $w_2$ is the remaining, which corresponds to the coordinates that are within $(-\alpha/8\sqrt{m}, \alpha/8\sqrt{m})$.
- $w_3$ corresponds to the coordinates that are $\leq -\alpha/8\sqrt{m}$, and

We write $v = (v_1, v_2, v_3)$ according to the same partition. We consider the following three cases.

- For each index $k$ in the first block, we have $(\phi(w_1 + rv_1))_k \neq (w_1 + rv_1)_k$ only if $(rv_1)_k \leq -\alpha/8\sqrt{m}$. However, if this happens, we have
$$0 \leq (\phi(w_1 + rv_1))_k - (w_1 + rv_1)_k \leq \max\{(rv_1)_k - \alpha/8\sqrt{m}, 0\} \ .$$

Applying Lemma A.5, we know with probability at least $1 - e^{-\Omega(\sqrt{m})}$,
$$\delta_1 \stackrel{\text{def}}{=} \phi(w_1 + rv_1) - (w_1 + rv_1) \quad \text{satisfies} \quad \|\delta_1\|^2 \leq 2\sqrt{m}\frac{2r^2}{m} \leq 4\alpha^2/\sqrt{m} \ .$$

- Similarly, for each index $k$ in the third block, we have $(\phi(w_1 + rv_1))_k \neq 0$ only if $(rv_1)_k \geq \alpha/8\sqrt{m}$. Therefore, we can similarly derive that
$$\delta_3 \stackrel{\text{def}}{=} \phi(w_3 + rv_3) \quad \text{satisfies} \quad \|\delta_3\|^2 \leq 4\alpha^2/\sqrt{m} \ .$$

- For the second block, we claim that
$$\delta_2 \stackrel{\text{def}}{=} \phi(w_2 + rv_2) \quad \text{satisfies} \quad \|\delta_2\|_2 \leq \alpha^{3/2}/4 \ . \tag{B.9}$$

To prove (B.9), we use triangle inequality,
$$\|\phi(w_2 + rv_2)\|_2 \leq \|\phi(w_2)\|_2 + \|\phi(rv_2)\|_2$$
Since $\|\phi(w_2)\|_\infty \leq \alpha/8\sqrt{m}$ and the size of support of $\phi(w_2)$ is at most $\alpha m$, we have
$$\|\phi(w_2)\|_2 \leq ((\alpha/8\sqrt{m})^2 \alpha m)^{1/2} \leq \frac{\alpha^{3/2}}{8}.$$
Since $v \sim \mathcal{N}(0, \frac{2}{m}I)$, and since the size of support of $v_2$ is at most $\alpha m$, we have $\|v_2\| \leq 2\sqrt{\alpha}$ with probability at least $1 - e^{-\Omega(\alpha m)}$ (due to chi-square distribution concentration). Thus
$$\|\phi(rv_2)\|_2 = r \cdot \|\phi(v_2)\|_2 \leq \frac{\alpha}{16} \cdot 2\sqrt{\alpha} = \frac{\alpha^{3/2}}{8} .$$
Together, by triangle inequality we have $\|\phi(w_2 + rv_2)\| \leq \frac{\alpha^{3/2}}{4}$ . This finishes the proof of (B.9).

Denoting by $\delta = (\delta_1, \delta_2, \delta_3)$, we have
$$\phi(w + rv) = w_1 + rv_1 + \delta.$$
Taking the norm on both sides,
$$\begin{aligned}
\|(I - UU^\top)\phi(w + rv)\| &= \|(I - UU^\top)(w_1 + rv_1 + \delta)\| \\
&\geq \|(I - UU^\top)(w_1 + rv_1)\| - \|(I - UU^\top)\delta\| \quad \text{(by triangle inequality)} \\
&\geq \|(I - UU^\top)(w_1 + rv_1)\| - \|\delta\| \quad \text{(by } \|(I - UU^\top)\delta\| \leq \|\delta\|) \\
&\geq \|(I - UU^\top)(w_1 + rv_1)\| - \frac{\alpha^{3/2}}{4} - \frac{4\alpha}{m^{1/4}} \\
&\geq \|(I - UU^\top)(w_1 + rv_1)\| - \frac{\alpha^{3/2}}{2} \\
&\geq \|(I - UU^\top)w_1 + rv_1\| - r\|U^\top v_1\| - \frac{\alpha^{3/2}}{2} .
\end{aligned}$$

Now, since $U \in \mathbb{R}^{m \times \widetilde{d}}$, it is not hard to show that $\|U^\top v_1\|_2^2 \leq \frac{2\widetilde{d}}{m}$ with probability at least $1 - \exp(-\Omega(m))$. Using $\frac{2\widetilde{d}}{m} \leq \frac{\alpha}{2}$ (owing to our assumption $m \geq 4\widetilde{d}/\alpha$), we have $r\|U^\top v_1\| \leq \frac{r\alpha}{2}$.

On the other hand, the random vector $z = (I - UU^\top)w_1 + rv_1$ follows from distribution $\mathcal{N}(\mu, \frac{2r^2}{m})$ for some fixed vector $\mu = (I - UU^\top)w_1$ and has at least $\frac{m}{2}(1 - \alpha)$ dimensions. By chi-square concentration, we have $\|z\| \geq r(1 - 3\alpha/2)$ with probability at least $1 - e^{-\Omega(\alpha^2 m)}$. Putting these together, we have

$$\|(I - UU^\top)\phi(w + rv)\| \geq r(1 - 3\alpha/2) - \frac{r\alpha}{2} - \frac{\alpha^{3/2}}{2} = r(1 - 2\alpha) - \frac{\alpha^{3/2}}{2} .$$

$\square$

### B.3 Forward $\delta$-Separateness

We first give the definition of $\delta$-Separable,

**Definition B.8** ($\delta$-separable vectors). *For any two vectors $x, y$, we say $x$ and $y$ are $\delta$-separable if*
$$\left\|(I - \frac{yy^\top}{\|y\|_2^2})x\right\| \geq \delta \quad \text{and} \quad \left\|(I - \frac{xx^\top}{\|x\|_2^2})y\right\| \geq \delta ..$$
*We say a finite set $X$ is $\delta$-separable if for any two vectors $x, y \in X$, $x$ and $y$ are $\delta$-separable.*

The goal of this subsection is to prove the $\delta$-separateness over all layers $\ell$.

**Lemma B.9** (c.f. (4.3)). *Let $\{x_{i,1}\}_{i\in[n]}$ be $\delta$-separable with $\delta \leq \frac{1}{10^6 L^2 \log^3 m}$. With probability at least $1 - e^{-\Omega(\sqrt{m})}$, for all $\ell = 0, 1, \ldots, L-1$, for all $i, j \in [n]$ with $i \neq j$, we have*

$$(I - U_\ell U_\ell^\top)h_{i,\ell+1} \text{ and } (I - U_\ell U_\ell^\top)h_{j,\ell+1} \text{ are } \frac{\delta}{2}\text{-separable.}$$

(Note that since we have assumed $\|x_{i,1} - x_{j,1}\| \geq \delta$ in Assumption 2.1 and assumed without loss of generality that $(x_{i,1})_{d_x} = \frac{1}{\sqrt{2}}$, it automatically satisfies that $\{x_{i,1}\}_{i\in[n]}$ is $O(\delta)$-separable.)

To prove Lemma B.9, we inductively (with the increasing order of $\ell$) show for each $i \in [n]$, for each $\ell \in \{0, 1, \cdots, L-1\}$, we have

$$(I - U_\ell U_\ell^\top)h_{i,\ell+1} \text{ and } (I - U_\ell U_\ell^\top)h_{j,\ell+1} \text{ are } \delta_\ell \text{ separable for } \delta_\ell \stackrel{\text{def}}{=} \delta(1 - 2\alpha - \frac{1}{4L})^\ell \quad (\text{B.10})$$

where $\alpha \stackrel{\text{def}}{=} 16\delta \log m$.

We skip the base case and only prove (B.10) for $\ell \geq 1$ by assuming (B.10) already holds for $\ell - 1$.[15]

**Claim B.10.** *For $i \neq j$, if $(I - U_{\ell-1}U_{\ell-1}^\top)h_{i,\ell}$ and $(I - U_{\ell-1}U_{\ell-1}^\top)h_{j,\ell}$ are $\delta_{\ell-1}$-separable, then letting $U = [U_\ell, \widehat{h}]$ where $\widehat{h} = \frac{(I - U_\ell U_\ell^\top)h_{j,\ell+1}}{\|(I - U_\ell U_\ell^\top)h_{j,\ell+1}\|_2}$, we have*

$$\|(I - UU^\top)h_{i,\ell+1}\| \geq \delta_\ell = \delta_{\ell-1}(1 - 3\alpha) .$$

*holds with probability at least $1 - \exp(-\Omega(\sqrt{m}))$.*

Since it is easy to verify that

$$\left\|\left(I - \widehat{h}\widehat{h}^\top\right)(I - U_\ell U_\ell^\top)h_{i,\ell+1}\right\| = \|(I - UU^\top)h_{i,\ell+1}\|_2 ,$$

Claim B.10 immediately implies (B.10) for layer $\ell$, and thus finishes the proof of Lemma B.9. Therefore, we only need to prove Claim B.10 below.

*Proof of Claim B.10.* Let $x = (I - U_{\ell-1}U_{\ell-1}^\top)h_{j,\ell}$ and $y = (I - U_{\ell-1}U_{\ell-1}^\top)h_{i,\ell}$. If $x$ and $y$ are $\delta_{\ell-1}$-separable, we split $y$ into two parts where $y_1$ is parallel to $x$ and $y_2$ is orthogonal to $x$

$$y = y_1 + y_2, \quad y_1 = \frac{\langle x, y \rangle x}{\|x\|_2^2}, \quad y_2 = (I - xx^\top/\|x\|_2^2)y$$

This also implies that the randomness in $y_1$ is independent of the randomness in $y_2$.

It is easy to see that $\|y_1\|_2 = \frac{\langle x, y \rangle}{\|x\|_2}$, so we can rewrite $Wy$ as follows

$$Wy = Wy_1 + Wy_2 = W\frac{\langle x, y \rangle x}{\|x\|_2^2} + Wy_2 = (\|y_1\|_2/\|x\|_2)Wx + Wy_2.$$

Therefore,

$$\begin{aligned}
Wh_{i,\ell} + Ax_{i,\ell+1} &= WU_{\ell-1}U_{\ell-1}^\top h_{i,\ell} + W(I - U_{\ell-1}U_{\ell-1}^\top)h_{i,\ell} + Ax_{i,\ell+1} \\
&= WU_{\ell-1}U_{\ell-1}^\top h_{i,\ell} + Wy + Ax_{i,\ell+1} &\text{(by definition of } y\text{)} \\
&= WU_{\ell-1}U_{\ell-1}^\top h_{i,\ell} + (\|y_1\|_2/\|x\|_2)Wx + Ax_{i,\ell+1} + Wy_2 &\text{(by rewritting } Wy\text{)} \\
&= M_1z_1 + M_2z_2 + M_3z_3 + M_4z_4
\end{aligned}$$

---

[15]The proof of the base case is a replication of Claim B.10 and only simpler. Indeed, for the base case of $\ell = 0$, one can view $h_{i,\ell} = x_{i,1}$ and $h_{j,\ell} = x_{j,1}$, and view $U_\ell$ as an empty matrix. Then, the same analysis of Claim B.10 but with slightly different notations will apply.

where

$$M_1 = WU_{\ell-1} \qquad M_2 = W\frac{x}{\|x\|_2} \qquad M_3 = A \qquad M_4 = W\frac{y_2}{\|y_2\|_2}$$

$$z_1 = U_{\ell-1}^\top h_{i,\ell} \qquad z_2 = \|y_1\|_2 \qquad z_3 = x_{i,\ell+1} \qquad z_4 = \|y_2\|_2 \ , \tag{B.11}$$

and we know the entries of $M_1, M_2, M_3, M_4$ are i.i.d. from $\mathcal{N}(0, \frac{2}{m})$, owing to a similar treatment as Footnote 13. For the vector $M_4 z_4$, we further rewrite it as

$$M_4 z_4 = \nu \cdot (z_4 - c_5 \alpha)_+ + \nu' z_4'$$

where $c_5 = \frac{1}{16 \log m}$ is a fixed parameter, and $\nu$ and $\nu'$ denote two vectors that are independently generated from the same distribution $\mathcal{N}(0, \frac{2}{m}I)$ as vector $M_4 \in \mathbb{R}^m$, and

$$(z_4 - c_5\alpha)_+ = \begin{cases} 0, & \text{if } z_4 < c_5\alpha; \\ \sqrt{z_4^2 - c_5^2\alpha^2}, & \text{if } z_4 \geq c_5\alpha. \end{cases} \qquad z_4' = \begin{cases} z_4, & \text{if } z_4 < c_5\alpha; \\ c_5\alpha, & \text{if } z_4 \geq c_5\alpha. \end{cases}$$

Together, we can write

$$\begin{aligned}(I - UU^\top)h_{i,\ell+1} &= (I - UU^\top)\phi(Wh_{i,\ell} + Ax_{i,\ell+1}) \\ &= (I - UU^\top)\phi(M_1 z_1 + M_2 z_2 + M_3 z_3 + M_4 z_4) \\ &= (I - UU^\top)\phi(M_1 z_1 + M_2 z_2 + M_3 z_3 + \nu \cdot (z_4 - c_5\alpha)_+ + \nu' z_4') \\ &= (I - UU^\top)\phi(w + \nu' z_4')\end{aligned}$$

where the last step follows by defining

$$w = M_1 z_1 + M_2 z_2 + M_3 z_3 + \nu \cdot (z_4 - c_5\alpha)_+ \ .$$

Our plan is to first use the randomness in $w$ to argue that $w$ is $(\alpha, \gamma/4\sqrt{m})$-good. Then we conditioned $w$ is good, and prove that the norm of $(I - UU^\top)\phi(w + \nu' z_4')$ is lower bounded (using (B.7)).

Now, suppose that $z_1 \in \mathbb{R}^{n(\ell-1)}$, $z_2 \in \mathbb{R}$, and $z_4 \in \mathbb{R}$ are fixed (instead of computed based on the randomness of $W$ and $A$), and satisfies[16]

$$\frac{1}{\sqrt{2}} \leq \|z_1\| \leq 2L + 6 \text{ and } |z_2|, |z_4| \leq 2L + 6. \tag{B.12}$$

we can use Corollary A.16 to obtain the following statement: $w$ is $(\alpha, \sigma/4)$-good with probability at least $1 - \exp(-\Omega(\alpha^2 m))$ where $\sigma = \left(\frac{2}{m}\|z_1\|_2^2 + \frac{2}{m}z_2^2 + \frac{2}{m}\|z_3\|_2^2 + \frac{2}{m}(z_4^2 - c_5^2\alpha^2)\right)^{1/2}$. Using $\|z_1\|_2^2 \geq 1/2$, we can lower bound $\sigma^2$ as

$$\sigma^2 \geq \frac{2}{m}(\|z_1\|_2^2 + z_2^2 + z_4^2 - c_5^2\alpha^2) \geq \frac{2}{m}(\frac{1}{2} - c_5^2\alpha^2)$$

In other words, $w$ is $(\alpha, \frac{1}{4\sqrt{2}\sqrt{m}})$-good. Next, applying standard $\varepsilon$-net argument, we have with probability at least $1 - \exp(-\Omega(\alpha^2 m))$, for all $z_1, z_2, z_4$ satisfying (B.12), it satisfies $w$ is $(\alpha, \frac{1}{8\sqrt{m}})$-good. This allows us to plug in the random choice of $z_1, z_2, z_4$ in (B.11).

We apply Lemma B.7 with the following setting

$$U = \left[U_\ell, \frac{(I - U_\ell U_\ell^\top)h_{j,\ell+1}}{\|(I - U_\ell U_\ell^\top)h_{j,\ell+1}\|_2}\right], \quad w = w, \quad v = \nu', \quad r = z_4' \in [0, c_5\alpha]$$

where $w$ is $(\alpha, \gamma/8\sqrt{m})$-good. (We can do so because the randomness of $v$ is independent of the randomness of $U$ and $w$.) Lemma B.7 tells us that, with probability at least $1 - \exp(-\Omega(\sqrt{m}))$ over the randomness of $v$,
$$\|(I - UU^\top) \cdot \phi(w + rv)\| \geq r(1 - 2\alpha) - \frac{\alpha^{1.5}}{4} \ .$$

---

[16]Note that if $z_1, z_2$ and $z_4$ are random, then they satisfy such constraints by Lemma B.3.

By induction hypothesis, we know
$$r = z'_4 \geq \min\{z_4, c_5\alpha\} \geq \min\{\delta_{\ell-1}, c_5\alpha\} \geq \delta_{\ell-1} \ ,$$

where the last step follows by $\delta_{\ell-1} \leq c_5\alpha$. Thus, we have
$$\|(I - UU^\top) \cdot \phi(w + rv)\| \geq r(1 - 2\alpha) - \frac{\alpha^{1.5}}{4} \geq \delta_{\ell-1}(1 - 2\alpha) - \frac{\alpha^{1.5}}{4} \geq \delta_{\ell-1}(1 - 2\alpha - \frac{1}{4L}) = \delta_\ell \ ,$$

where the last inequality uses $\alpha^{1.5} \leq \frac{\delta_{\ell-1}}{L}$. $\square$

## B.4 Intermediate Layers: Spectral Norm

The following lemma bounds the spectral norm of (consecutive) intermediate layers.

> **Lemma B.11** (c.f. (4.4)). *With probability at least $1 - \exp(-\Omega(m/L^2))$, we have for all $L \geq \ell_2 \geq \ell_1 \geq 0$ and $i \in [n]$*
> $$\left\| \prod_{\ell=\ell_2}^{\ell_1} D_{i,\ell} W \right\|_2 \leq O(L^3),$$

We start with an important claim whose proof is almost identical to Lemma B.3.

**Claim B.12.** *Given $\ell > b \geq 1$, given $i \in [n]$, and given $z_{b-1} \in \mathbb{R}^m$ a fixed vector, letting*
$$z_\ell = D_{i,\ell} W D_{i,\ell-1} \cdots D_{i,b} W z_{b-1} \ ,$$
*we have with probability at least $1 - \exp(-\Omega(m/L^2))$, it satisfies $\|z_\ell\|_2 \leq (1 + 1/L)^{\ell-b+1} \|z_{b-1}\|_2$.*

*Proof.* Let $U$ denote the following column orthonormal matrix using Gram-Schmidt (its first $n(\ell-1)$ columns coincide with $U_{\ell-1}$):
$$U \stackrel{\text{def}}{=} \mathsf{GS}\left(h_{1,1}, \ldots, h_{n,1}, h_{1,2}, \ldots, h_{n,2}, \ldots, h_{1,\ell-1}, \ldots, h_{n,\ell-1}, z_{b-1}, \cdots z_{\ell-1}\right) \ .$$

We can rewrite $\|z_\ell\|_2$ as follows:
$$\begin{aligned}
\|z_\ell\|_2 &= \|D_{i,\ell} W z_{\ell-1}\|_2 \\
&= \left\|\mathbf{1}_{Wh_{i,\ell-1}+Ax_{i,\ell} \geq 0} \cdot W z_{\ell-1}\right\|_2 \\
&= \left\|\mathbf{1}_{WUU^\top h_{i,\ell-1}+Ax_{i,\ell} \geq 0} \cdot WUU^\top z_{\ell-1}\right\|_2 \\
&= \left\|\mathbf{1}_{My+Ax \geq 0} M z\right\|_2 \ .
\end{aligned}$$

where in the last step we have defined $M = WU$, $y = U^\top h_{i,\ell-1}$, $x = x_{i,\ell}$ and $z = U^\top z_{\ell-1}$. We stress here that the entries of $M$ and $A$ are i.i.d. from $\mathcal{N}(0, \frac{2}{m})$.[17]

Now, Claim B.13 tells us for fixed $x \in \mathbb{R}^d$ and fixed $y, z \in \mathbb{R}^{n\ell+\ell-b}$, we have with probability $1 - \exp(-\Omega(m/L^2))$ (over the randomness of $M, A$),
$$\left\|\mathbf{1}_{My+Ax \geq 0} M z\right\|_2 \leq \|z\|_2 (1 + 1/2L) \ .$$

After taking $\varepsilon$-net over all possible $y, z$, we have that for fixed $x \in \mathbb{R}^d$ but all $y, z \in \mathbb{R}^{n\ell+\ell-b}$:
$$\left\|\mathbf{1}_{My+Ax \geq 0} M z\right\|_2 \leq \|z\|_2 (1 + 1/L) \ .$$

---
[17]This follows from a similar argument as Footnote 13, taking into account the additional fact that each $z_j$ may only depend on the randomness of $A$, $WU_{\ell-1}$, and $Wz_{b-1}, \ldots Wz_{j-1}$.

We can thus plug in the choice $y = U^\top h_{i,\ell-1}$ and $z = U^\top z_{\ell-1}$ (both of which may depend on the randomness of $W$ and $A$). Using $\|z\|_2 \leq \|z_{\ell-1}\|_2$, we have

$$\|z_\ell\|_2 \leq \|z_{\ell-1}\|_2 (1 + 1/L) \ .$$

Finally, taking union bound over all possible $\ell$ and applying induction, we have

$$\|z_\ell\|_2 \leq (1 + 1/L)^{\ell-b+1} \|z_{b-1}\|_2. \qquad \square$$

Now, to prove the spectral norm bound in Lemma B.11, we need to go from "for each $z_{b-1}$ (see Claim B.12)" to "for all $z_{b-1}$". Since $z_{b-1}$ has $m$ dimensions, we cannot afford taking union bound over all possible $z_{b-1}$ (or its $\varepsilon$-net).

To bypass this issue, we partition the coordinates of $z_{b-1}$ into $L^3$ trunk (each of length $m/L^3$). We write $z_{b-1} = \sum_{j=1}^{L^3} (z_{b-1})_j$ where each $(z_{b-1})_j \in \mathbb{R}^m$ denotes a vector that only has non-zero entries on $m/L^3$ coordinates. For each $j \in [L^3]$, we have with probability $1 - \exp(-m/L^2)$,

$$\|(z_\ell)_j\|_2 \leq (1 + 1/L)^{\ell-b} \cdot \|(z_{b-1})_j\|_2 \leq 2\|(z_{b-1})_j\|_2 \ .$$

By applying an $\varepsilon$-net argument over all such possible (but sparse) $(z_{b-1})_j$, we have with probability at least $1 - 2^{O(m/L^3)} \exp(-\Omega(m/L^2)) \geq 1 - \exp(-\Omega(m/L^2))$, the above equation holds for all possible $(z_{b-1})_j$.

Next, taking a union bound over all $j \in [L^3]$, we have with probability at least $1 - \exp(-\Omega(m/L^2))$:

$$\forall z_{b-1} \in \mathbb{R}^m : \|z_\ell\|_2 \leq O(L^3) \|z_{b-1}\|_2 \ .$$

Taking a union bound over all $\ell, b$, we complete the proof of Lemma B.11. $\blacksquare$

### B.4.1 Tools

**Claim B.13.** *For fixed $x \in \mathbb{R}^d, y, z \in \mathbb{R}^k$. Let $M \in \mathbb{R}^{m \times k}$ and $A \in \mathbb{R}^{m \times d}$ denote random Gaussian matrices where each entry is i.i.d. sampled from $\mathcal{N}(0, 2/m)$. We have with probability at least $1 - \exp(-\Omega(m/L^2))$*

$$\|\mathbf{1}_{My+Ax \geq 0} \cdot Mz\|_2 \leq \|z\|_2 (1 + 1/2L) \ .$$

*Proof.* Without loss of generality we assume $\|z\|_2 = 1$. We can rewrite $My$ as follows

$$My = M(zz^\top)y + M(I - zz^\top)y$$
$$= Mz \cdot z^\top y + \frac{M(I - zz^\top)y}{\|(I - zz^\top)y\|_2} \cdot \|(I - zz^\top)y\|_2$$
$$= M_1 z_1 + M_2 z_2,$$

where $M_1, M_2 \in \mathbb{R}^m$ and $z_1, z_2 \in \mathbb{R}$ are defined as follows

$$M_1 = Mz, \quad M_2 = \frac{M(I - zz^\top)y}{\|(I - zz^\top)y\|_2}$$
$$z_1 = z^\top y, \quad z_2 = \|(I - zz^\top)y\|_2$$

It is easy to see that $M_1$ is independent of $M_2$. We can rewrite

$$\|\mathbf{1}_{My+Ax \geq 0} \cdot Mz\|_2 = \|\mathbf{1}_{M_1 z_1 + M_2 z_2 + Ax \geq 0} \cdot M_1\|_2$$

Using Fact A.11 together with concentration bounds (for binomial distribution and for chi-square distribution), we have with probability at least $1 - \exp(-\Omega(m/L^2))$,

$$\|\mathbf{1}_{M_1 z_1 + M_2 z_2 + Ax \geq 0} \cdot M_1\|_2 \leq \|z\|_2 (1 + 1/2L).$$

Thus, we complete the proof. $\square$

## B.5 Intermediate Layers: Sparse Spectral Norm

This section proves two results corresponding to the spectral norm of intermediate layers with respect to *sparse* vectors. We first show Lemma B.14 and our Corollary B.15 and B.16 shall be direct applications of Lemma B.14.

**Lemma B.14.** *For every $k \in [m]$ and $t \geq 2$, with probability at least*
$$1 - nL^2 e^{O(k \log(m))} (e^{-\Omega(m/L^2)} + e^{-\Omega(nLt^2)})$$
*it satisfies, for all $i \in [n]$, for all $L > \ell_2 \geq \ell_1 \geq 1$, for all $k$-sparse vectors $z, y \in \mathbb{R}^m$*
$$\left| y^\top W \left( \prod_{\ell=\ell_2}^{\ell_1} D_{i,\ell} W \right) z \right| \leq \frac{5t\sqrt{nL}}{m^{1/2}} \cdot \|y\|_2 \cdot \|z\|_2.$$

*Proof.* Without loss of generality we assume $\|y\| = \|z\| = 1$. Fixing $\ell_2 \geq \ell_1$, fixing $i$, and fixing $z_{\ell_1 - 1} = z$, we have according to Claim B.12, letting $z_{\ell_2} = D_{i,\ell_2} W D_{i,\ell_2 - 1} W \cdots D_{i,\ell_1} z_{\ell_1 - 1}$, then with probability at least $1 - e^{-\Omega(m/L^2)}$
$$\|z_{\ell_2}\|_2 \leq (1 + 1/L)^{\ell_2 - \ell_1 - 1} \|z_{\ell_1 - 1}\|_2 \leq 2 \ .$$

Applying $\varepsilon$-net over all $k$-sparse vectors $z$, we have with probability at least $1 - e^{O(k \log m)} e^{-\Omega(m/L^2)}$, it satisfies $\|z_{\ell_2}\|_2 \leq 3$ for all $k$-sparse $z$:

Similar to the proof of Lemma B.11, we let $U$ denote the following column orthonormal matrix using Gram-Schmidt:
$$U \stackrel{\text{def}}{=} \mathsf{GS}\left(h_{1,1}, \ldots, h_{n,1}, h_{1,2}, \ldots, h_{n,2}, \ldots, h_{1,\ell_2 - 1}, \ldots, h_{n,\ell_2 - 1}, z_{\ell_1 - 1}, \cdots z_{\ell_2}\right) \ ,$$
and we have
$$\left| y^\top W \left( \prod_{\ell=\ell_2}^{\ell_1} D_{i,\ell} W \right) z \right| = \left| y^\top W U U^\top z_{\ell_2} \right| \leq \left\| y^\top W U \right\| \cdot \| U^\top z_{\ell_2} \| \leq 3 \left\| y^\top W U \right\| \ .$$

Next, observe the entries of $WU \in \mathbb{R}^{m \times n(\ell_2 - 1) + (\ell_2 - \ell_1 + 2)}$ are i.i.d. drawn from $\mathcal{N}(0, \frac{2}{m})$ (following a similar argument as Footnote 17). Therefore, if $y$ is a fixed vector, then $y^\top WU$ is in distribution identical to a Gaussian vector $\mathcal{N}(0, \frac{2}{m} I)$ of $n(\ell_2 - 1) + (\ell_2 - \ell_1 + 2) \leq 2nL$ dimensions. By chi-square distribution tail bound (see Lemma A.3), we have for $t \geq 2$:
$$\mathbf{Pr}[\|y^\top WU\|^2 \geq \frac{nLt^2}{m}] \leq e^{-\Omega(nLt^2)} \ .$$

Applying $\varepsilon$-net over all $k$-sparse vectors $y$, we have with probability at least $1 - e^{O(k \log m)} e^{-\Omega(nLt^2)}$, it satisfies $\|y^\top WU\|^2 \leq \frac{2nLt^2}{m}$ for all $k$-sparse vectors $y$.

Conditioning on both events happen, we have
$$\left| y^\top W \left( \prod_{\ell=\ell_2}^{\ell_1} D_{i,\ell} W \right) z \right| \leq 3 \frac{\sqrt{2nL} t}{\sqrt{m}} \ .$$

Taking union bound over all possible $i, \ell_1, \ell_2$ we finish the proof. □

Choosing $k = s^2 m^{2/3}$ and $t = \frac{sm^{1/3} \log m}{5\sqrt{nL}}$ in Lemma B.14, we have **Corollary B.15** (c.f. (4.5)). *Let $s \in [m^{-1/4}, m^{1/6}]$ be a fixed real. With probability at least $1 - e^{-\Omega(s^2 m^{2/3} \log^2 m)}$, we have for all $i \in [n]$, for all $L > \ell_2 \geq \ell_1 \geq 1$, for all $s^2 m^{2/3}$-sparse vectors $y, z \in \mathbb{R}^m$,*

$$\left| z^\top W \left( \prod_{\ell=\ell_2}^{\ell_1} D_{i,\ell} W \right) y \right| \leq \frac{s \log m}{m^{1/6}} \cdot \|z\|_2 \cdot \|y\|_2.$$

Choosing $k = 1$ and $t = \frac{\sqrt{nLd \log m}}{5}$ in Lemma B.14, we have

**Corollary B.16** (c.f. (4.5)). *Let $\rho = nLd \log m$. With probability at least $1 - \exp(-\Omega(\rho^2))$, it satisfies for all $i \in [n]$, for all $L > \ell_2 \geq \ell_1 \geq 1$, and for all 1-sparse vectors $y, z \in \mathbb{R}^m$,*

$$\left| z^\top W \left( \prod_{\ell=\ell_2}^{\ell_1} D_{i,\ell} W \right) y \right| \leq \frac{\rho}{m^{1/2}} \cdot \|z\|_2 \cdot \|y\|_2.$$

### B.6 Backward Propagation

This section proves upper bound on the backward propagation against *sparse* vectors. We first show Lemma B.17 and our Corollary B.18 and B.19 shall be immediate corollaries.

**Lemma B.17.** *For any $k \in [m]$, $t \geq 2$ and any $a \in \mathbb{R}^d$, with probability at least*

$$1 - nL^2 e^{O(k \log m)} (e^{-\Omega(m/L^2)} + e^{-\Omega(t^2)}),$$

*over the randomness of $W, A, B$, we have for all $i \in [n]$, for all $L \geq \ell_2 \geq \ell_1 \geq 1$, and for all $k$-sparse $y \in \mathbb{R}^m$,*

$$\left| a^\top B \left( \prod_{\ell=\ell_2}^{\ell_1} D_{i,\ell} W \right) y \right| \leq \frac{t}{\sqrt{d}} \cdot \|a\|_2 \cdot \|y\|_2.$$

*Proof of Lemma B.17.* Using Claim B.12, we know that for fixed $z_{\ell_1 - 1} = y$, letting

$$z_{\ell_2} = D_{i,\ell_2} W D_{i,\ell_2 - 1} W \cdots D_{i,\ell_1} W z_{\ell_1 - 1},$$

with probability at least $1 - \exp(-\Omega(m/L^2))$ we have

$$\|z_{\ell_2}\|_2 \leq (1 + 1/L)^{\ell_2 - \ell_1 + 1} \|z_{\ell_1 - 1}\|_2 \leq 3\|z_{\ell_1 - 1}\|_2.$$

Next, fixing vectors $a$ and $z_{\ell_2}$ and letting $B$ be the only source of randomness, we know $a^\top B z_{\ell_2}$ follows from $\mathcal{N}(0, \|a\|_2^2 \|z_{\ell_2}\|_2^2 / d)$. Then with probability at least $1 - \exp(-\Omega(t^2))$ over $B$,

$$|a^\top B z_{\ell_2}| \leq t \cdot \|a\|_2 \|z_{\ell_2}\|_2 / \sqrt{d}.$$

Taking a union of the above two events, we get with probability $1 - \exp(-\Omega(m/L^2)) - \exp(-\Omega(t^2))$,

$$|a^\top B z_{\ell_2}| \leq \frac{3t}{\sqrt{d}} \cdot \|a\|_2 \cdot \|y\|_2.$$

At this point, we apply $\varepsilon$-net argument for all $k$-sparse vectors $y \in \mathbb{R}^m$ (the size of which is at most $e^{O(k \log(m/k))}$). Taking a union bound over all such vectors in the $\varepsilon$-net, we have with probability at least

$$1 - \exp(O(d + k \log(m/k)))(\exp(-m/L^2) + \exp(-\Omega(t^2))),$$

for all $k$-sparse vector $y \in \mathbb{R}^m$,
$$|a^\top B z_{\ell_2}| \leq \frac{3t}{\sqrt{d}} \cdot \|a\|_2 \cdot \|y\|_2 \ .$$

Finally, we also take a union bound over all $\ell_2, \ell_1$ and $i$, and there are at most $O(nL^2)$ choices. $\square$

Using Lemma B.17 with $k = s^2 m^{2/3}$ and $t = sm^{1/3} \log m$, and taking union bound over all $a \in \mathbb{R}^d$, give

**Corollary B.18** (c.f. (4.6)). *Let $s \in [m^{-1/4}, m^{1/6}]$ be a fixed real. With probability at least $1 - \exp(-\Omega(s^2 m^{2/3} \log^2 m))$, we have for all $i \in [n]$, for all $L \geq \ell_2 \geq \ell_1 \geq 1$, for all $(s^2 \cdot m^{2/3})$-sparse $y \in \mathbb{R}^m$, for all $a \in \mathbb{R}^d$,*
$$\left| a^\top B \left( \prod_{\ell=\ell_2}^{\ell_1} D_{i,\ell} W \right) y \right| \leq (sm^{1/3} \log m) \cdot \|a\|_2 \cdot \|y\|_2 \ .$$

Using Lemma B.17 with $k = 1$ and $t = nLd \log m$, and taking union bound over all $a \in \mathbb{R}^d$, give

**Corollary B.19** (c.f. (4.6)). *Let $\rho = nLd \log m$. With probability at least $1 - \exp(-\Omega(\rho^2))$, we have for all $i \in [n]$, for all $L \geq \ell_2 \geq \ell_1 \geq 1$, for all 1-sparse vector $y \in \mathbb{R}^m$, and for all $a \in \mathbb{R}^d$,*
$$\left| a^\top B \left( \prod_{\ell=\ell_2}^{\ell_1} D_{i,\ell} W \right) y \right| \leq \rho \cdot \|a\|_2 \cdot \|y\|_2.$$

## C  Stability After Adversarial Perturbation

Throughout this section, we consider some random initialization $\widetilde{W}, A, B$, and some adversarially chosen perturbation $W' \in \mathbb{R}^{m \times m}$ which may depend on the randomness of $\widetilde{W}, A, B$. We introduce the following notations in this section

**Definition C.1.**

$$\begin{aligned}
\widetilde{g}_{i,0} = \widetilde{h}_{i,0} = 0 &\qquad g_{i,0} = h_{i,0} = 0 &\text{for } i \in [n] \\
\widetilde{g}_{i,\ell} = \widetilde{W}\widetilde{h}_{i,\ell-1} + Ax_{i,\ell} &\qquad g_{i,\ell} = (\widetilde{W} + W')h_{i,\ell-1} + Ax_{i,\ell} &\text{for } i \in [n] \text{ and } \ell \in [L] \\
\widetilde{h}_{i,\ell} = \phi(\widetilde{W}\widetilde{h}_{i,\ell-1} + Ax_{i,\ell}) &\qquad h_{i,\ell} = \phi((\widetilde{W} + W')h_{i,\ell-1} + Ax_{i,\ell}) &\text{for } i \in [n] \text{ and } \ell \in [L] \\
h'_{i,\ell} = h_{i,\ell} - \widetilde{h}_{i,\ell} &\qquad g'_{i,\ell} = g_{i,\ell} - \widetilde{g}_{i,\ell} &\text{for } i \in [n] \text{ and } \ell \in [L]
\end{aligned}$$

Define diagonal matrices $\widetilde{D}_{i,\ell}$ and $D_{i,\ell}$ by letting

$$(\widetilde{D}_{i,\ell})_{k,k} = \mathbb{1}_{(\widetilde{g}_{i,\ell})_k \geq 0} \text{ and } (D_{i,\ell})_{k,k} = \mathbb{1}_{(g_{i,\ell})_k \geq 0}.$$

Accordingly, we let diagonal matrix $D'_{i,\ell} = D_{i,\ell} - \widetilde{D}_{i,\ell}$.

**Roadmap.**

- Section C.1 proves the stability at forward when the perturbation matrix $W'$ has small spectral norm. It gives bounds on $\|g'_{i,\ell}\|_2$, $\|h'_{i,\ell}\|_2$, $\|D'_{i,\ell}\|_0$, and $\|D'_{i,\ell}g_{i,\ell}\|_2$.

- Section C.2 proves the stability of intermediate layers. It gives bound on $\|\prod_{\ell=\ell_2}^{\ell_1} D_{i,\ell}\widetilde{W}\|_2$ and $\|\prod_{\ell=\ell_2}^{\ell_1} D_{i,\ell}W\|_2$.

- Section C.3 analyzes the stability for backward. It bounds the difference $\|a^\top B(\prod_{\ell=\ell_2}^{\ell_1} \widetilde{D}_{i,\ell}\widetilde{W} - \prod_{\ell=\ell_2}^{\ell_1} D_{i,\ell}\widetilde{W})\|_2$.

- Section C.4 considers a special type of rank-one perturbation matrix $W'$, and provides stability bounds on the forward and backward propagation.

As discussed in Section 5, the results of Section C.1, C.2 and C.3 shall be used twice, once for the final training updates (see Appendix G), and once for the randomness decomposition (see Appendix E). In contrast, the results of Section C.4 shall only be used once in Appendix E.

### C.1  Forward

The goal of this section is to prove Lemma C.2,

> **Lemma C.2** (forward stability, c.f. (5.1)). *Letting $\varrho = nLd\delta^{-1}\log(m/\varepsilon)$, for any $\tau_0 \in [\varrho^{-100}, \varrho^{100}]$, with probability at least $1 - e^{-\Omega(L^6\tau_0^{4/3}m^{1/3})}$ over the randomness of $\widetilde{W}, A, B$, for every $W' \in \mathbb{R}^{m \times m}$ with $\|W'\|_2 \leq \frac{\tau_0}{\sqrt{m}}$, for every $i \in [n]$ and $\ell \in [L]$, we have*
> 
> *(a) $\|g'_{i,\ell}\|_2, \|h'_{i,\ell}\|_2 \leq O(L^6\tau_0)/m^{1/2}$,*
> 
> *(b) $\|D'_{i,\ell}\|_0 \leq O(L^{10/3}\tau_0^{2/3}) \cdot m^{2/3}$, and*
> 
> *(c) $\|D'_{i,\ell}g_{i,\ell}\|_2 \leq O(L^5\tau_0)/m^{1/2}$.*

*Proof of Lemma C.2.* Suppose $C > 1$ is a large enough constant so that the hidden constant in Lemma B.11 can be $C$. We inductively prove that one can write $g'_{i,\ell} = g'_{i,\ell,1} + g'_{i,\ell,2}$ where

$$\text{I:} \quad \|g'_{i,\ell,1}\|_2 \leq \tau_1 \cdot \frac{1}{m^{1/2}} \quad \text{II:} \quad \|g'_{i,\ell,2}\|_\infty \leq \tau_2 \cdot \frac{1}{m} \quad \text{III:} \quad \|g'_{i,\ell}\|_2 \leq \tau_3 \cdot \frac{1}{m^{1/2}}$$
$$\text{IV:} \quad \|D'_{i,\ell}\|_0 \leq \tau_4 \cdot m^{2/3} \quad \text{V:} \quad \|D'_{i,\ell}g_{i,\ell}\|_2 \leq \tau_5 \cdot \frac{1}{m^{1/2}} \quad \text{VI:} \quad \|h'_{i,\ell}\|_2 \leq (\tau_3 + \tau_5)\frac{1}{m^{1/2}} \quad . \quad \text{(C.1)}$$

Above, we choose parameters

$$\tau_1 = 5CL^4(L+2)\tau_0 \quad \tau_2 = 4L\tau_5 \log m \quad \tau_3 = \tau_1 + \tau_2 \quad \tau_4 = 10(\tau_1)^{2/3} \quad \tau_5 = 3\tau_1 \quad .$$

We emphasize that all these parameters are polynomial in $\varrho$ so negligible when comparing to $m$.

Throughout the proof, we focus on some fixed $i \in [n]$ without loss of generality, and one can always take a union bound at the end. We drop the subscript $i$ for notational simplicity. In order to prove (C.1), we first assume that it holds for all $1, 2, \ldots, \ell - 1$. In particular, we assume for all $\ell' \leq \ell - 1$,

$$\|g'_{\ell',1}\|_2 \leq \tau_1 \cdot \frac{1}{m^{1/2}} \quad \text{and} \quad \|g'_{\ell',2}\|_\infty \leq \tau_2 \cdot \frac{1}{m} \quad .$$

A useful observation is $g'_{\ell'}$ can be split into three terms

$$g'_{\ell'} = \underbrace{W'D_{\ell'-1}g_{\ell'-1}}_{z_{\ell'-1,1}} + \underbrace{\widetilde{W}D'_{\ell'-1}g_{\ell'-1}}_{z_{\ell'-1,2}} + \underbrace{\widetilde{W}\widetilde{D}_{\ell'-1}g'_{\ell'-1}}_{z_{\ell'-1,3}} \quad . \quad \text{(C.2)}$$

After recursively applying (C.2), we can write

$$g'_\ell = g'_{\ell,1} = \sum_{\ell_a=1}^{\ell-1} \left(\widetilde{W}\widetilde{D}_{\ell-1} \cdots \widetilde{W}\widetilde{D}_{\ell-\ell_a+1}\right) z_{\ell-\ell_a,1} + \left(\widetilde{W}\widetilde{D}_{\ell-1} \cdots \widetilde{W}\widetilde{D}_{\ell-\ell_a+1}\right) z_{\ell-\ell_a,2} \quad . \quad \text{(C.3)}$$

Applying Claim C.3, with probability at least $1 - e^{-\Omega(\tau_1^{4/3}m^{1/3})}$, we have for all $\ell' \leq \ell - 1$,

$$\|z_{\ell',1}\|_2 \leq \frac{\tau_0}{\sqrt{m}}\left(4\ell' + 8 + \frac{\tau_1 + \tau_2}{\sqrt{m}}\right) \quad \text{(C.4)}$$

Applying Claim C.6, we have with probability at least $1 - e^{-\Omega(\tau_1^{4/3}m^{1/3})}$, one can write

$$\left(\widetilde{W}\widetilde{D}_{\ell-1} \cdots \widetilde{W}\widetilde{D}_{\ell-\ell_a+1}\right) z_{\ell-\ell_a,2} = z_{\ell-\ell_a,2\sharp} + z_{\ell-\ell_a,2\flat}$$

where

$$\|z_{\ell-\ell_a,2\sharp}\|_2 \leq \frac{3\tau_5\sqrt{\tau_4}\log^2 m}{m^{2/3}}, \quad \|z_{\ell-\ell_a,2\flat}\|_\infty \leq \frac{4\tau_5 \log m}{m} \quad . \quad \text{(C.5)}$$

As a result, we can define $g'_\ell = g'_{\ell,1} + g'_{\ell,2}$ for

$$g'_{\ell,1} = \sum_{\ell_a=1}^{\ell-1} \left(\widetilde{W}\widetilde{D}_{\ell-1} \cdots \widetilde{W}\widetilde{D}_{\ell-\ell_a+1}\right) z_{\ell-\ell_a,1} + z_{\ell-\ell_a,2\sharp} \quad \text{and} \quad g'_{\ell,2} = \sum_{\ell_a=1}^{\ell-1} z_{\ell-\ell_a,2\flat} \quad .$$

We first bound $\|g'_{\ell,1}\|_2$,

$$\|g'_{\ell,1}\|_2 \leq \sum_{\ell_a=1}^{\ell-1} \|\widetilde{W}\widetilde{D}_{i,\ell-1}\cdots\widetilde{W}\widetilde{D}_{i,\ell-\ell_a+1}\|_2 \cdot \|z_{\ell-\ell_a,1}\|_2 + \|z_{\ell-\ell_a,2\sharp}\|_2$$

$$\overset{\text{①}}{\leq} \sum_{\ell_a=1}^{\ell} (CL^3 \|z_{\ell-\ell_a,1}\|_2 + \|z_{\ell-\ell_a,2\sharp}\|_2)$$

$$\overset{\text{②}}{\leq} CL^4 \frac{\tau_0}{\sqrt{m}}\left(4L + 8 + \frac{\tau_1+\tau_2}{\sqrt{m}}\right) + L\frac{3\tau_5\sqrt{\tau_4}\log^2 m}{m^{2/3}}$$

$$\leq 5CL^4 \cdot \frac{\tau_0(L+2)}{\sqrt{m}} \overset{\text{③}}{\leq} \tau_1\frac{1}{\sqrt{m}}\ .$$

Above, inequality ① follows from Lemma B.11, inequality ② follows from (C.4) and (C.5), and ③ follows from our choice of $\tau_1 = 5CL^4(L+2)\tau_0$. We next bound $\|g'_{\ell,2}\|_\infty$,

$$\|g'_{\ell,2}\|_\infty \leq \sum_{\ell_a=1}^{\ell-1} \|z_{\ell-\ell_a,2^\flat}\|_\infty \overset{\text{①}}{\leq} L \cdot \frac{4\tau_5\log m}{m} \overset{\text{②}}{\leq} \tau_2\frac{1}{m}$$

where inequality ① is due to (C.5), and ② follows from our choice of $\tau_2 = 4L\tau_5\log m$. Thus, we have showed I and II of (C.1):

$$\|g'_{\ell,1}\|_2 \leq \tau_1 \cdot \frac{1}{m^{1/2}} \quad \text{and} \quad \|g'_{\ell,2}\|_\infty \leq \tau_2 \cdot \frac{1}{m}\ .$$

They together further imply

$$\|g'_\ell\|_2 \leq (\tau_1+\tau_2)\frac{1}{m^{1/2}} = \tau_3\frac{1}{m^{1/2}}$$

so III of (C.1) holds. IV and V of (C.1) are implied by Claim C.4, and VI is implied because

$$\|h'_\ell\|_2 = \|\phi(g_\ell) - \phi(\widetilde{g}_\ell)\|_2 \leq \|g_\ell - \widetilde{g}_\ell\|_2 \leq \|g'_\ell\|_2\ .\qquad\square$$

### C.1.1 Tools

**Claim C.3.** *Suppose $g_{\ell-1} = \widetilde{g}_{\ell-1} + g'_{\ell-1} = \widetilde{g}_{\ell-1} + g'_{\ell-1,1} + g'_{\ell-1,2}$ where*

$$\|g'_{\ell-1,1}\|_2 \leq \frac{\tau_1}{\sqrt{m}} \quad \text{and} \quad \|g'_{\ell-1,2}\|_\infty \leq \frac{\tau_2}{m}\ .$$

*Then, we have with probability at least $1 - e^{-\Omega(m/L^2)}$*

$$\|W'D_{\ell-1}g_{\ell-1}\|_2 \leq \frac{\tau_0}{\sqrt{m}}\left(4\ell + 8 + \frac{\tau_1+\tau_2}{\sqrt{m}}\right).$$

*Proof of Claim C.3.* Using triangle inequality, we can calculate

$$\|W'D_{\ell-1}g_{\ell-1}\|_2 \leq \|W'\|_2 \cdot \|D_{\ell-1}\|_2 \cdot (\|\widetilde{g}_{\ell-1}\|_2 + \|g'_{\ell-1,1}\|_2 + \|g'_{\ell-1,2}\|_2)$$

Using Lemma B.3, we have $\|\widetilde{g}_{\ell-1}\| \leq 4\ell + 8$. Using elementary calculation, we have

$$\|g'_{\ell-1,1}\|_2 + \|g'_{\ell-1,2}\|_2 \leq \frac{\tau_1}{\sqrt{m}} + \sqrt{m}\frac{\tau_2}{m} \leq \frac{\tau_1+\tau_2}{\sqrt{m}}$$

Together we finish the proof. $\square$

**Claim C.4.** *With probability at least $1 - e^{-\Omega(\tau_1^{4/3} m^{1/3})}$ the following holds. Whenever*

$$\|g'_{\ell-1,1}\|_2 \leq \tau_1 \cdot \frac{1}{m^{1/2}}, \quad \|g'_{\ell-1,2}\|_\infty \leq \tau_2 \cdot \frac{1}{m},$$

*then letting $\tau_4 = 10(\tau_1)^{2/3}$ and $\tau_5 = 3\tau_1$, we have*

$$\|D'_{\ell-1} g_{\ell-1}\|_2 \leq \tau_5 \cdot \frac{1}{m^{1/2}}, \quad \|D'_{\ell-1}\|_0 \leq \tau_4 \cdot m^{2/3}$$

*Proof of Claim C.4.* We choose parameters $\xi = \frac{(\tau_1)^{2/3}}{10^{1/3} m^{5/6}}$ and $\alpha = 10\xi\sqrt{m}$ in the proof. We have $\|g'_{\ell-1,2}\|_\infty \leq \xi$.

First of all, using similar (but simpler) proof as Lemma B.6, one can show with probability at least $1 - \exp(-\Omega(\alpha^2 m))$, the vector $\widetilde{g}_{\ell-1}$ is $(\alpha, \frac{1}{5\sqrt{m}})$-good (recall Definition A.14).[18] This implies that $\widetilde{g}_{\ell-1}$ has at most $\alpha m = 10\xi m^{3/2}$ coordinates $j$ satisfying $|(\widetilde{g}_{\ell-1})_j| \leq \frac{\alpha}{5\sqrt{m}} = 2\xi$.

For each $j \in [m]$, if it satisfies $(D'_{\ell-1})_{j,j} \neq 0$, then the sign of $\widetilde{g}_{\ell-1}$ and $\widetilde{g}_{\ell-1} + g'_{\ell-1,1} + g'_{\ell-1,2}$ must differ on coordinate $j$. As a result:

$$|(g'_{\ell-1,1} + g'_{\ell-1,2})_j| > |(\widetilde{g}_{\ell-1})_j| .$$

Define $y = D'_{\ell-1} g_{\ell-1}$. There are two possibilities for such $j$ with $(D'_{\ell-1})_{j,j} \neq 0$.

- Case 1: $|(\widetilde{g}_{\ell-1})_j| \leq 2\xi$. Let such coordinates be $S_1 \subset [m]$, and we have $|S_1| \leq 10\xi m^{3/2}$ using the above argument.

  Next, for each such $j \in S_1$, we must have $|y_j| = |(\widetilde{g}_{\ell-1} + g'_{\ell-1,1} + g'_{\ell-1,2})_j| \leq |(g'_{\ell-1,1} + g'_{\ell-1,2})_j| \leq |(g'_{\ell-1,1})_j| + \xi$ so we can calculate the $\ell_2$ norm of $y$ on $S_1$:

  $$\sum_{i \in S_1} y_j^2 \leq 2\|g'_{\ell-1,1}\|^2 + 2\xi^2|S_1| \leq \frac{2\tau_1^2}{m} + 20\xi^3 m^{3/2} = \frac{4\tau_1^2}{m} .$$

- Case 2: $|(\widetilde{g}_{\ell-1})_j| > 2\xi$. Let such coordinates be $S_2 \subset [m] \setminus S_1$. In this case we must have $|(g'_{\ell-1,1})_j| \geq |(\widetilde{g}_{\ell-1})_j| - |(g'_{\ell-1,2})_j| > 2\xi - \xi = \xi$. Therefore, $|S_2| \leq \frac{\|g'_{\ell-1,1}\|_2^2}{\xi^2} \leq \frac{\tau_1^2}{m\xi^2}$ .

  Next, for each $j \in S_2$, we must have

  $$|y_j| = |(\widetilde{g}_{\ell-1} + g'_{\ell-1,1} + g'_{\ell-1,2})_j| \leq |(g'_{\ell-1,1} + g'_{\ell-1,2})_j| \leq |(g'_{\ell-1,1})_j| + \xi/2 \leq \frac{5}{4}|(g'_{\ell-1,1})_j| .$$

  and therefore

  $$\sum_{j \in S_2} y_j^2 \leq 2 \sum_{j \in S_2} (g'_{\ell-1,1})_j^2 \leq \frac{2\tau_1^2}{m} .$$

---

[18]Indeed,

$$\widetilde{g}_{\ell-1} = \widetilde{W}\widetilde{h}_{\ell-2} + Ax_{\ell-1} = M_1 z_1 + M_2 z_2.$$

where $M_1 \in \mathbb{R}^{m \times n(\ell-2)}$, $M_3 \in \mathbb{R}^{m \times d}$ and $z_1 \in \mathbb{R}^{n(\ell-1)}$, $z_3 \in \mathbb{R}^d$ are defined as

$$M_1 = \widetilde{W} U_{\ell-2} \qquad\qquad M_2 = A$$
$$z_1 = U_{\ell-2}^\top \widetilde{h}_{\ell-2} \qquad\qquad z_2 = x_{\ell-1} . \tag{C.6}$$

The entries of $M_1$ and $M_2$ are i.i.d. from $\mathcal{N}(0, \frac{2}{m})$ (recall Footnote 13). Suppose $z_1$ and $z_2$ are fixed (instead of depending on the randomness of $\widetilde{W}$ and $A$) and satisfies $\frac{1}{\sqrt{2}} \leq \|z_1\| \leq 2L + 6$. We can apply Corollary A.16 to obtain the following statement: with probability at least $1 - \exp(-\Omega(\alpha^2 m))$, $\widetilde{g}_{\ell-1}$ is $(\alpha, \sigma/4)$-good where $\sigma^2 = \frac{2}{m}\|z_1\|_2^2 + \frac{2}{m}\|z_3\|_2^2 \geq \frac{1}{m}$. In other words, $w$ is $(\alpha, \frac{1}{4\sqrt{m}})$-good. Applying standard $\varepsilon$-net argument, we have with probability at least $1 - \exp(-\Omega(\alpha^2 m))$, for all vectors $z_1 \in \mathbb{R}^{n(\ell-2)}$ satisfying $\frac{1}{\sqrt{2}} \leq \|z_1\| \leq 2L + 6$, it satisfies $\widetilde{g}_{\ell-1}$ is $(\alpha, \frac{1}{5\sqrt{m}})$-good. This allows us to plug in the random choice of $z_1$ in (C.6), to conclude that $\widetilde{g}_{\ell-1}$ is $(\alpha, \frac{1}{5\sqrt{m}})$-good.

In sum, we conclude that

$$\|D'_{\ell-1}\|_0 \leq |S_1| + |S_2| \leq 10\xi m^{3/2} + \frac{\tau_1^2}{m\xi^2} < 10(\tau_1)^{2/3} m^{2/3} = \tau_4 \cdot m^{2/3}$$

and

$$\|y\|_2 \leq \frac{\sqrt{4\tau_1^2 + 2\tau_1^2}}{\sqrt{m}} \leq \frac{3\tau_1}{\sqrt{m}} = \tau_5 \cdot \frac{1}{m^{1/2}} \ . \qquad \square$$

**Claim C.5.** *With probability at least $1 - e^{-\Omega(\tau_4 m^{2/3} \log^2 m)}$, for all $L \geq \ell + 1 \geq b \geq 1$, for all*

$$x \in \mathbb{R}^m \quad \text{with} \quad \|x\|_2 \leq \frac{\tau_5}{\sqrt{m}} \quad \text{and} \quad \|x\|_0 \leq \tau_4 \cdot m^{2/3}$$

*we have that the vector $y = \widetilde{W}\widetilde{D}_\ell \widetilde{W} \cdots \widetilde{D}_b \widetilde{W} x$ can be written as*

$$y = y_1 + y_2 \quad \text{where} \quad \|y_1\|_2 \leq \frac{3\tau_5 \sqrt{\tau_4} \log^2 m}{m^{2/3}} \quad \text{and} \quad \|y_2\|_\infty \leq \frac{4\tau_5 \log m}{m}$$

*Proof of Claim C.5.* Let $s = \tau_4 \cdot m^{2/3}$ for notational simplicity. Let $z_{b-1} = x$ and assume for now that $x$ is a fixed vector. Letting $z_\ell = \widetilde{D}_\ell \widetilde{W} \cdots \widetilde{D}_b z_{b-1}$, we have according to Claim B.12, with probability at least $1 - \exp(-\Omega(m/L^2))$,

$$\|z_\ell\|_2 \leq (1 + 1/L)^{\ell-b-1} \|z_{b-1}\|_2 \leq 3\|z_{b-1}\|_2.$$

Similar to the proof of Lemma B.11, we let $U$ denote the following column orthonormal matrix using Gram-Schmidt:

$$U \stackrel{\text{def}}{=} \mathsf{GS}\left(h_{1,1}, \ldots, h_{n,1}, h_{1,2}, \ldots, h_{n,2}, \ldots, h_{1,\ell-1}, \ldots, h_{n,\ell-1}, z_{b-1}, \cdots z_\ell\right) \ ,$$

and we have

$$\widetilde{W}\widetilde{D}_\ell \widetilde{W} \cdots \widetilde{D}_b \widetilde{W} x = \widetilde{W} U U^\top z_{\ell_2} \ .$$

Observe the entries of $\widetilde{W} U \in \mathbb{R}^{m \times n(\ell-1) + (\ell-b+2)}$ are i.i.d. drawn from $\mathcal{N}(0, \frac{2}{m})$ (following a similar argument as Footnote 17). Therefore, according to Lemma A.17, we have can write $y = \widetilde{W}\widetilde{D}_\ell \widetilde{W} \cdots \widetilde{D}_b \widetilde{W} x$ as $y = y_1 + y_2$ with

$$\|y_1\| \leq \frac{\sqrt{s} \log^2 m}{\sqrt{m}} \cdot \|U^\top z_{\ell_2}\| \quad \text{and} \quad \|y_2\|_\infty \leq \frac{\log m}{\sqrt{m}} \cdot \|U^\top z_{\ell_2}\| \ .$$

Plugging in $\|U^\top z_\ell\|_2 \leq \|z_\ell\|_2 \leq 3\|z_{b-1}\|_2 = 3\|x\|_2 \leq \frac{3\tau_5}{\sqrt{m}}$, we have

$$\|y_1\| \leq \frac{3\tau_5 \sqrt{s} \log^2 m}{m} \quad \text{and} \quad \|y_2\|_\infty \leq \frac{3\tau_5 \log m}{m} \ .$$

Finally, taking $\varepsilon$-net over all $s$-sparse vectors $x$, we have the desired result. $\qquad \square$

Combining Claim C.4 and Claim C.5, we have

**Claim C.6.** *With probability at least $1 - e^{-\Omega(\tau_1^{4/3} m^{1/3})}$, whenever*

$$\|g'_{\ell-1,1}\|_2 \leq \frac{\tau_1}{\sqrt{m}}, \quad \|g'_{\ell-1,2}\|_\infty \leq \frac{\tau_2}{m} \ ,$$

*we have the vector $y = \widetilde{W}\widetilde{D}_\ell \widetilde{W} \cdots \widetilde{D}_b \widetilde{W} D'_{\ell-1} g_{\ell-1}$ can be written as*

$$y = y_1 + y_2 \quad \text{where} \quad \|y_1\|_2 \leq \frac{3\tau_5 \sqrt{\tau_4} \log^2 m}{m^{2/3}} \quad \text{and} \quad \|y_2\|_\infty \leq \frac{4\tau_5 \log m}{m}$$

## C.2 Intermediate Layers

The goal of this subsection is to prove Lemma C.7.

> **Lemma C.7** (intermediate stability, c.f. (5.2)). *Letting $\varrho = nLd\log m$, for any $\tau_0 \in [\varrho^{-100}, \varrho^{100}]$, with probability at least $1 - e^{-\Omega(L^6 \tau_0^{4/3} m^{1/3})}$ over the randomness of $\widetilde{W}, A, B$, for every $W' \in \mathbb{R}^{m \times m}$ with $\|W'\|_2 \leq \frac{\tau_0}{\sqrt{m}}$, for every $i \in [n]$, for every $L \geq \ell_2 \geq \ell_1 \geq 1$, the following holds*
> 
> - $\left\| \prod_{\ell=\ell_2}^{\ell_1} (\widetilde{D}_{i,\ell} + D'_{i,\ell}) \widetilde{W} \right\|_2 \leq O(L^7)$ .
> - $\left\| \prod_{\ell=\ell_2}^{\ell_1} (\widetilde{D}_{i,\ell} + D'_{i,\ell})(\widetilde{W} + W') \right\|_2 \leq O(L^7)$ .

We show Claim C.8 and then use it to prove Corollary C.8.

**Claim C.8.** *Let $s \in [m^{-1/4}, m^{1/8}]$ be any fixed real. With probability at least $1 - e^{-\Omega(s^2 m^{2/3} \log^2 m)}$, it satisfies that for every $i \in [n]$, for every $L \geq \ell_2 \geq \ell_1 \geq 1$, if $D'_{i,\ell} \in \mathbb{R}^{m \times m}$ is an arbitrary diagonal matrix with entries in $\{-1, 0, 1\}$ of sparsity $\|D'_{i,\ell}\|_0 \leq s^2 m^{2/3}$ (for each $\ell = \ell_1, \ldots, \ell_2$), then*

$$\left\| \prod_{\ell=\ell_2}^{\ell_1} (\widetilde{D}_{i,\ell} + D'_{i,\ell}) \widetilde{W} \right\|_2 \leq O(L^7) \ .$$

*Proof of Claim C.8.* We define set $\mathcal{C}$ as follows

$$\mathcal{C} = \left\{ \prod_{\ell=\ell_2}^{\ell_1} (D'_{i,\ell})^{c_\ell} (\widetilde{D}_{i,\ell})^{1-c_\ell} \widetilde{W} \;\middle|\; c_\ell \in \{0,1\}, \forall \ell \in [\ell_1, \ell_2] \right\} .$$

For $\ell$, we define set $\mathcal{C}_\ell$ as follows

$$\mathcal{C}_\ell = \left\{ C \in \mathcal{C} \;\middle|\; D' \text{ appears } \ell \text{ times in } C \right\}$$

From Lemma B.11, we can bound the $\ell = 0$ term

$$\left\| \sum_{C \in \mathcal{C}_\ell} C \right\|_2 \leq O(L^3).$$

For each $C \in \mathcal{C}_\ell$ with $\ell > 0$, we have

$$\|C\|_2 \leq O(L^3) \cdot \left( \frac{s \log m}{m^{1/6}} \right)^{\ell-1} \cdot O(L^3)$$

where we apply Lemma B.11 twice(one on the left, one on the right) and apply Corollary B.15 $\ell - 1$ times. The probability of the above event is at least $1 - e^{\Omega(s^2 \cdot m^{2/3} \log^2 m)}$. Therefore, for each $\ell \in [\ell_2 - \ell_1]$, we can bound

$$\left\| \sum_{C \in \mathcal{C}_\ell} C \right\|_2 \leq O(L^6) \binom{\ell_2 - \ell_1}{\ell} \left( \frac{s \log m}{m^{1/6}} \right)^{\ell-1}$$

which is at most $O(L^7)$ when $\ell = 1$ and $o(L^7)$ for $\ell > 1$ by our parameter choices. Putting it altogether,

$$\text{LHS} = \left\| \sum_{\ell=0}^{\ell_2-\ell_1} \sum_{C \in \mathcal{C}_\ell} C \right\|_2 \leq \sum_{\ell=0}^{\ell_2-\ell_1} \left\| \sum_{C \in \mathcal{C}_\ell} C \right\|_2 \leq O(L^3 + L^7) \qquad \square$$

*Proof of Lemma C.7.* Applying Lemma C.2c, we know with probability $\geq 1 - e^{-\Omega(L^6 \tau_0^{4/3} m^{1/3})}$ it satisfies $\|D'_{i,\ell}\|_0 \leq s^2 m^{2/3}$ for $s^2 = O(L^{10/3} \tau_0^{2/3})$. Therefore, applying Claim C.8 we have

$$\left\| \prod_{\ell=\ell_2}^{\ell_1} (\widetilde{D}_{i,\ell} + D'_{i,\ell}) \widetilde{W} \right\|_2 \leq O(L^7) \ .$$

On the other hand, since $\|W'\| \leq \frac{\tau_0}{\sqrt{m}}$, we also have

$$\left\| \prod_{\ell=\ell_2}^{\ell_1} (\widetilde{D}_{i,\ell} + D'_{i,\ell})(\widetilde{W} + W') \right\|_2 \leq O(L^7) \ .$$

by expanding out the $2^{\ell_2 - \ell_1 + 1}$ terms with respect to $W'$. □

## C.3 Backward

We state the main result in this section.

---

**Lemma C.9** (backward stability, c.f. (5.3)). *Letting $\varrho = nLd \log m$, for any $\tau_0 \in [\varrho^{-100}, \varrho^{100}]$, with probability at least $1 - e^{-\Omega(L^6 \tau_0^{4/3} m^{1/3})}$ over the randomness of $\widetilde{W}, A, B$, for every $W' \in \mathbb{R}^{m \times m}$ with $\|W'\|_2 \leq \frac{\tau_0}{\sqrt{m}}$, for every $1 \leq \ell_1 \leq \ell_2 \leq L$, for every $i \in [n]$, the followings hold*

(a) $\left\| a^\top B \prod_{\ell=\ell_2}^{\ell_1} \widetilde{D}_{i,\ell} \widetilde{W} - a^\top B \prod_{\ell=\ell_2}^{\ell_1} D_{i,\ell} \widetilde{W} \right\|_2 \leq O(\tau_0^{1/3} L^6 \log m \cdot m^{1/3}) \cdot \|a\|_2.$

(b) $\left\| a^\top B \prod_{\ell=\ell_2}^{\ell_1} D_{i,\ell} \widetilde{W} - a^\top B \prod_{\ell=\ell_2}^{\ell_1} D_{i,\ell} W \right\|_2 \leq O(\tau_0 L^{15}) \cdot \|a\|_2.$

(c) $\left\| a^\top B \prod_{\ell=\ell_2}^{\ell_1} \widetilde{D}_{i,\ell} \widetilde{W} - a^\top B \prod_{\ell=\ell_2}^{\ell_1} \widetilde{D}_{i,\ell} W \right\|_2 \leq O(\tau_0 L^7) \cdot \|a\|_2.$

(d) $\left\| a^\top B \prod_{\ell=\ell_2}^{\ell_1} \widetilde{D}_{i,\ell} W - a^\top B \prod_{\ell=\ell_2}^{\ell_1} D_{i,\ell} W \right\|_2 \leq O(\tau_0^{1/3} L^6 \log m \cdot m^{1/3}) \cdot \|a\|_2.$

---

We first use Lemma C.2 to derive that, with probability $\geq 1 - e^{-\Omega(L^6 \tau_0^{4/3} m^{1/3})}$, $\|D'_{i,\ell}\|_0 \leq s^2 m^{2/3}$, for parameter $s = O(L^{5/3} \tau_0^{1/3})$. We prove the five statements separately.

*Proof of Lemma C.9a.* Without loss of generality we assume $\|a\| = 1$. We define set $\mathcal{C}$ to be

$$\mathcal{C} = \left\{ \prod_{\ell=\ell_2}^{\ell_1} (D'_{i,\ell})^{c_\ell} (\widetilde{D}_{i,\ell})^{1-c_\ell} \widetilde{W} \ \Big| \ c_{\ell_1}, \ldots, c_{\ell_2} \in \{0, 1\} \right\} \ .$$

For each $\ell \in [\ell_2 - \ell_1]$, we define set $\mathcal{C}_\ell$ to be

$$\mathcal{C}_\ell = \{ C \in \mathcal{C} \mid D' \text{ appears } \ell \text{ times in } C \}.$$

Ignoring subscripts for the ease of presentation, by Corollary B.18, we know

$$\|a^\top B \widetilde{D} \widetilde{W} \cdots \widetilde{D} \widetilde{W} D'\|_2 \leq (s \log m) m^{1/3} \ .$$

By Corollary B.15, we know

$$\left\| D' \widetilde{W} \widetilde{D} \cdots \widetilde{D} \widetilde{W} D' \right\|_2 \leq (s \log m) m^{-1/6} \ .$$

Thus, for each $\ell$ and $C \in \mathcal{C}_\ell$,

$$\|a^\top B C\|_2 \leq \|a^\top B \widetilde{D} \widetilde{W} \cdots \widetilde{D} \widetilde{W} D'\|_2 \cdot \left\| D' \widetilde{W} \widetilde{D} \cdots \widetilde{D} \widetilde{W} D' \right\|_2^{\ell - 1} \cdot \left\| D' \widetilde{W} \widetilde{D} \cdots \widetilde{D} \widetilde{W} \right\|_2$$

$$\leq (s \cdot m^{1/3} \log m) \times (s \cdot m^{-1/6} \log m)^{\ell - 1} \times O(L^3) \ .$$

Finally, the LHS of the Lemma C.9a becomes

$$\text{LHS} = \left\| \sum_{\ell=1}^{\ell_2-\ell_1} \sum_{C \in \mathcal{C}_\ell} a^\top BC \right\|_2 \leq \sum_{\ell=1}^{\ell_2-\ell_1} \sum_{C \in \mathcal{C}_\ell} \|a^\top BC\|_2$$

$$\leq \sum_{\ell=1}^{\ell_2-\ell_1} \binom{\ell_2-\ell_1}{\ell}(s \cdot m^{1/3}\log m) \times (s \cdot m^{-1/6}\log m)^{\ell-1} \times O(L^3) \leq O\big(L^4 \cdot s\log m \cdot m^{1/3}\big) \ .$$

$\square$

*Proof of Lemma C.9b.* Again we assume $\|a\| = 1$ without loss of generality. We define set $\mathcal{C}$ to be

$$\mathcal{C} = \left\{ \prod_{\ell=\ell_2}^{\ell_1} D_{i,\ell}(W')^{c_\ell} \widetilde{W}^{1-c_\ell} \ \middle| \ c_{\ell_1},\ldots,c_{\ell_2} \in \{0,1\} \right\}$$

For each $\ell \in [\ell_2 - \ell_1]$, we define set $\mathcal{C}_\ell$

$$\mathcal{C}_\ell = \{ C \in \mathcal{C} \mid W' \text{ appears } \ell \text{ times in } C \} \ .$$

Ignoring subscripts for the ease of presentation, we have for each $C \in \mathcal{C}_\ell$,

$$\|a^\top BC\|_2 \leq \|a^\top B\|_2 \cdot \|\widetilde{W}D\cdots\widetilde{W}D\|_2 \cdot \|W'\|_2 \cdot \big(\|D\widetilde{W}\cdots\widetilde{W}D\|_2\|W'\|_2\big)^{\ell-1} \cdot \|D\widetilde{W}\cdots D\widetilde{W}\|_2 \tag{C.7}$$

$$\leq O(\sqrt{m}L^7) \cdot \left(\frac{\tau_0}{\sqrt{m}}\right) \cdot \left(\frac{O(\tau_0 L^7)}{\sqrt{m}}\right)^{\ell-1} \cdot O(L^7) \ . \tag{C.8}$$

Above, we have used $\|W'\|_2 \leq \frac{\tau_0}{\sqrt{m}}$, $\|\widetilde{W}D\cdots\widetilde{W}D\|_2 \leq O(L^7)$ from Lemma C.7, and $\|B\|_2 \leq O(\sqrt{m})$ with high probability. Finally, by triangle inequality, the LHS of Lemma C.9b becomes

$$\text{LHS} = \left\| \sum_{\ell=1}^{\ell_2-\ell_1} \sum_{C \in \mathcal{C}_\ell} a^\top BC \right\|_2 \leq \sum_{\ell=1}^{\ell_2-\ell_1} \sum_{C \in \mathcal{C}_\ell} \left\| a^\top BC \right\|_2$$

$$\leq \sum_{\ell=1}^{\ell_2-\ell_1} \binom{\ell_2-\ell_1}{\ell} O(\sqrt{m}L^7) \cdot \left(\frac{\tau_0}{\sqrt{m}}\right) \cdot \left(\frac{O(\tau_0 L^7)}{\sqrt{m}}\right)^{\ell-1} \cdot O(L^7)$$

$$\leq O(\tau_0 L^{15})$$

$\square$

*Proof of Lemma C.9c.* It is similar to the proof of Lemma C.9b. $\square$

*Proof of Lemma C.9d.* It is similar to the proof of Lemma C.9a. $\square$

### C.4 Special Rank-One Perturbation

We prove two lemmas regarding the special rank-one perturbation with respect to a coordinate $k \in [m]$. The first one talks about forward propagation.

**Lemma C.10** (c.f. (5.4)). *Let $\rho = nLd\log m$, $\varrho = nLd\delta^{-1}\log(m/\varepsilon)$, $\tau_0 \in [\varrho^{-100}, \varrho^{50}]$, $N \in [1, \varrho^{100}]$. If $\widetilde{W}, A$ are at random initialization, then with probability at least $1 - e^{-\Omega(\rho^2)}$, for any rank-one perturbation $W' = yz^\top$ with $\|z\| = 1$, $\|y\|_0 = N$, $\|y\|_\infty \leq \frac{\tau_0}{\sqrt{m}}$, it satisfies*

$$\forall i \in [n], \forall \ell \in [L], \forall k \in [m]: \quad |((\widetilde{W} + W')h'_{i,\ell})_k| \leq O\Big(\frac{\rho^8 N^{2/3}\tau_0^{5/6}}{m^{2/3}}\Big) \ .$$

*Proof.* Without loss of generality we only prove the statement for fixed $i, \ell, k$ because one can take union bound at the end. We can rewrite $Wh'_{i,\ell}$ as follows

$$\begin{aligned}
(\widetilde{W} + W')h'_{i,\ell} &= W'h'_{i,\ell} + \widetilde{W}h'_{i,\ell} \\
&= W'h'_{i,\ell} + \widetilde{W}(h_{i,\ell} - \widetilde{h}_{i,\ell}) \\
&= W'h'_{i,\ell} + \widetilde{W}((\widetilde{D}_{i,\ell} + D'_{i,\ell})g'_{i,\ell} - \widetilde{D}_{i,\ell}\widetilde{g}_{i,\ell}) \\
&= W'h'_{i,\ell} + \widetilde{W}\widetilde{D}_{i,\ell}g'_{i,\ell} + \widetilde{W}D'_{i,\ell}g_{i,\ell} \\
&= W'h'_{i,\ell} + \widetilde{W}\widetilde{D}_{i,\ell}((\widetilde{W} + W')h_{i,\ell-1} - \widetilde{W}\widetilde{h}_{i,\ell-1}) + \widetilde{W}D'_{i,\ell}g_{i,\ell} \\
&= W'h'_{i,\ell} + \widetilde{W}\widetilde{D}_{i,\ell}(\widetilde{W}h'_{i,\ell-1} + W'h_{i,\ell-1}) + \widetilde{W}D'_{i,\ell}g_{i,\ell} \\
&= W'h'_{i,\ell} + \widetilde{W}\widetilde{D}_{i,\ell}\underbrace{\widetilde{W}h'_{i,\ell-1}}_{\text{recurse}} + \widetilde{W}\widetilde{D}_{i,\ell}W'h_{i,\ell-1} + \widetilde{W}D'_{i,\ell}g_{i,\ell}
\end{aligned}$$

After recursively calculating as above, using $h'_{i,0} = 0$, and applying triangle inequality, we have

$$|((\widetilde{W} + W')h'_{i,\ell})_k| \leq |(W'h'_{i,\ell})_k| + \sum_{a=0}^{\ell}|(\widetilde{W}\widetilde{D}_{i,\ell}\cdots\widetilde{W}\widetilde{D}_{i,a}W'h_{i,a})_k| + \sum_{a=0}^{\ell}|(\widetilde{W}\widetilde{D}_{i,\ell}\cdots\widetilde{W}\widetilde{D}_{i,a+1}\widetilde{W}D'_{i,a}g_{i,a})_k| \tag{C.9}$$

We bound the three types of terms on the RHS of (C.9) as follows.

- First, we can bound

$$|(W'h'_{i,\ell})_k| = |\mathbf{e}_k^\top yz^\top h'_{i,\ell}| = |\mathbf{e}_k^\top y| \cdot |z^\top h'_{i,\ell}| \leq \|y\|_\infty \cdot \|h'_{i,\ell}\|_2 \leq \frac{\tau_0}{\sqrt{m}} \cdot \Big(\frac{O(L^6\tau_0\sqrt{N})}{\sqrt{m}}\Big)$$

where the last step follows by $\|y\|_\infty \leq \frac{\tau_0}{\sqrt{m}}$, $\|W'\|_2 \leq \frac{\sqrt{N}\tau_0}{\sqrt{m}}$ and Lemma C.2a.

- Second, we can bound

$$\begin{aligned}
|(\widetilde{W}\widetilde{D}_{i,\ell}\cdots\widetilde{W}\widetilde{D}_{i,a}W'h_{i,a})_k| &= |\mathbf{e}_k^\top \widetilde{W}\widetilde{D}_{i,\ell}\cdots\widetilde{W}\widetilde{D}_{i,a}yz^\top h_{i,a}| \\
&= |\mathbf{e}_k^\top \widetilde{W}\widetilde{D}_{i,\ell}\cdots\widetilde{W}\widetilde{D}_{i,a}y| \cdot |z^\top h_{i,a}| \\
&\overset{①}{\leq} |\mathbf{e}_k^\top \widetilde{W}\widetilde{D}_{i,\ell}\cdots\widetilde{W}\widetilde{D}_{i,a}y| \cdot \|h_{i,a}\|_2 \\
&\overset{②}{\leq} |\mathbf{e}_k^\top \widetilde{W}\widetilde{D}_{i,\ell}\cdots\widetilde{W}\widetilde{D}_{i,a}y| \cdot O(L) \\
&\leq N \cdot \|y\|_\infty \max_j |\mathbf{e}_k^\top \widetilde{W}\widetilde{D}_{i,\ell}\cdots\widetilde{W}\widetilde{D}_{i,a}\mathbf{e}_j| \cdot O(L) \\
&\overset{③}{\leq} N \cdot \|y\|_\infty \cdot \frac{\rho}{\sqrt{m}} \cdot O(L) \\
&\overset{④}{\leq} N \cdot (\tau_0\frac{1}{\sqrt{m}}) \cdot \frac{\rho}{\sqrt{m}} \cdot O(L) \leq O\Big(\frac{N\tau_0\rho L}{m}\Big).
\end{aligned}$$

where ① follows by $\|z\|_2 \leq 1$, ② follows by Lemma B.3, ③ follows by Corollary B.16, and ④ follows by $\|y\|_\infty \leq \tau_0 \frac{1}{\sqrt{m}}$.

- Third, we can bound,

$$|(\widetilde{W}\widetilde{D}_{i,\ell}\cdots\widetilde{W}\widetilde{D}_{i,a+1}\widetilde{W}D'_{i,a}g_{i,a})_k| \leq \|\mathbf{e}_k\widetilde{W}\widetilde{D}_{i,\ell}\cdots\widetilde{W}\widetilde{D}_{i,a+1}\widetilde{W}D'_{i,a}\|_2 \cdot \|D'_{i,a}g_{i,a}\|_2$$
$$\leq O\left(\frac{L^{5/3}\tau_0^{1/3}N^{1/6}\log m}{m^{1/6}}\right) \cdot O\left(\frac{L^5\tau_0^{1/2}\sqrt{N}}{\sqrt{m}}\right)$$

where the last step follows by Corollary B.15 (which relies on Lemma C.2b to give $s = O(L^{5/3}\tau_0^{1/3}N^{1/6})$) together with Lemma C.2c.

Putting the three bounds into (C.9) (where the term one is the dominating term), we have

$$|((\widetilde{W}+W')h'_{i,\ell})_k| \leq O\left(\frac{\rho^8 N^{2/3}\tau_0^{5/6}}{m^{2/3}}\right) \quad . \qquad \square$$

The next one talks about backward propagation.

**Lemma C.11** (c.f. (5.5)). *Let $\rho = nLd\log m$, $\varrho = nLd\delta^{-1}\log(m/\varepsilon)$, $\tau_0 \in [\varrho^{-100}, \varrho^{50}]$, $N \in [1, \varrho^{100}]$. If $\widetilde{W}, A, B$ are at random initialization, and given some perturbation $W' = yz^\top$ with $\|z\| = 1$, $\|y\|_0 = N$, $\|y\|_\infty \leq \frac{\tau_0}{\sqrt{m}}$ where $W'$ can only depend on the randomness of $\widetilde{W}$ and $A$ (but not on $B$), then with probability at least $1 - e^{-\Omega(\rho^2)}$, for all $i \in [n]$, all $1 \leq \ell_1 \leq \ell_2 \leq L$, all $k \in [m]$:*

*(a)* $\left|a^\top B\left(\prod_{\ell=\ell_2}^{\ell_1} D_{i,\ell}\widetilde{W}\right)\mathbf{e}_k - a^\top B\left(\prod_{\ell=\ell_2}^{\ell_1} \widetilde{D}_{i,\ell}\widetilde{W}\right)\mathbf{e}_k\right| \leq \|a\|_2 \cdot O\left(\frac{\rho^3\tau_0^{1/3}N^{1/6}}{m^{1/6}}\right)$,

*(b)* $\left|a^\top B\left(\prod_{\ell=\ell_2}^{\ell_1} D_{i,\ell}W\right)\mathbf{e}_k - a^\top B\left(\prod_{\ell=\ell_2}^{\ell_1} D_{i,\ell}\widetilde{W}\right)\mathbf{e}_k\right| \leq \|a\|_2 \cdot O\left(\frac{\rho^{12}N^{5/6}\tau_0^{5/3}}{m^{1/3}}\right)$.

*Proof.* We prove for a fixed $i \in [n]$ and drop the subscript in $i$ for notational simplicity. Without loss of generality we assume $\|a\| = 1$.

(a) We define set $\mathcal{C}$ to be

$$\mathcal{C} = \left\{\prod_{\ell=\ell_2}^{\ell_1}(D'_\ell)^{c_\ell}(\widetilde{D}_\ell)^{1-c_\ell}\widetilde{W} \;\middle|\; c_\ell \in \{0,1\}, \forall \ell \in [\ell_1, \ell_2]\right\} .$$

For each $\ell \in [\ell_2 - \ell_1]$, we define set $\mathcal{C}_\ell$ to be

$$\mathcal{C}_\ell = \{C \in \mathcal{C} \mid D' \text{ appears } \ell \text{ times in } C\}.$$

Ignoring the subscripts for the ease of presentation, by Lemma B.11, we know

$$\|\widetilde{D}\widetilde{W}\cdots\widetilde{D}\widetilde{W}\cdot D'\|_2 \leq \|\widetilde{D}\widetilde{W}\cdots\widetilde{D}\widetilde{W}\|_2 \leq O(L^3) \quad .$$

By Corollary B.15 (which relies on Lemma C.2b to give $s = O(L^{5/3}\tau_0^{1/3}N^{1/6})$), we know

$$\left\|D'\widetilde{W}\widetilde{D}\cdots\widetilde{D}\widetilde{W}D'\right\|_2 \leq O(\frac{L^{5/3}\tau_0^{1/3}N^{1/6}}{m^{1/6}}\log m) \quad ,$$

$$\forall k \in [m]: \left\|D'\widetilde{W}\widetilde{D}\cdots\widetilde{D}\widetilde{W}\mathbf{e}_k\right\|_2 \leq O(\frac{L^{5/3}\tau_0^{1/3}N^{1/6}}{m^{1/6}}\log m) \quad .$$

Thus, for each $k \in [m]$ and $C \in \mathcal{C}_\ell$, we can bound it
$$\|C\mathbf{e}_k\| \leq L^3 \times \left(O(\frac{L^{5/3}\tau_0^{1/3}N^{1/6}}{m^{1/6}}\log m)\right)^\ell .$$

As a consequence, letting
$$v = \left(\prod_{\ell=\ell_2}^{\ell_1} D_{i,\ell}\widetilde{W}\right)\mathbf{e}_k - \left(\prod_{\ell=\ell_2}^{\ell_1} \widetilde{D}_{i,\ell}\widetilde{W}\right)\mathbf{e}_k$$

we have
$$\|v\| = \left\|\sum_{\ell=1}^{\ell_2-\ell_1} \sum_{C \in \mathcal{C}_\ell} C\mathbf{e}_k\right\|$$
$$\leq \sum_{\ell=1}^{\ell_2-\ell_1} \sum_{C \in \mathcal{C}_\ell} \|C\mathbf{e}_k\| \leq \sum_{\ell=1}^{\ell_2-\ell_1} \binom{L}{\ell} \cdot \left(O(\frac{L^{5/3}\tau_0^{1/3}N^{1/6}}{m^{1/6}}\log m)\right)^\ell \leq O\left(\frac{\rho^3\tau_0^{1/3}N^{1/6}}{m^{1/6}}\right) .$$

Finally, we use the randomness of $B$ (recall that $v$ does not depend on $B$), we conclude that with probability at least $1 - e^{-\Omega(m)}$, it satisfies
$$\text{LHS} = \left\|a^\top Bv\right\| \leq O\left(\frac{\rho^3\tau_0^{1/3}N^{1/6}}{m^{1/6}}\right) .$$

(b) This time, we define set $\mathcal{C}$ to be
$$\mathcal{C} = \left\{\prod_{\ell=\ell_2}^{\ell_1} D_\ell(W')^{c_\ell}\widetilde{W}^{1-c_\ell} \,\middle|\, c_\ell \in \{0,1\}, \forall \ell \in [\ell_1, \ell_2]\right\}$$

For each $\ell \in [\ell_2 - \ell_1]$, we define set $\mathcal{C}_\ell$
$$\mathcal{C}_\ell = \{C \in \mathcal{C} \mid W' \text{ appears } \ell \text{ times in } C\}.$$

One can carefully derive that[19]

$$|a^\top BD\widetilde{W}\cdots\widetilde{W}Dy| \leq O\left(\frac{\rho^4 N^{5/6}\tau_0^{5/3}}{m^{1/3}}\right) \enspace.$$

By Lemma C.7 we have

$$|z^\top D\widetilde{W}\cdots\widetilde{W}Dy| \leq \|z\|_2 \cdot O(L^7) \cdot \|y\|_2 \leq O\left(\frac{L^7\sqrt{N}\tau_0}{\sqrt{m}}\right) \enspace,$$

$$|z^\top D\widetilde{W}\cdots\widetilde{W}D\mathbf{e}_k| \leq \|z\|_2 \cdot O(L^7) \cdot \|\mathbf{e}_k\|_2 \leq O(L^7) \enspace.$$

Therefore, for each $C \in \mathcal{C}_\ell$ we can upper bound $\|a^\top BC\|_2$ as follows

$$|a^\top BC\mathbf{e}_k| \leq O\left(\frac{\rho^4 N^{5/6}\tau_0^{5/3}}{m^{1/3}}\right) \cdot \left(O\left(\frac{L^7\sqrt{N}\tau_0}{\sqrt{m}}\right)\right)^{\ell-1} \cdot O(L^7)$$

and therefore

$$\text{LHS} = \left|\sum_{\ell=1}^{\ell_2-\ell_1} \sum_{C\in\mathcal{C}_\ell} a^\top BC\mathbf{e}_k\right|$$

$$\leq \sum_{\ell=1}^{\ell_2-\ell_1} \binom{\ell_2-\ell_1}{\ell} \cdot O\left(\frac{\rho^4 N^{5/6}\tau_0^{5/3}}{m^{1/3}}\right) \cdot \left(O\left(\frac{L^7\sqrt{N}\tau_0}{\sqrt{m}}\right)\right)^{\ell-1} \cdot O(L^7) \leq O\left(\frac{\rho^{12} N^{5/6}\tau_0^{5/3}}{m^{1/3}}\right) \enspace.$$

$\square$

---

[19]Since we have done this too many times, let us quickly point out the calculation. By Corollary B.19, we have with probability at least $1 - e^{-\Omega(\rho^2)}$

$$|a^\top B\widetilde{D}\widetilde{W}\cdots\widetilde{W}\widetilde{D}y| \leq \|a\|_2 \cdot \sqrt{|N|}\log m \cdot \|y\|_2 \leq \|y\|_\infty \|y\|_0 \cdot \rho \leq \frac{\tau_0|N|}{\sqrt{m}} \enspace.$$

Using binomial expansion again, we have

$$|a^\top B\widetilde{D}\widetilde{W}\cdots\widetilde{W}\widetilde{D}y - a^\top BD\widetilde{W}\cdots\widetilde{W}Dy| \leq \sum_{\ell=1}^{L}\binom{L}{\ell}\|a^\top B\widetilde{D}\widetilde{W}\cdots\widetilde{W}D'\|\cdot\|D'\widetilde{W}\cdots\widetilde{W}D'\|_2^{\ell-1}\cdot\|D'\widetilde{W}\cdots\widetilde{W}\widetilde{D}y\|$$

$$\leq \sum_{\ell=1}^{L}\binom{L}{\ell}(sm^{1/3}\log m)\cdot\left(\frac{s\log m}{m^{1/6}}\right)^{\ell-1}\cdot\left(\frac{s\log m}{m^{1/6}}\right)\cdot\|y\|$$

$$\leq O(L)(sm^{1/3}\log m)\left(\frac{s\log m}{m^{1/6}}\right)\cdot\frac{\sqrt{N}\tau_0}{\sqrt{m}} = O\left(\frac{\rho^4 N^{5/6}\tau_0^{5/3}}{m^{1/3}}\right)$$

Here, we have used Corollary B.18 and Corollary B.15 with parameter $s = O(L^{5/3}\tau_0^{1/3}N^{1/6})$ chosen from Lemma C.2b, as well as $\|y\| \leq \frac{\sqrt{N}\tau_0}{\sqrt{m}}$.

# D  Indicator and Backward Coordinate Bounds

In this section, let $W \in \mathbb{R}^{m \times m}$, $A \in \mathbb{R}^{m \times d_x}$ and $B \in \mathbb{R}^{d \times m}$ denote three random matrices where each entry of $W$ and $A$ is i.i.d. sampled from $\mathcal{N}(0, \frac{2}{m})$ and each entry of $B$ is i.i.d. sampled from $\mathcal{N}(0, \frac{1}{d})$. We let $\{\mathsf{loss}_{i,\ell} \in \mathbb{R}^d \colon i \in [n], \ell \in [L]\setminus\{1\}\}$ be arbitrary fixed vectors, and let $i^*, \ell^* = \arg\max_{i,\ell} \|\mathsf{loss}_{i,\ell}\|_2$.

- In Section D.1, we prove that there are sufficiently many coordinates $k \in [m]$ such that the quantity $|(g_{i,\ell+1})_k| = |(Wh_{i,\ell} + Ax_{i,\ell+1})_k|$ is small for $(i,\ell) = (i^*, \ell^*)$ but large for $(i,\ell) \in [n] \times \{\ell^*, \ell^* + 1, L\} \setminus (i^*, \ell^*)$.

- In Section D.2, we prove that there are sufficiently many coordinates $k \in [m]$ such that $|\sum_a (\mathsf{Back}^\top_{i^*,\ell^* \to a} \cdot \mathsf{loss}_{i^*,a})_k|$ is large.

## D.1  Indicator Coordinate Bound

For this section, we fix some choice of $i^* \in [n]$ and $\ell^* \in [L]\setminus\{1\}$, and define two parameters

**Definition D.1.**
$$\boxed{\beta_+ \stackrel{\mathrm{def}}{=} \frac{\delta}{\rho^2} \quad \text{and} \quad \beta_- \stackrel{\mathrm{def}}{=} \frac{\delta}{\rho^{10}}}$$

**Lemma D.2** (indicator coordinate bound, c.f. (6.3)). *Suppose $W$ and $A$ follow from random initialization. Given fixed set $N_1 \subseteq [m]$ and define*
$$N_4 \stackrel{\mathrm{def}}{=} \left\{ k \in N_1 \ \middle| \ \begin{array}{ll} |(Wh_{i,\ell} + Ax_{i,\ell+1})_k| \leq \beta_-/\sqrt{m}, & \text{if } i = i^*, \ell = \ell^*; \\ |(Wh_{i,\ell} + Ax_{i,\ell+1})_k| \geq \beta_+/\sqrt{m}, & \text{if } i \neq i^* \text{ and } \ell = \ell^*; \\ |(Wh_{i,\ell} + Ax_{i,\ell+1})_k| \geq \beta_+/\sqrt{m}, & \text{if } i \in [n] \text{ and } \ell > \ell^*. \end{array} \right\}$$
*Then, as long as $|N_1| \geq n^2 L^2/\beta_-$, we have*
$$\Pr_{W,A}\left[|N_4| \geq \frac{\beta_-|N_1|}{64L}\right] \geq 1 - e^{-\Omega(\beta_-|N_1|/L)}.$$

**Lemma D.3.** *Suppose $W$ and $A$ follow from random initialization. Given fixed set $N_1 \subseteq [m]$ and define*
$$N_2 \stackrel{\mathrm{def}}{=} \left\{ k \in N_1 \ \middle| \ |\langle W_k, h_{i^*, \ell^*}\rangle + \langle A_k, x_{i^*, \ell^*+1}\rangle| \leq \frac{\beta_-}{\sqrt{m}} \right\}.$$
*Then, we have*
$$\Pr_{W,A}\left[|N_2| \geq \frac{\beta_-|N_1|}{16L}\right] \geq 1 - e^{-\Omega(\beta_-|N_1|/L)}.$$

*Proof.* Let $i = i^*$ and $\ell = \ell^*$. Similar to the proof of Lemma B.3, we can write
$$y = Wh_{i,\ell} + Ax_{i,\ell+1} = WU_\ell U_\ell^\top h_{i,\ell} + Ax_{i,\ell+1} = M_1 z_1 + M_2 z_2$$
where $M_1 \in \mathbb{R}^{m \times n\ell}$, $M_2 = A$, $z_1 = U_\ell^\top h_{i,\ell}$ and $z_2 = x_{i,\ell+1}$. Again, the entries of $M_1$ and $M_2$ are i.i.d. from $\mathcal{N}(0, \frac{2}{m})$.

Now, suppose $z_1$ is fixed (instead of depending on the randomness of $W$ and $A$), and it satisfies $\|z_1\| \leq 6L$. We have each entry of $y$ is is distributed as $\mathcal{N}\left(0, \frac{2\|z_1\|^2 + 2\|z_2\|^2}{m}\right)$. By property of Gaussian

(see Fact A.12), we have for each $y \in [m]$,
$$\mathbf{Pr}\left[|y_k| \leq \frac{0.9\beta_-}{\sqrt{m}}\right] \geq \frac{2}{3}\frac{0.9\beta_-}{\sqrt{2\|z_1\|^2 + 2\|z_2\|^2}} \geq \frac{\beta_-}{15L} \enspace .$$

Since we have $|N_1|$ independent random variables, applying a Chernoff bound, we have
$$\mathbf{Pr}\left[\left|\left\{k \in N_1 \colon |y_k| \leq \frac{0.9\beta_-}{\sqrt{m}}\right\}\right| \geq \frac{\beta_-}{16L}|N_1|\right] \geq 1 - \exp(-\Omega(\beta_-|N_1|/L)) \enspace .$$

Finally, by taking a union bound with respect to the $\varepsilon$-net over all possible choices of $z_1 \in \mathbb{R}^{n\ell}$, we have that
$$\mathbf{Pr}\left[\left|\left\{k \in N_1 \colon |y_k| \leq \frac{\beta_-}{\sqrt{m}}\right\}\right| \geq \frac{\beta_-}{16L}|N_1|\right] \geq 1 - \exp(-\Omega(\beta_-|N_1|/L)) \enspace .$$

for all $z_1$ and a fixed $z_2$. Plugging in the random choice of $z_1 = U_\ell^\top h_{i,\ell}$ (and we have $\|z_1\| \leq 6L$ by Lemma B.3), we have the desired result. $\square$

Observe that the set $N_2$ produced by Lemma D.3 only depends on the randomness of $A$, $WU_{\ell^*-1}$ and $Wh_{i^*,\ell^*}$. In other words, for any unit vector $z \in \mathbb{R}^m$ that is orthogonal to the columns of $U_{\ell^*-1}$ and orthogonal to $h_{i^*,\ell^*}$, we have that $Wz$ is independent of $N_2$.

We next proceed to refine $N_2$:

**Lemma D.4** ($i \neq i^*, \ell = \ell^*$). *Suppose $W$ and $A$ follow from random initialization, and $N_2 \subseteq [m]$ is a given set (that may only depend on the randomness of $A$, $WU_{\ell^*-1}$ and $Wh_{i^*,\ell^*}$). Define*
$$N_3 = \left\{k \in N_2 \;\middle|\; |\langle W_k, h_{i,\ell^*}\rangle + \langle A_k, x_{i,\ell^*+1}\rangle| \geq \frac{\beta_-}{\sqrt{m}}, \forall i \in [n]\backslash i^*\right\} .$$

*Then*
$$\mathbf{Pr}\left[|N_3| \geq \frac{1}{2}|N_2| \;\middle|\; A, WU_{\ell^*-1}, Wh_{i^*,\ell^*}\right] \geq 1 - e^{-\Omega(|N_2|/n)}.$$

*Proof.* First fix some $i \in [n]\backslash i^*$ and $\ell = \ell^*$. Define set $N_{3,i}$,
$$N_{3,i} \stackrel{\text{def}}{=} \left\{k \in N_2 \;\middle|\; |\langle W_k, h_{i,\ell}\rangle + \langle A_k, x_{i,\ell+1}\rangle| \geq \frac{\beta_-}{\sqrt{m}}\right\} \enspace .$$

Let $v \stackrel{\text{def}}{=} \frac{(I-U_{\ell-1}U_{\ell-1}^\top)h_{i^*,\ell}}{\|(I-U_{\ell-1}U_{\ell-1}^\top)h_{i^*,\ell}\|}$, and let column orthonormal matrix
$$U \stackrel{\text{def}}{=} \mathsf{GS}(U_{\ell-1}, h_{i^*,\ell}) = [U_{\ell-1}, v] \enspace .$$

We have
$$\langle W_k, h_{i,\ell}\rangle + \langle A_k, x_{i,\ell+1}\rangle = \left(WUU h_{i,\ell} + Ax_{i,\ell+1} + W(I-UU^\top)h_{i,\ell}\right)_k \enspace .$$

The rest of the proof is focusing on having the $WU$ and $A$ being fixed (so the first two vectors $WUUh_{i,\ell} + Ax_{i,\ell+1}$ are fixed), and letting $W(I-UU^\top)h_{i,\ell}$ be the only random variable. We have
$$W(I-UU^\top)h_{i,\ell} \sim \mathcal{N}\left(0, \frac{2\|(I-UU^\top)h_{i,\ell}\|_2^2 \cdot I}{m}\right) \enspace .$$

and therefore for a fixed vector $\mu$ it satisfies
$$Wh_{i,\ell} + Ax_{i,\ell+1} = WUUh_{i,\ell} + Ax_{i,\ell+1} + W(I-UU^\top)h_{i,\ell} \sim \mathcal{N}\left(\mu, \frac{2\|(I-UU^\top)h_{i,\ell}\|_2^2 \cdot I}{m}\right) \enspace .$$

According to Lemma B.9, with probability at least $1-\exp(-\Omega(\sqrt{m}))$, it satisfies $\|(I-UU^\top)h_{i,\ell}\|_2 \geq \frac{\delta}{2}$. By property of Gaussian distribution (see Fact A.12), we have for each $k \in N_2$:

$$\mathbf{Pr}\left[|(Wh_{i,\ell} + Ax_{i,\ell+1})_k| \geq \frac{\beta_-}{\sqrt{m}} \,\Big|\, WU, A\right] \geq 1 - \frac{\beta_-}{\delta} \geq 1 - \frac{1}{4n} \ .$$

Applying Chernoff bound, with probability at least $1 - e^{-\Omega(|N_2|/n)}$, we have $|N_{3,i}| \geq (1 - \frac{1}{2n})|N_2|$. Applying union bound over all possible $i \in [n] \setminus \{i^*\}$, we have $N_3 = \bigcap_{i \neq i^*} N_{3,i}$ has cardinality at least $\frac{|N_2|}{2}$. □

**Lemma D.5** ($\ell > \ell^*$). *Suppose $W$ and $A$ follow from random initialization, and $N_3 \subseteq [m]$ is a given set (that may only depend on the randomness of $A$, $WU_{\ell^*}$). Define*

$$N_4 = \left\{k \in N_3 \,\Big|\, |\langle W_k, h_{i,\ell}\rangle + \langle A_k, x_{i,\ell+1}\rangle| \geq \frac{\beta_-}{\sqrt{m}}, \forall i \in [n], \ell > \ell^*\right\} \ .$$

*Then*

$$\mathbf{Pr}\left[|N_4| \geq \frac{1}{2}|N_3| \,\Big|\, A, WU_{\ell^*}\right] \geq 1 - e^{-\Omega(|N_3|/nL)} \ .$$

*Proof.* Letting $N_{4,\ell^*} \stackrel{\text{def}}{=} N_3$. For any $i \in [n]$ and $\ell = \ell^*+1, \ell^*+2, \ldots, L$, we define $N_{4,i,\ell}$ as

$$N_{4,i,\ell} \stackrel{\text{def}}{=} \left\{k \in N_{4,\ell-1} \,\Big|\, |\langle W_k, h_{i,\ell}\rangle + \langle A_k, x_{i,\ell+1}\rangle| \geq \frac{\beta_-}{\sqrt{m}}\right\} \quad \text{and} \quad N_{4,\ell} \stackrel{\text{def}}{=} \bigcap_{i \in [n]} N_{4,i,\ell} \ .$$

We rewrite

$$\langle W_k, h_{i,\ell}\rangle + \langle A_k, x_{i,\ell+1}\rangle = (WU_{\ell-1}U_{\ell-1}^\top h_{i,\ell} + Ax_{i,\ell+1} + W(I - U_{\ell-1}U_{\ell-1}^\top)h_{i,\ell})_k \ .$$

The rest of the proof is focusing on having the $WU_{\ell-1}$ and $A$ being fixed (so the first two vectors $WUUh_{i,\ell} + Ax_{i,\ell+1}$ are fixed and $N_{4,\ell-1}$ is fixed), and letting $W(I - U_{\ell-1}U_{\ell-1}^\top)h_{i,\ell}$ be the only random variable. We have

$$W(I - U_{\ell-1}U_{\ell-1}^\top)h_{i,\ell} \sim \mathcal{N}\left(0, \frac{2\|(I - U_{\ell-1}U_{\ell-1}^\top)h_{i,\ell}\|_2^2 \cdot I}{m}\right) \ .$$

and therefore for a fixed vector $\mu$ it satisfies

$$Wh_{i,\ell} + Ax_{i,\ell+1} = WU_{\ell-1}U_{\ell-1}h_{i,\ell} + Ax_{i,\ell+1} + W(I - U_{\ell-1}U_{\ell-1}^\top)h_{i,\ell} \sim \mathcal{N}\left(\mu, \frac{2\|(I - U_{\ell-1}U_{\ell-1}^\top)h_{i,\ell}\|_2^2 \cdot I}{m}\right) \ .$$

According to Lemma B.6, with probability at least $1 - \exp(-\Omega(\sqrt{m}))$, it satisfies $\|(I - UU^\top)h_{i,\ell}\|_2 \geq \frac{\delta}{2}$. By property of Gaussian distribution (see Fact A.12), we have for each $k \in N_{4,\ell-1}$:

$$\mathbf{Pr}\left[|(\widetilde{W}h_{i,\ell} + Ax_{i,\ell+1})_k| \geq \frac{\beta_-}{\sqrt{m}} \,\Big|\, WU, A\right] \geq 1 - \frac{\beta_-}{\delta} \geq 1 - \frac{1}{4nL} \ .$$

Applying Chernoff bound, we have

$$\mathbf{Pr}\left[|N_{4,i,\ell}| \geq (1 - \frac{1}{3nL})|N_{4,\ell-1}| \,\Big|\, WU_{\ell-1}, A\right] \geq 1 - e^{-\Omega(|N_{4,\ell-1}|/nL)} \ .$$

Applying union bound over all possible $i \in [n]$, we have $N_{4,\ell} = \bigcap_{i \in [n]} N_{4,i,\ell}$ has cardinality at least $(1 - \frac{1}{3L})|N_{4,\ell-1}|$. After telescoping we have $|N_4| = |N_{4,L}| \geq \frac{1}{2}|N_3|$. □

### D.2 Backward Coordinate Bound

Recall the following notions

**Definition D.6.** *Given matrices $W, A, B$. For each $i$, for each $\ell$, we define*
$$\mathsf{Back}_{i,\ell \to \ell} = B \in \mathbb{R}^{d \times m}.$$
*For each $i$, for each $a \geq \ell + 1$, we define*
$$\mathsf{Back}_{i,\ell \to a} = B D_{i,a} W \cdots D_{i,\ell+1} W \in \mathbb{R}^{d \times m}.$$

The goal of this section is to prove Lemma D.7.

**Lemma D.7** (backward coordinate bound, c.f. (6.2)). *Let $\rho = ndL \log m$ and $\varrho = nLd\delta^{-1} \log(m/\varepsilon)$. Let $\{\mathsf{loss}_{i,\ell} \in \mathbb{R}^d \colon i \in [n], \ell \in [L] \setminus \{1\}\}$ be fixed vectors and $i^*, \ell^* = \arg\max_{i,\ell} \|\mathsf{loss}_{i,\ell}\|_2$. If $W \in \mathbb{R}^{m \times m}, A \in \mathbb{R}^{m \times d_x}, B \in \mathbb{R}^{d \times m}$ are random, and $N_4 \subseteq [m]$ is a set with $|N_4| \in [\rho^4, \varrho^{100}]$ ($N_4$ can depend on the randomness of $W$ and $A$, but not depend on $B$). Define*
$$N_5 \stackrel{\text{def}}{=} \left\{ k \in N_4 \;\middle|\; \left| \sum_{a=\ell^*}^{L} (\mathsf{Back}_{i^*, \ell^* \to a}^\top \cdot \mathsf{loss}_{i^*, a})_k \right| \geq \frac{\|\mathsf{loss}_{i^*, \ell^*}\|_2}{6\sqrt{d}nL} \right\}.$$
*Then with probability at least $1 - e^{-\Omega(\rho^2)}$, we have*
$$|N_5| \geq \left(1 - \frac{1}{2nL}\right) |N_4|.$$

*Proof.* For notational simplicity, in the proof we use $\ell$ to denote $\ell^*$, use $N$ to denote $N_4$, and drop the subscript $i^*$. We denote $B \in \mathbb{R}^{d \times m}$ as $[b_1, b_2, \cdots, b_m]$ where each $b_i \in \mathbb{R}^d$. We denote by $C_{a,\ell+1} = D_a W \cdots D_{\ell+1} W$.

We define a function $v_k(b_1, b_2, \cdots, b_m) \in \mathbb{R}$ as follows
$$v_k(b_1, b_2, \cdots, b_m) \stackrel{\text{def}}{=} \sum_{a=\ell}^{L} (\mathsf{Back}_{\ell \to a}^\top \cdot \mathsf{loss}_{i,a})_k.$$

We can rewrite
$$v_k(b_1, b_2, \cdots, b_m) = \left( B^\top \cdot \mathsf{loss}_{i,\ell} + \sum_{a=\ell+1}^{L} \mathsf{Back}_{\ell \to a}^\top \cdot \mathsf{loss}_{i,a} \right)_k$$
$$= \left( B^\top \cdot \mathsf{loss}_{i,\ell} + \sum_{a=\ell+1}^{L} (C_{a,\ell+1})^\top B^\top \cdot \mathsf{loss}_{i,a} \right)_k$$
$$= \langle b_k, \mathsf{loss}_{i,\ell} \rangle + \sum_{a=\ell+1}^{L} \left\langle \sum_{j=1}^{m} (C_{a,\ell+1})_{k,j} \cdot b_j, \mathsf{loss}_{i,a} \right\rangle$$
$$= \left\langle b_k, \mathsf{loss}_{i,\ell} + \sum_{a=\ell+1}^{L} (C_{a,\ell+1})_{k,j} \cdot \mathsf{loss}_{i,a} \right\rangle + \sum_{j \in [m] \setminus k} \left\langle b_j, \sum_{a=\ell+1}^{L} (C_{a,\ell+1})_{k,j} \cdot \mathsf{loss}_{i,a} \right\rangle$$

where we have used $(C_{a,\ell+1})_{k,j}$ to denote the $(k, j)$ entry of matrix $C_{a,\ell+1}$.

Choose parameter $\mathfrak{L}_h = \frac{3\sqrt{d}nL}{\|\mathsf{loss}_{i,\ell}\|_2}$ and we can define function $h \colon \mathbb{R} \to [0, 1]$ by
$$h(t) \stackrel{\text{def}}{=} \begin{cases} |t| \cdot \mathfrak{L}_h, & \text{if } |t| \leq 1/\mathfrak{L}_h; \\ 1, & \text{otherwise.} \end{cases}$$

We have $h(t)$ is $\mathfrak{L}_h$-Lipschitz continuous.

We define two probabilistic events $E_1, E_2$.

- Event $E_1$ depends on the randomness of $W$ and $A$:
$$E_1 \overset{\text{def}}{=} \left\{ |z^\top C_{a,\ell+1} y| \leq \frac{\rho}{\sqrt{m}} \|z\| \|y\| \text{ for all } a, \ell \text{ and for all 1-sparse vectors } y, z \right\}$$

Corollary B.16 says $\mathbf{Pr}[E_1] \geq 1 - e^{-\Omega(\rho^2)}$.

- Event $E_2$ depends on the randomness of $B$:
$$E_2 \overset{\text{def}}{=} \left\{ |B_{i,j}| \leq \frac{\rho}{\sqrt{d}} \text{ for all } i, j \right\}$$

By Gaussian tail bound, we have $\mathbf{Pr}[E_2] \geq 1 - e^{-\Omega(\rho^2)}$.

In the rest of the proof, we assume $W$ and $A$ are fixed and satisfy $E_1$. We let $B$ be the only source of randomness but we condition on $B$ satisfies $E_2$.

For simplicity we use subscript $-j$ to denote $[m] \setminus j$, and subscript $-N$ to denote $[m] \setminus N$. For instance, we shall write $b = (b_{-k}, b_k) = (b_N, b_{-N})$.

**Step 1.** Fixing $b_{-N}$ and letting $b_N$ be the only randomness, we claim for each $k \in N$:
$$\mathbb{E}_{b_N}[h(v_k)] \geq \mathbf{Pr}_{b_N}\left[|v_k| \geq \frac{1}{\mathfrak{L}_h}\right] \geq 1 - \frac{1}{4nL} \ .$$

We prove this inequality as follows. We can rewrite
$$v_k = \left\langle b_k, \mathsf{loss}_{i,\ell} + \sum_{a=\ell+1}^{L} (C_{a,\ell+1})_{k,k} \cdot \mathsf{loss}_{i,a} \right\rangle + \sum_{j \in N \setminus \{k\}} \left\langle b_j, \sum_{a=\ell+1}^{L} (C_{a,\ell+1})_{k,j} \cdot \mathsf{loss}_{i,a} \right\rangle$$
$$+ \sum_{j \in [m] \setminus N} \left\langle b_j, \sum_{a=\ell+1}^{L} (C_{a,\ell+1})_{k,j} \cdot \mathsf{loss}_{i,a} \right\rangle$$

and $v_k$ is distributed as Gaussian random variable $\mathcal{N}(\mu, \sigma^2)$, and $\mu$ and $\sigma^2$ are defined as follows
$$\mu = \sum_{j \in [m] \setminus N} \left\langle b_j, \sum_{a=\ell+1}^{L} (C_{a,\ell+1})_{k,j} \cdot \mathsf{loss}_{i,a} \right\rangle$$
$$\sigma^2 = \frac{1}{d}\left\|\mathsf{loss}_{i,\ell} + \sum_{a=\ell+1}^{L} (C_{a,\ell+1})_{k,k} \cdot \mathsf{loss}_{i,a}\right\|_2^2 + \frac{1}{d}\sum_{j \in N \setminus \{k\}} \left\|\sum_{a=\ell+1}^{L} (C_{a,\ell+1})_{k,j} \cdot \mathsf{loss}_{i,a}\right\|_2^2$$

Meanwhile, we calculate that
$$\left\|\sum_{a=\ell+1}^{L} (C_{a,\ell+1})_{k,j} \cdot \mathsf{loss}_{i,a}\right\|_2 \overset{\text{①}}{\leq} \sum_{a=\ell+1}^{L} |(C_{a,\ell+1})_{k,j}| \cdot \|\mathsf{loss}_{i,a}\|_2$$
$$\overset{\text{②}}{\leq} \sum_{a=\ell+1}^{L} |(C_{a,\ell+1})_{k,j}| \cdot \|\mathsf{loss}_{i,\ell}\|_2 \overset{\text{③}}{\leq} \frac{L\rho}{\sqrt{m}} \cdot \|\mathsf{loss}_{i,\ell}\|_2 \quad (D.1)$$

where ① follows by triangle inequality, ② follows by $i, \ell = \arg\max_{i',\ell'} \|\mathsf{loss}_{i',\ell'}\|$, and ③ follows from event $E_1$. Therefore,

$$\sigma^2 \geq \frac{1}{d}\left(\|\mathsf{loss}_{i,\ell}\|_2 - \left\|\sum_{a=\ell+1}^{L}(C_{a,\ell+1})_{k,j}\cdot \mathsf{loss}_{i,a}\right\|_2\right)^2$$

$$\geq \frac{1}{d}\left(1 - \frac{L\rho}{\sqrt{m}}\right)^2 \|\mathsf{loss}_{i,\ell}\|_2^2 \geq \frac{1}{2d}\|\mathsf{loss}_{i,\ell}\|_2^2$$

where the first step follows by $\|a+b\|_2 \geq \|a\|_2 - \|b\|_2$, the second step follows by (D.1). In sum, we have that $v_k$ is a random Gaussian with $\sigma^2 \geq \frac{1}{2d}\|\mathsf{loss}_{i,\ell}\|_2^2$. This means,

$$\Pr_{b_N}\left[|v_k| \geq \frac{1}{\mathfrak{L}_h}\right] \geq \Pr_{b_N}\left[|v_k| \geq \frac{\|\mathsf{loss}_{i,\ell}\|_2}{3\sqrt{d}nL}\right] \geq 1 - \frac{1}{4nL} \ .$$

according to Fact A.12.

**Step 2.** For every $b_1, b_2, \cdots, b_m$ and $b_k' \in \mathbb{R}^d$ satisfying event $E_2$, for every $j \neq k$, we have

$$|v_j(b_{-k}, b_k) - v_j(b_{-k}, b_k')| = \left|\left\langle b_k - b_k', \sum_{a=\ell+1}^{L}(C_{a,\ell+1})_{k,k}\cdot \mathsf{loss}_{i,a}\right\rangle\right|$$

$$\leq \|b_k - b_k'\|_2 \cdot \left\|\sum_{a=\ell+1}^{L}(C_{a,\ell+1})_{k,k}\cdot \mathsf{loss}_{i,a}\right\|_2$$

$$\leq \rho \cdot \frac{L\rho}{\sqrt{m}}\|\mathsf{loss}_{i,\ell}\|_2 \leq \frac{\rho^3}{\sqrt{m}}\|\mathsf{loss}_{i,\ell}\|_2 \quad \text{(D.2)}$$

where the second step follows by $|\langle a, b\rangle| \leq \|a\|_2 \cdot \|b\|_2$, the third step follows by (D.1).

**Step 3.** Fixing $b_{-N}$ and letting function $g(b_1, b_2, \cdots, b_m) \stackrel{\text{def}}{=} \sum_{k \in N} h(v_k(b_1, b_2, \cdots, b_m))$. We have

$$|g(b_j, b_{-j}) - g(b_j', b_{-j})| = \left|\sum_{k\in N} h(v_k(b_j, b_{-j})) - \sum_{k \in N} h(v_k(b_j', b_{-j}))\right|$$

$$\leq \sum_{k \in N}\left|h(v_k(b_j, b_{-j})) - h(v_k(b_j', b_{-j}))\right|$$

$$= \left|h(v_j(b_j, b_{-j})) - h(v_j(b_j', b_{-j}))\right| + \sum_{k \in N\setminus\{j\}}\left|h(v_k(b_j, b_{-j})) - h(v_k(b_j', b_{-j}))\right|$$

$$\overset{①}{\leq} 1 + \sum_{k \in N\setminus\{j\}}\left|h(v_k(b_j, b_{-j})) - h(v_k(b_j', b_{-j}))\right|$$

$$\leq 1 + \sum_{k \in N\setminus\{j\}}\mathfrak{L}_h \cdot |v_k(b_j, b_{-j}) - v_k(b_j', b_{-j})|$$

$$\overset{②}{\leq} 1 + |N| \cdot \frac{3\sqrt{d}nL}{\|\mathsf{loss}_{i,\ell}\|_2}\cdot\left((nLd\log m)^3 \frac{\|\mathsf{loss}_{i,\ell}\|_2}{\sqrt{m}}\right)$$

$$\leq 1 + \frac{|N|\cdot 3(nLd\log m)^4}{\sqrt{m}} \leq 2 \ .$$

Above, inequality ① follows by $h(t) \in [0,1]$, inequality ② follows from (D.2). Using McDiarmid inequality (see Lemma A.18), we have

$$\Pr_{b_N}\left[\left|g(b_1,\cdots,b_m) - \mathbb{E}_{b_N}[g]\right| \geq \varepsilon\right] \leq \exp\left(-\frac{\varepsilon^2}{8|N|}\right)$$

$$\implies \Pr_{b_N}\left[\left|g(b_1,\cdots,b_m) - \mathbb{E}_{b_N}[g]\right| \geq \frac{|N|}{4nL}\right] < \exp\left(-\frac{|N|}{128(nL)^2}\right)$$

where we choose $\varepsilon = \frac{|N|}{4nL}$.

Finally, combining this with Step 1 —which gives $\mathbb{E}_{b_N}[g(b_1,\cdots,b_m)] \geq |N|(1-\frac{1}{4nL})$— we have with probability at least $1 - e^{-\Omega(|N|/(nL)^2)}$ over randomness of $b_N$,

$$g(b_1,\cdots,b_m) \geq |N|\left(1 - \frac{1}{2nL}\right) \ .$$

This implies at least $(1-\frac{1}{2nL})$-fraction of $k \in N$ will have $|v_k| \geq \frac{1}{2\mathfrak{L}_h} = \frac{\|\mathsf{loss}_{i,\ell}\|_2}{6nL\sqrt{d}}$. □

# E  Gradient Bound at Random Initialization (Theorem 5)

The goal of this section is to understand the lower and upper bounds of gradients. Instead of analyzing the true gradient directly—where the forward and backward propagation have correlated randomness— we assume that the loss vectors $\{\mathsf{loss}_{i,a}\}_{i\in[n],a\in\{2,\ldots,L\}}$ are fixed (no randomness) in this section, as opposed to being defined as $\mathsf{loss}_{i,a} = Bh_{i,a} - y^*_{i,a}$ which is random. We call this the "fake loss." We define a corresponding "fake gradient" with respect to this fixed loss.

**Definition E.1** (fake gradient, c.f. (6.1)). *Given fixed vectors $\{\mathsf{loss}_{i,a}\}_{i\in[n],a\in\{2,\ldots,L\}}$, we define*

$$\widehat{\nabla}_k f(W) \overset{\text{def}}{=} \sum_{i=1}^n \sum_{a=2}^L \sum_{\ell=1}^{a-1} (\mathsf{Back}^\top_{i,\ell+1\to a} \cdot \mathsf{loss}_{i,a})_k \cdot h_{i,\ell} \cdot \mathbf{1}_{\langle W_k, h_{i,\ell}\rangle + \langle A_k, x_{i,\ell+1}\rangle \geq 0}$$

$$\widehat{\nabla} f(W) \overset{\text{def}}{=} \sum_{i=1}^n \sum_{a=2}^L \sum_{\ell=1}^{a-1} D_{i,\ell+1}(\mathsf{Back}^\top_{i,\ell+1\to a} \cdot \mathsf{loss}_{i,a}) \cdot h_{i,\ell}^\top$$

$$\widehat{\nabla} f_i(W) \overset{\text{def}}{=} \sum_{a=2}^L \sum_{\ell=1}^{a-1} D_{i,\ell+1}(\mathsf{Back}^\top_{i,\ell+1\to a} \cdot \mathsf{loss}_{i,a}) \cdot h_{i,\ell}^\top$$

In our analysis, we also write it as

$$\widehat{\nabla}_k f(W) = \sum_{i=1}^n \sum_{\ell=1}^{L-1} (u_{i,\ell})_k \cdot h_{i,\ell} \cdot \mathbf{1}_{(g_{i,\ell+1})_k \geq 0}$$

$$\widehat{\nabla} f(W) = \sum_{i=1}^n \sum_{\ell=1}^{L-1} D_{i,\ell+1} u_{i,\ell} \cdot h_{i,\ell}^\top$$

$$\widehat{\nabla} f_i(W) = \sum_{\ell=1}^{L-1} D_{i,\ell+1} u_{i,\ell} \cdot h_{i,\ell}^\top$$

where vectors $u_{i,\ell}, g_{i,\ell+1} \in \mathbb{R}^m$ are defined as

$$u_{i,\ell} \overset{\text{def}}{=} \sum_{a=\ell+1}^L \mathsf{Back}^\top_{i,\ell+1\to a} \cdot \mathsf{loss}_{i,a} \qquad\qquad g_{i,\ell+1} \overset{\text{def}}{=} Wh_{i,\ell} + Ax_{i,\ell+1} \ .$$

We first state a simple upper bound on the gradients.

**Lemma E.2.** *Letting $\rho = nLd\log m$, given fixed vectors $\{\mathsf{loss}_{i,\ell} \in \mathbb{R}^d\}_{i \in [n], \ell \in [L]\setminus\{1\}}$, with probability at least $1 - e^{-\Omega(m/L^2)}$ over $W, A, B$, we have*

$$\|\widehat{\nabla} f(W)\|_F \leq O(nL^6\sqrt{m}) \cdot \max_{i,\ell}\{\|\mathsf{loss}_{i,\ell}\|\} \quad \text{and}$$

$$\forall i \in [n]: \|\widehat{\nabla} f_i(W)\|_F \leq O(L^6\sqrt{m}) \cdot \max_{i,\ell}\{\|\mathsf{loss}_{i,\ell}\|\} .$$

*With probability at least $1 - e^{-\Omega(\rho^2)}$, we have for every $k \in [m]$:*

$$\|\widehat{\nabla}_k f(W)\|_2 \leq O(n\rho L^3) \cdot \max_{i,\ell}\{\|\mathsf{loss}_{i,\ell}\|\} .$$

*Proof of Lemma E.2.* For each $i, a, \ell$, we can calculate that

$$\left\| D_{i,\ell+1}(\mathsf{Back}_{i,\ell+1\to a}^\top \cdot \mathsf{loss}_{i,a}) \cdot h_{i,\ell}^\top \right\|_F = \left\| D_{i,\ell+1}(\mathsf{Back}_{i,\ell+1\to a}^\top \cdot \mathsf{loss}_{i,a}) \right\|_2 \cdot \|h_{i,\ell}^\top\|_2$$

$$\leq \|\mathsf{Back}_{i,\ell+1\to a}\|_2 \cdot \|\mathsf{loss}_{i,a}\|_2 \cdot \|h_{i,\ell}^\top\|_2$$

$$\stackrel{①}{\leq} O(\sqrt{m}) \cdot O(L^3) \cdot O(L) \cdot \|\mathsf{loss}_{i,a}\|_2 .$$

where inequality ① uses Lemma B.11, Lemma B.3, and $\|B\|_2 \leq O(\sqrt{m})$ with high probability. Applying triangle inequality and using the definition of fake gradient (see Definition E.1), we finish the proof of the first statement.

As for the second statement, we replace the use of Lemma B.11 with Corollary B.19. □

The rest of this section is devoted to proving a (much more involved) lower bound on this fake gradient.

**Theorem 5** (gradient lower bound at random init, restated)**.** *Letting $\rho = nLd\log m$, given fixed vectors $\{\mathsf{loss}_{i,\ell} \in \mathbb{R}^d\}_{i \in [n], \ell \in [L]\setminus\{1\}}$, with probability at least $1 - e^{-\Omega(\rho^2)}$ over $W, A, B$, we have*

$$\|\widehat{\nabla} f(W)\|_F^2 \geq \Omega\left(\frac{\delta}{\rho^{14}}\right) \times m \times \max_{i,\ell}\{\|\mathsf{loss}_{i,\ell}\|^2\} .$$

In the rest of this section, we first present a elegant way to decompose the randomness in Section E.1 (motivated by smooth analysis [60]). We give a lower bound on the *expected* fake gradient in Section E.2. We then calculate the stability of fake gradient against rank-one perturbations in Section E.3. Finally, in Section E.4, we apply our extended McDiarmid's inequality to prove Theorem 5.

### E.1 Randomness Decomposition

We introduce a parameter $\theta$ as follows:

**Definition E.3.** *Choose any*

$$\boxed{\theta \in \left[\rho^4 \cdot \beta_-, \rho^{-3} \cdot \beta_+\right]}$$

*where the notions $\beta_-$ and $\beta_+$ come from Definition D.1. We have $\theta \leq \frac{\delta}{\rho}$.*

In the next lemma, we introduce a way to decompose the randomness of $W$ into $W = W_2 + W'$ for two corrollated random matrices $W_2$ and $W'$. We will make sure that the entries of $W_2$ are also i.i.d. from $N(0, \frac{2}{m})$. This definition requires us to choose a specific pair $(i^*, \ell^*) \in [n] \times [L]$.

**Lemma E.4.** *Suppose $W$ and $A$ follow from random initialization. Given any fixed $i^* \in [n]$, $\ell^* \in [L]$, and parameter $\theta \in \left(0, \frac{1}{2\rho}\right]$, letting random vector $v_{i^*,\ell^*} \in \mathbb{R}^m$ be defined as*

$$v_{i^*,\ell^*} \stackrel{\text{def}}{=} \frac{(I - U_{\ell^*-1}U_{\ell^*-1}^\top)h_{i^*,\ell^*}}{\|(I - U_{\ell^*-1}U_{\ell^*-1}^\top)h_{i^*,\ell^*}\|_2} \enspace ,$$

*then we can write $W = W_2 + W'$ for two random matrices $W_2 \in \mathbb{R}^{m \times m}$ and $W' \in \mathbb{R}^{m \times m}$ where*

(a) *entries of $W_2$ are i.i.d. drawn from $\mathcal{N}(0, \frac{2}{m})$ (so $W_2$ is in the same distribution as $W$);*

(b) *$W'$ is correlated with $W_2$ and can be written as $W' = uv_{i^*,\ell^*}^\top$ for some $u \in \mathbb{R}^m$;*

(c) *With probability at least $1 - e^{-\Omega(\rho^2)}$ over the randomness of $W_2$:*

$$\forall k: \quad \Pr_{W'}\left[u_k = (W'v_{i^*,\ell^*})_k \geq \frac{\theta}{2\sqrt{m}} \,\bigg|\, W_2\right] \geq \frac{1}{4} \quad \text{and}$$

$$\Pr_{W'}\left[u_k = (W'v_{i^*,\ell^*})_k \leq -\frac{\theta}{2\sqrt{m}} \,\bigg|\, W_2\right] \geq \frac{1}{4}$$

(d) *With probability at least $1 - e^{-\Omega(\rho^2)}$ over the randomness of $W_2$:*

$$\Pr_{W'}\left[\|u\|_\infty = \|W'v_{i^*,\ell^*}\|_\infty \leq \frac{3\theta\rho}{2\sqrt{m}} \,\bigg|\, W_2\right] \geq 1 - e^{-\Omega(\rho^2)} \enspace .$$

*Proof.* Since $v_{i^*,\ell^*}$ only depends on the randomness of $WU_{\ell^*-1}$, we have $Wv_{i^*,\ell^*} \sim \mathcal{N}(0, \frac{2I}{m})$. As a result, letting $g_1, g_2 \in \mathbb{R}^m$ be two independent random Gaussian vectors sampled from $\mathcal{N}(0, \frac{2I}{m})$, we can couple them to make sure

$$Wv_{i^*,\ell^*} = \sqrt{1-\theta^2}g_1 + \theta g_2 \enspace .$$

We define $W_2, W' \in \mathbb{R}^{n \times n}$ as follows

$$W_2 \stackrel{\text{def}}{=} WU_{\ell^*-1}U_{\ell^*-1}^\top + g_1 v_{i^*,\ell^*}^\top + W(I - U_{\ell^*-1}U_{\ell^*-1}^\top)(I - v_{i^*,\ell^*}v_{i^*,\ell^*}^\top)$$

$$W' \stackrel{\text{def}}{=} uv_{i^*,\ell^*}^\top \quad \text{where} \quad u \stackrel{\text{def}}{=} \theta g_2 - (1 - \sqrt{1-\theta^2})g_1$$

- It is easy to verify that

$$W_2 + W' = W(I - v_{i^*,\ell^*}v_{i^*,\ell^*}^\top) + g_1 v_{i^*,\ell^*}^\top + (\theta g_2 - (1-\sqrt{1-\theta^2})g_1)v_{i^*,\ell^*}^\top$$

$$= W(I - v_{i^*,\ell^*}v_{i^*,\ell^*}^\top) + (\theta g_2 + \sqrt{1-\theta^2}g_1)v_{i^*,\ell^*}^\top$$

$$= W(I - v_{i^*,\ell^*}v_{i^*,\ell^*}^\top) + Wv_{i^*,\ell^*}v_{i^*,\ell^*}^\top$$

$$= W.$$

- We verify that the entries of $W_2$ are i.i.d. Gaussian in two steps.

  On one hand, let $U = [\widehat{h}_1, \ldots, \widehat{h}_m] \in \mathbb{R}^{m \times m}$ be an arbitrary orthonormal matrix where its first $nL$ columns are identical to $U_L \in \mathbb{R}^{m \times nL}$. For each $k \in [nL]$, although $\widehat{h}_k$ may depend on the randomness of $W = W_2 + W'$, it can only depend on $W[\widehat{h}_1, \ldots, \widehat{h}_{k-1}]$. Therefore, conditioning on the randomness of $W_2[\widehat{h}_1, \ldots, \widehat{h}_{k-1}]$, we have that $W_2\widehat{h}_k$ is still $\mathcal{N}(0, \frac{2I}{m})$. This shows that all entries of $W_2U$ still follow from $N(0, \frac{2}{m})$.

  On the other hand, let $V = W_2U = [v_1, \ldots, v_m] \in \mathbb{R}^{m \times m}$, then we have $v_k = W_2\widehat{h}_k$ is independent of $\widehat{h}_k$ as argued above. Since each $v_k \sim \mathcal{N}(0, \frac{2I}{m})$, we have that all entries of $VU^\top = \sum_k v_k \widehat{k}_k^\top$ are i.i.d. from $N(0, \frac{2}{m})$. Since $W_2 = VU^\top$ this concludes the proof.

- By standard Gaussian tail bound, we have with probability at least $1 - e^{-\Omega(\rho^2)}$, it satisfies $\|g_1\|_\infty \leq \frac{\rho}{\sqrt{m}}$ and therefore $\|(1 - \sqrt{1-\theta^2})g_1\|_\infty \leq \frac{\theta^2 \rho}{\sqrt{m}}$. If this happens, for every $k \in [m]$, we have (see for instance Fact A.12) that $\mathbf{Pr}_{g_2}[(\theta g_2)_k > \frac{\theta}{\sqrt{m}}] \geq \frac{1}{4}$ and $\mathbf{Pr}_{g_2}[(\theta g_2)_k < -\frac{\theta}{\sqrt{m}}] \geq \frac{1}{4}$. Recalling
$$u = (\theta g_2 - (1 - \sqrt{1 - \theta^2})g_1 \ .$$
and using the fact that $\theta \leq \frac{1}{2\rho}$, we conclude that
$$\mathbf{Pr}_{W'}\left[u_k > \frac{\theta}{2\sqrt{m}} \,\Big|\, W_2, A\right] \geq \frac{1}{4} \quad \text{and} \quad \mathbf{Pr}_{W'}\left[u_k < -\frac{\theta}{2\sqrt{m}} \,\Big|\, W_2, A\right] \geq \frac{1}{4} \ .$$

- Again by Gaussian tail bound, we have with probability at least $1 - e^{-\Omega(\rho^2)}$, it satisfies $\|g_1\|_\infty, \|g_2\|_\infty \leq \frac{\rho}{\sqrt{m}}$. Therefore, $\|(1 - \sqrt{1-\theta^2})g_1\|_\infty \leq \frac{\theta^2 \rho}{\sqrt{m}}$ and using our assumption on $\theta$, we have
$$\|u\|_\infty = \left\|(\theta g_2 - (1 - \sqrt{1 - \theta^2})g_1\right\|_\infty \leq \frac{3\theta \rho}{2\sqrt{m}} \ .$$

$\square$

**Lemma E.5.** *Given any fixed $i^* \in [n]$, $\ell^* \in [L]$, and parameter $\theta \in \left(0, \frac{1}{2\rho}\right]$, and suppose suppose $W$ and $A$ are at random initialization and $W = W_2 + W'$ where $W' = uv_{i^*,\ell^*}^\top$ (following Lemma E.4). Now, we introduce another two ways of decomposing randomness based on this definition.*

(a) *Given fixed $k \in [m]$, we can write $W = W_0 + W'_k$ where $W'_k = u_k v_{i^*,\ell^*}^\top$ for $u_k = (0, \ldots, 0, u_k, 0, \ldots, 0)$ that is non-zero only at the $k$-th coordinate. We again have the entries of $W_0$ are i.i.d. from $N(0, \frac{2}{m})$.*

(b) *Given fixed $N \subseteq [m]$, we can write $W = W_1 + W'_N$ where $W'_N = u_N v_{i^*,\ell^*}^\top$ for $u_N$ being the projection of $u$ onto coordinates in $N$ (so $\|u\|_0 = |N|$). We again have the entries of $W_0$ are i.i.d. from $N(0, \frac{2}{m})$.*

(c) *Given fixed $N \subseteq [m]$, we can write $W = W_2 + W'_N + W'_{-N}$ where $W'_N = u_N v_{i^*,\ell^*}^\top$ and $W'_{-N} = u_{-N} v_{i^*,\ell^*}^\top$. Here, $u_{-N}$ is the projection of $u$ onto coordinates in $[m] \setminus N$.*

*Proof.* The proofs are analogous to Lemma E.4. $\square$

### E.2 Gradient Lower Bound in Expectation

The goal of this subsection is to prove Lemma E.6 and then translate it into our Core Lemma A (see Lemma E.7).

**Lemma E.6** (c.f. (6.6)). *Let $\rho = nLd \log m$ and $\varrho = nLd\delta^{-1}\log(m/\varepsilon)$. Fix integer $N \in [\rho^6/\beta_-, \varrho^{100}]$, $\theta$ from Definition E.3, fix vectors $\{\text{loss}_{i,\ell} \in \mathbb{R}^d\}_{i\in[n],\ell\in[L]\setminus\{1\}}$, let $(i^*, \ell^*) = \arg\min_{i,\ell} \{\|\text{loss}_{i,\ell}\|\}$. Suppose $W_1, A, B$ are at random initialization, and suppose $W'_N$ is defined according to Lemma E.5b so that $W = W_1 + W'_N$ is also at random initialization. With probability at least $1 - e^{-\Omega(\rho^2)}$ over the randomness of $W_1, A, B$, we have*
$$\left|\left\{k \in N \,\Big|\, \mathbb{E}_{W'_N}\left[\|\widehat{\nabla}_k f(W_1 + W'_N)\|_2\right] \geq \Omega\Big(\frac{\|\text{loss}_{i^*,\ell^*}\|_2}{nL\sqrt{d}}\Big)\right\}\right| \geq \frac{\beta_-|N|}{100L}$$

### E.2.1 Proof of Lemma E.6

Throughout this proof we denote by $\widetilde{W} = W_1$. We introduce some (old and new) notations:

$$\widetilde{\mathsf{Back}}_{i,\ell \to \ell} = B, \qquad\qquad \mathsf{Back}_{i,\ell \to \ell} = B$$
$$\widetilde{\mathsf{Back}}_{i,\ell \to a} = B\widetilde{D}_{i,a}\widetilde{W}\cdots \widetilde{D}_{i,\ell+1}\widetilde{W}, \qquad \mathsf{Back}_{i,\ell \to a} = BD_{i,a}W\cdots D_{i,\ell+1}W,$$

and we let $\mathsf{Back}'_{i,\ell \to a} = \mathsf{Back}_{i,\ell \to a} - \widetilde{\mathsf{Back}}_{i,\ell \to a}$. For each $i \in [n]$, $\ell \in [L]$, we define

$$\widetilde{g}_{i,\ell} = Ax_{i,\ell} + \widetilde{W}\widetilde{D}_{i,\ell-1}\widetilde{g}_{i,\ell-1} \qquad g_{i,\ell} = Ax_{i,\ell} + WD_{i,\ell-1}g_{i,\ell-1} \qquad g'_{i,\ell} = g_{i,\ell} - \widetilde{g}_{i,\ell}$$
$$\widetilde{h}_{i,\ell} = \widetilde{D}_{i,\ell}(Ax_{i,\ell} + \widetilde{W}\widetilde{h}_{i,\ell-1}) \qquad h_{i,\ell} = D_{i,\ell}(Ax_{i,\ell} + Wh_{i,\ell-1}) \qquad h'_{i,\ell} = h_{i,\ell} - \widetilde{h}_{i,\ell}$$
$$\widetilde{u}_{i,\ell} = \sum_{a=\ell}^{L} \widetilde{\mathsf{Back}}_{\ell \to a}^{\top} \mathsf{loss}_{i,a} \qquad u_{i,\ell} = \sum_{a=\ell}^{L} \mathsf{Back}_{\ell \to a}^{\top} \mathsf{loss}_{i,a} \qquad u'_{i,\ell} = u_{i,\ell} - \widetilde{u}_{i,\ell}$$

We also let

$$v_{i^*,\ell^*} = \frac{(I - U_{\ell^*-1}U_{\ell^*-1}^{\top})\widetilde{h}_{i^*,\ell^*}}{\|(I - U_{\ell^*-1}U_{\ell^*-1}^{\top})\widetilde{h}_{i^*,\ell^*}\|_2}$$

and recall from Lemma E.4d, we have

$$W'_N = yv_{i^*,\ell^*}^{\top} \text{ for some } y \in \mathbb{R}^m \text{ with } \|y\|_0 = |N| \text{ and } \|y\|_\infty \leq \frac{\tau_0}{\sqrt{m}} \stackrel{\text{def}}{=} \frac{3\theta\rho}{2\sqrt{m}} \in \left[\frac{\varrho^{-20}}{\sqrt{m}}, \frac{\varrho^{20}}{\sqrt{m}}\right]$$

*Proof of Lemma E.6.* We let $N_1 = N$, apply Lemma D.2 to obtain $N_4 \subseteq N_1$ with $|N_4| \geq \frac{\beta_-|N_1|}{64L}$, and apply Lemma D.7 to obtain $N_5 \subseteq N_4$. According to the statements of these lemmas, we know that with probability at least $1 - e^{-\Omega(\rho^2)}$, the random choice of $W_1, A, B$ will satisfy $|N_5| \geq \frac{\beta_-|N_1|}{100L}$.

Fixing such $k \in N_5$, we use $u_{i,\ell}$ to denote $(u_{i,\ell})_k$ and $g_{i,\ell}$ to denote $(g_{i,\ell})_k$ for notational simplicity. We can rewrite $\widehat{\nabla}_k f(W)$ as follows and apply triangle inequality:

$$\|\widehat{\nabla}_k f(W)\| = \left\| u_{i^*,\ell^*} \cdot h_{i^*,\ell^*} \cdot \mathbf{1}_{g_{i^*,\ell^*+1} \geq 0} + \sum_{(i,\ell) \neq (i^*,\ell^*)} u_{i,\ell} \cdot h_{i,\ell} \cdot \mathbf{1}_{g_{i,\ell+1} \geq 0} \right\|$$
$$\geq \underbrace{\left\| \widetilde{u}_{i^*,\ell^*} \cdot \widetilde{h}_{i^*,\ell^*} \cdot \mathbf{1}_{g_{i^*,\ell^*+1} \geq 0} + \sum_{(i,\ell) \neq (i^*,\ell^*)} \widetilde{u}_{i,\ell} \cdot \widetilde{h}_{i,\ell} \cdot \mathbf{1}_{g_{i,\ell+1} \geq 0} \right\|_2}_{\clubsuit}$$
$$- \underbrace{\sum_{i,\ell} \left\| u'_{i,\ell} \cdot \widetilde{h}_{i,\ell} + \widetilde{u}_{i,\ell} \cdot h'_{i,\ell} + u'_{i,\ell} \cdot h'_{i,\ell} \right\|_2}_{\spadesuit} \qquad\qquad (E.1)$$

where recall $g_{i,\ell+1} = \langle W_k, h_{i,\ell}\rangle + \langle A_k, x_{i,\ell+1}\rangle$.

**Step 1.** To bound the $\spadesuit$ term on the RHS of (E.1), we consider the following bounds:

1. Using Lemma C.11, we have with probability at least $1 - e^{-\Omega(\rho^2)}$, for all $i, \ell, a$

$$|(\widetilde{\mathsf{Back}}_{i,\ell \to a} \cdot \mathsf{loss}_{i,a})_k - (\mathsf{Back}_{i,\ell \to a} \cdot \mathsf{loss}_{i,a})_k| \leq O\left(\frac{\rho^3 \tau_0^{1/3} N^{1/6}}{m^{1/6}}\right) \cdot \|\mathsf{loss}_{i,a}\| \ .$$

This implies, by triangle inequality and the fact that $\|\mathsf{loss}_{i,a}\| \leq \|\mathsf{loss}_{i^*,a^*}\|$,

$$|u'_{i,\ell}| = |u_{i,\ell} - \widetilde{u}_{i,\ell}| \leq O\left(\frac{\rho^4 \tau_0^{1/3} N^{1/6}}{m^{1/6}}\right) \cdot \|\mathsf{loss}_{i^*,a^*}\| \ .$$

2. Using Corollary B.19, we have with probability at least $1 - e^{-\Omega(\rho^2)}$, for all $i, \ell, a$

$$|(\widetilde{\mathsf{Back}}_{i,\ell \to a} \cdot \mathsf{loss}_{i,a})_k| \leq \rho \|\mathsf{loss}_{i,a}\| \ .$$

This implies, by triangle inequality and the fact that $\|\mathsf{loss}_{i,a}\| \leq \|\mathsf{loss}_{i^*,a^*}\|$,
$$|\widetilde{u}_{i,\ell}| \leq \rho^2 \|\mathsf{loss}_{i^*,a^*}\| \ .$$

3. Using Lemma B.3 and Lemma C.2a we have
$$\|\widetilde{h}_{i,\ell}\|_2 \leq O(L) \quad \text{and} \quad \|h'_{i,\ell}\|_2 \leq O(L^6 \tau_0 \sqrt{N})/m^{1/2}$$

All together, they imply
$$\sum_{i,\ell} \left\| u'_{i,\ell} \cdot \widetilde{h}_{i,\ell} + \widetilde{u}_{i,\ell} \cdot h'_{i,\ell} + u'_{i,\ell} \cdot h'_{i,\ell} \right\|_2 \leq O\Big(\frac{\rho^6 \tau_0^{1/3} N^{1/6}}{m^{1/6}}\Big) \cdot \|\mathsf{loss}_{i^*,a^*}\| \ . \tag{E.2}$$

**Step 2.** We next turn to the ♣ term on the RHS of (E.1). We divide into three cases to analyze the sign change of $g_{i,\ell+1}$ for all possible $i$ and $\ell$ in the above formula (under the randomness of $W'_N$). Before we do so, first note that Lemma C.10 implies
$$|(Wh'_{i,\ell})_k| = |(W_1 + W'_N)h'_{i,\ell}| \leq O\Big(\frac{\rho^8 N^{2/3} \tau_0^{5/6}}{m^{2/3}}\Big) \tag{E.3}$$

1. Consider $g_{i,\ell+1}$ for the case of $i = i^*, \ell = \ell^*$. We want to show that the sign of $g_{i,\ell+1}$ changes (at least with constant probability) with respect to the randomness of $W'_N$. One can easily check that:[20]
$$g_{i,\ell+1} = \langle A_k, x_{i,\ell+1} \rangle + (W_1 \widetilde{h}_{i,\ell})_k + (Wh'_{i,\ell})_k + (W'_N v_{i,\ell})_k \cdot \|(I - U_{\ell-1} U_{\ell-1}^\top) h_{i,\ell}\|_2. \tag{E.4}$$

By Lemma D.2, we have $|\langle A_k, x_{i,\ell} \rangle + (W_1 \widetilde{h}_{i,\ell})_k| \leq \frac{\beta_-}{\sqrt{m}}$ Thus, putting this and (E.3) into (E.4), we can write
$$g_{i,\ell+1} = \Xi + (W'_N v_{i,\ell})_k \cdot \|(I - U_{\ell-1} U_{\ell-1}^\top) \widetilde{h}_{i,\ell}\|_2$$
for some value $\Xi$ with $|\Xi| \leq \frac{2\beta_-}{\sqrt{m}}$. By Lemma B.6, we have $\|(I - U_{\ell-1} U_{\ell-1}^\top) \widetilde{h}_{i,\ell}\|_2 \geq \Omega\big(\frac{1}{L^2 \log^3 m}\big)$. By Lemma E.4c, we have $(W'_N v_{i,\ell})_k$ shall be larger than $\frac{\theta}{2\sqrt{m}}$ and smaller than $-\frac{\theta}{2\sqrt{m}}$ each with probability at least $1/4$. Since by our choice of $\theta$ (see Definition E.3),
$$\frac{\theta}{2\sqrt{m}} \cdot \Omega\Big(\frac{1}{L^2 \log^3 m}\Big) \gg \frac{2\beta_-}{\sqrt{m}} \geq |\Xi|$$

This concludes that, with probability at least $1 - e^{-\Omega(\rho^2)}$ over $W_1$ and $A$, it satisfies
$$\Pr_{W'_N}[g_{i,\ell+1} > 0 \mid W_1, A] \geq \frac{1}{5} \quad \text{and} \quad \Pr_{W'_N}[g_{i,\ell+1} < 0 \mid W_1, A] \geq \frac{1}{5}$$

---

[20] We can write $(g_{i,\ell+1})_k = \langle A_k, x_{i,\ell+1} \rangle + (W(\widetilde{h}_{i,\ell} + h'_{i,\ell}))_k = \langle A_k, x_{i,\ell+1} \rangle + (W\widetilde{h}_{i,\ell})_k + (Wh'_{i,\ell})_k$. For the second term in the above equation, we have

$$(W\widetilde{h}_{i,\ell})_k = \left(WU_{\ell-1}U_{\ell-1}^\top \widetilde{h}_{i,\ell} + \frac{W(I - U_{\ell-1}U_{\ell-1}^\top)\widetilde{h}_{i,\ell}}{\|(I - U_{\ell-1}U_{\ell-1}^\top)\widetilde{h}_{i,\ell}\|_2} \|(I - U_{\ell-1}U_{\ell-1}^\top)\widetilde{h}_{i,\ell}\|_2 \right)_k$$

$$\overset{①}{=} \left(WU_{\ell-1}U_{\ell-1}^\top \widetilde{h}_{i,\ell} + Wv_{i,\ell}\|(I - U_{\ell-1}U_{\ell-1}^\top \widetilde{h}_{i,\ell})\|_2 \right)_k$$

$$\overset{②}{=} \left(W_1 U_{\ell-1}U_{\ell-1}^\top \widetilde{h}_{i,\ell} + Wv_{i,\ell}\|(I - U_{\ell-1}U_{\ell-1}^\top \widetilde{h}_{i,\ell})\|_2 \right)_k$$

$$\overset{③}{=} \left(W_1 U_{\ell-1}U_{\ell-1}^\top \widetilde{h}_{i,\ell} + W_1 v_{i,\ell}\|(I - U_{\ell-1}U_{\ell-1}^\top \widetilde{h}_{i,\ell})\|_2 \right)_k + \left(W'_N v_{i,\ell} \cdot \|(I - U_{\ell-1}U_{\ell-1}^\top \widetilde{h}_{i,\ell})\|_2 \right)_k$$

$$= (W_1 \widetilde{h}_{i,\ell})_k + (W'_N v_{i,\ell})_k \cdot \|(I - U_{\ell-1}U_{\ell-1}^\top \widetilde{h}_{i,\ell})\|_2$$

where ① follows by definition of $v_{i,\ell}$, ② follows by definition of $W_1$, ③ follows by $W = W_1 + W'_N$.

2. Consider $g_{i,\ell+1}$ for the case of $i \in [n]\setminus\{i^*\}, \ell = \ell^*$ or $i \in [n], \ell > \ell^*$. We want to show the sign of $g_{i,\ell+1}$ is fixed, meaning with high probability independent of the choice of $W'_N$. One can easily check that:

$$g_{i,\ell+1} = \langle A_k, x_{i,\ell+1}\rangle + (W_1\widetilde{h}_{i,\ell})_k + (Wh'_{i,\ell})_k + (W'_N\widetilde{h}_{i,\ell})_k \ . \tag{E.5}$$

By Lemma D.2, we have $|\langle A_k, x_{i,\ell}\rangle + (W_1\widetilde{h}_{i,\ell})_k| \geq \frac{\beta_+}{\sqrt{m}}$. Thus, putting this and (E.3) into (E.6), we can write

$$g_{i,\ell+1} = \Xi + (W'_N\widetilde{h}_{i,\ell})_k = \Xi + (W'_N v_{i^*,\ell^*})_k \cdot \langle v_{i^*,\ell^*}, \widetilde{h}_{i,\ell}\rangle$$

for some value $\Xi$ with $|\Xi| \geq \frac{\beta_+}{2\sqrt{m}}$. By Lemma B.3, we have $|\langle v_{i^*,\ell^*}, \widetilde{h}_{i,\ell}\rangle| \leq \|\widetilde{h}_{i,\ell}\|_2 \leq O(L)$. By Lemma E.4d, we have $|(W'_N v_{i^*,\ell^*})_k| \leq \frac{2\theta\rho}{\sqrt{m}}$ and by our choice of $\theta$ (see Definition E.3),

$$\frac{2\theta\rho}{\sqrt{m}} \cdot O(L) \ll \frac{2\beta_+}{\sqrt{m}} \leq |\Xi|$$

This concludes that, with probability at least $1 - e^{-\Omega(\rho^2)}$ over $W_1$ and $A$, it satisfies

$$\exists s \in \{-1,1\}\colon \quad \Pr_{W'_N}[\operatorname{sgn}(g_{i,\ell+1}) = s \mid W_1, A] \geq 1 - e^{-\Omega(\rho^2)} \ .$$

3. Consider $g_{i,\ell+1}$ for the case of $i \in [n]$ and $\ell < \ell^*$. We want to show such $g_{i,\ell+1}$ is fixed, meaning independent of $W'_N$. One can easily check that:

$$g_{i,\ell+1} = \langle A_k, x_{i,\ell+1}\rangle + (W_1\widetilde{h}_{i,\ell})_k + (Wh'_{i,\ell})_k + (W'_N\widetilde{h}_{i,\ell})_k \ . \tag{E.6}$$

Now, since $\ell < \ell^*$, we know $h_{i,\ell} = \widetilde{h}_{i,\ell}$ because both of which only depend on the randomness of $WU_{\ell^*-1} = W_1 U_{\ell^*-1}$ and $A$; this implies $h'_{i,\ell} = 0$. Also, since $W'_N = (W'_N v_{i^*,\ell^*})v_{i^*,\ell^*}^\top$ but $\langle v_{i^*,\ell^*}, \widetilde{h}_{i,\ell}\rangle = 0$, we know that

$$g_{i,\ell+1} = \langle A_k, x_{i,\ell+1}\rangle + (W_1\widetilde{h}_{i,\ell})_k \ .$$

This is a fixed value, independent of $W'_N$.

Therefore, combining the three cases above, we conclude that with at least constant probability, the sign of $g_{i^*,\ell^*+1}$ can possibly change, but the sign of $g_{i,\ell+1}$ will not change for any other $(i,\ell) \neq (i^*,\ell^*)$. This means,

$$\mathbb{E}[\clubsuit] = \mathbb{E}\left[\left\|\widetilde{u}_{i^*,\ell^*} \cdot \widetilde{h}_{i^*,\ell^*} \cdot \mathbf{1}_{g_{i^*,\ell^*+1}\geq 0} + \sum_{(i,\ell)\neq(i^*,\ell^*)} \widetilde{u}_{i,\ell} \cdot \widetilde{h}_{i,\ell} \cdot \mathbf{1}_{g_{i,\ell+1}\geq 0}\right\|_2\right]$$
$$\geq \Omega(\|\widetilde{u}_{i^*,\ell^*}\widetilde{h}_{i^*,\ell^*}\|_2) = \Omega(|\widetilde{u}_{i^*,\ell^*}|\cdot\|\widetilde{h}_{i^*,\ell^*}\|_2)$$

Finally, for each $k \in N_5$, owing to Lemma D.7 and Lemma B.3, we have

$$|\widetilde{u}_{i^*,\ell^*}|_2 = \left|\sum_{a=\ell^*}^{L}(\widetilde{\operatorname{Back}}_{i^*,\ell^*\to a}^\top \operatorname{loss}_{i^*,a})_k\right| \geq \frac{\|\operatorname{loss}_{i^*,\ell^*}\|_2}{6\sqrt{d}nL} \quad \text{and} \quad \|\widetilde{h}_{i^*,\ell^*}\|_2 \geq \frac{1}{2} \ .$$

This means,

$$\mathbb{E}[\clubsuit] \geq \Omega\left(\frac{\|\operatorname{loss}_{i^*,\ell^*}\|_2}{\sqrt{d}nL}\right) \ . \tag{E.7}$$

Putting (E.2) and (E.7) back to (E.1), we conclude that

$$\forall k \in N_5\colon \quad \mathbb{E}[\|\widehat{\nabla}_k f(W)\|] \geq \Omega\left(\frac{\|\operatorname{loss}_{i^*,\ell^*}\|_2}{\sqrt{d}nL}\right) \ .$$

$\square$

### E.2.2 Core Lemma A

In Lemma E.6, we split $W$ into $W = W_1 + W'_N$ for a fixed $N \subseteq [m]$. In this subsection, we split $W$ into three parts $W = W_2 + W'_N + W'_{-N}$ following Lemma E.5c. Our purpose is to rewrite Lemma E.6 into the following variant:

> **Lemma E.7** (Core Lemma A). *Let $\rho = nLd \log m$ and $\varrho = nLd\delta^{-1} \log(m/\varepsilon)$. Fix vectors $\{\mathsf{loss}_{i,\ell} \in \mathbb{R}^d\}_{i \in [n], \ell \in [L] \setminus \{1\}}$, integer $|N| \in [\rho^6/\beta_-, \varrho^{100}]$, and parameter $t = m/|N|$. Let $N_1, \ldots, N_t$ be i.i.d. random subsets of $[m]$ with cardinality $|N_i| = |N|$. Then,*
> - *with probability $\geq 1 - e^{-\Omega(\rho^2)}$ over the randomness of $W_2, A, B$ and $N_1, \cdots N_t$, the following holds:*
>   - *for every $N \in \{N_1, N_2, \cdots, N_t\}$, the following holds:*
>     * *with probability at $1 - e^{-\Omega(\rho^2)}$ over the randomness of $W'_{-N}$, the following holds:*
>
> $$\sum_{k \in N} \mathop{\mathbb{E}}_{W'_N} \left[\|\widehat{\nabla}_k f(W_2 + W'_N + W'_{-N})\|_2^2\right] \geq \Omega\left(\frac{\beta_- |N|}{\rho^2}\right) \cdot \max_{i,\ell} \|\mathsf{loss}_{i,\ell}\|_2^2.$$

*Proof of Lemma E.7.* We denote by $(i^*, \ell^*) = \arg\max_{i,\ell} \|\mathsf{loss}_{i,\ell}\|_2$. We first note that Lemma E.6 directly implies

- with probability $\geq 1 - e^{-\Omega(\rho^2)}$ over the randomness of $W_2, W'_{-N}, A, B, N$, the following holds:

$$\left|\left\{k \in N \;\middle|\; \mathop{\mathbb{E}}_{W'_N}\left[\|\widehat{\nabla}_k f(W_2 + W'_{-N} + W'_N)\|_2\right] \geq \Omega\left(\frac{\|\mathsf{loss}_{i^*,\ell^*}\|_2}{nL\sqrt{d}}\right)\right\}\right| \geq \frac{\beta_-}{100L}|N|$$

or putting it in another way, using our choice of $q$,

- with probability $\geq 1 - e^{-\Omega(\rho^2)}$ over the randomness of $W_2, W'_{-N}, A, B, N$, the following holds ($q = \frac{\beta_- |N|}{c\rho^2}$ for some sufficiently large constant):

$$\sum_{k \in N} \mathop{\mathbb{E}}_{W'_N} \left[\|\widehat{\nabla}_k f(W_2 + W'_{-N} + W_N)\|_2^2\right] \geq q \cdot \|\mathsf{loss}_{i^*,\ell^*}\|_2^2.$$

Applying simple tricks to switch the ordering of randomness (see Fact A.2), we have

- with probability $\geq 1 - e^{-\Omega(\rho^2)}$ over the randomness of $W_2, A, B$, the following holds:
  - with probability $\geq 1 - e^{-\Omega(\rho^2)}$ over the randomness of $N$, the following holds:
    * with probability $\geq 1 - e^{-\Omega(\rho^2)}$ over randomness of $W'_{-N}$, the following holds:

$$\sum_{k \in N} \mathop{\mathbb{E}}_{W'_N} [\|\widehat{\nabla}_k f(W_2 + W'_N + W'_{-N})\|_2^2] \geq q \cdot \|\mathsf{loss}_{i^*,\ell^*}\|_2^2.$$

Repeating the choice of $N$ for $t$ times, and using union bound, we have

- with probability $\geq 1 - e^{-\Omega(\rho^2)}$ over the randomness of $W_2, A, B$, the following holds:
  - with probability $\geq 1 - e^{-\Omega(\rho^2)}$ over the randomness of $N_1, N_2, \cdots, N_t$, the following holds:
    * for all $N \in \{N_1, N_2, \cdots, N_t\}$, the following holds:
      · with probability $\geq 1 - e^{-\Omega(\rho^2)}$ over randomness of $W'_{-N}$, the following holds:

$$\sum_{k \in N} \mathop{\mathbb{E}}_{W'_N} [\|\widehat{\nabla}_k f(W_2 + W'_N + W'_{-N})\|_2^2] \geq q \cdot \|\mathsf{loss}_{i^*,\ell^*}\|_2^2.$$

Combining the above statement with Fact A.1 (to merge randomness), we conclude the proof. □

## E.3 Gradient Stability

The goal of this subsection is to prove Lemma E.8 and then translate it into our Core Lemma B (see Lemma E.10).

**Lemma E.8** (c.f. (6.7)). *Fix parameter $\rho = nLd \log m$, vectors $\{\mathsf{loss}_{i,\ell} \in \mathbb{R}^d\}_{i \in [n], \ell \in [L] \setminus \{1\}}$, $(i^*, \ell^*) = \arg\min_{i,\ell} \{\|\mathsf{loss}_{i,\ell}\|\}$, index $j \in [m]$. Let $N \subseteq [m]$ be a random subset containing $j$ of fixed cardinality $|N|$, let $W, A, B$ be at random initialization, and $W_j''$ be defined as Definition E.5a so that $W + W_j''$ is also at random initialization. Then, with probability at least $1 - e^{-\Omega(\rho^2)}$*

$$\sum_{k \in [N]} \left| \|\widehat{\nabla}_k(W)\|_2^2 - \|\widehat{\nabla}_k(W + W_j'')\|_2^2 \right| \leq O\left(\rho^8 + \frac{\rho^{11} \theta^{1/3}}{m^{1/6}} |N|\right) \ . \tag{E.8}$$

### E.3.1 Proof of Lemma E.8

Throughout this proof we denote by $\widetilde{W} = W$ to emphasize that $W$ is at random initialization. We first show some properties before proving Lemma E.8.

**Claim E.9.** *Fix parameter $\rho = nLd \log m$, vectors $\{\mathsf{loss}_{i,\ell} \in \mathbb{R}^d\}_{i \in [n], \ell \in [L] \setminus \{1\}}$, $(i^*, \ell^*) = \arg\min_{i,\ell} \{\|\mathsf{loss}_{i,\ell}\|\}$, index $j \in [m]$. Let $\widetilde{W}, A, B$ be at random initialization, and $W_j''$ be defined as Lemma E.5a so that $W = \widetilde{W} + W_j''$ is also at random initialization. Then, then with probability $\geq 1 - e^{-\Omega(\rho^2)}$ over $\widetilde{W}, A, B, W_j''$,*

- $\left|\left\{ k \in [m] \ \middle| \ \left|\|\widehat{\nabla}_k f(\widetilde{W} + W_j'')\|_2^2 - \|\widehat{\nabla}_k f(\widetilde{W})\|_2^2\right| \leq O\left(\frac{\rho^{11} \theta^{1/3}}{m^{1/6}}\right) \right\}\right| \geq \left(1 - \frac{\rho^5 \theta^{2/3}}{m^{1/3}}\right) m \ .$

- $\left|\|\widehat{\nabla}_k f(\widetilde{W} + W_j'')\|_2^2 - \|\widehat{\nabla}_k f(\widetilde{W})\|_2^2\right| \leq O(\rho^6)$ *for every $k \in [m]$.*

*Proof of Claim E.9.* As before, let $\widetilde{D}_{i,\ell}$ denote the diagonal matrix consisting of the indicator function $\mathbb{1}_{(\widetilde{g}_{i,\ell+1})_k \geq 0}$ for its $k$-th diagonal entry. Let $D_{i,\ell}$ be the same thing but with respect to weight matrix $\widetilde{W} + W_j''$. Let $D_{i,\ell}'' = D_{i,\ell} - \widetilde{D}_{i,\ell}$ be the diagonal matrix of sign change.

- By Lemma C.2 (with $\tau_0 = \frac{3\theta\rho}{2}$ owing to Lemma E.4d), we have $\|D_\ell''\|_0 \leq \rho^4 \theta^{2/3} m^{2/3}$. Therefore, letting $J \overset{\text{def}}{=} [m] \setminus \cup_{i,\ell} \mathrm{supp}(D_{i,\ell})$ be the set of all indices of sign changes, we have

$$|J| \geq m - nL\rho^4 \theta^{2/3} m^{2/3} \geq \left(1 - \frac{\rho^5 \theta^{2/3}}{m^{1/3}}\right) \cdot m.$$

Next, for every index $k \in J$, we have

$$\|\widehat{\nabla}_k f(\widetilde{W} + W_j'') - \widehat{\nabla}_k f(\widetilde{W})\|_2$$

$$\overset{\text{①}}{=} \left\| \sum_{i=1}^n \sum_{\ell=1}^L (\widetilde{u}_{i,\ell})_k \cdot \widetilde{h}_{i,\ell} \cdot \mathbf{1}_{(\widetilde{g}_{i,\ell+1})_k \geq 0} - \sum_{i=1}^n \sum_{\ell=1}^L (u_{i,\ell})_k \cdot h_{i,\ell} \cdot \mathbf{1}_{(g_{i,\ell+1})_k \geq 0} \right\|_2$$

$$\overset{\text{②}}{=} \left\| \sum_{i=1}^n \sum_{\ell=1}^L (\widetilde{u}_{i,\ell})_k \cdot \widetilde{h}_{i,\ell} \cdot \mathbf{1}_{(g_{i,\ell+1})_k \geq 0} - \sum_{i=1}^n \sum_{\ell=1}^L (u_{i,\ell})_k \cdot h_{i,\ell} \cdot \mathbf{1}_{(g_{i,\ell+1})_k \geq 0} \right\|_2$$

$$\overset{\text{③}}{\leq} \sum_{i=1}^n \sum_{\ell=1}^L \|((\widetilde{u}_{i,\ell})_k \cdot \widetilde{h}_{i,\ell} - (u_{i,\ell})_k \cdot h_{i,\ell}) \cdot \mathbf{1}_{(g_{i,\ell+1})_k \geq 0}\|_2 \ .$$

where ① follows by definition, ② follows by our choice of $k \in J$, and ③ follows by triangle inequality. For each $i, \ell$, using the same proof as (E.2) (so invoking Lemma C.11, Corollary B.19, Lemma B.3 and Lemma C.2a), we have with probability at least $1 - e^{-\Omega(\rho^2)}$:

$$\|(\widetilde{u}_{i,\ell})_k \cdot \widetilde{h}_{i,\ell} - (u_{i,\ell})_k \cdot h_{i,\ell}\|_2 \leq O\big(\frac{\rho^6 \tau_0^{1/3}}{m^{1/6}}\big) \cdot \|\operatorname{loss}_{i^*, a^*}\| \ . \tag{E.9}$$

which implies

$$\|\widehat{\nabla}_k f(\widetilde{W} + W_j'') - \widehat{\nabla}_k f(\widetilde{W})\|_2 \leq O\big(\frac{\rho^6 \tau_0^{1/3}}{m^{1/6}}\big) \cdot \|\operatorname{loss}_{i^*, a^*}\| \ .$$

Combining this with $\|\widehat{\nabla}_k f(\widetilde{W})\|_2 \leq O(\rho^4)$ from Lemma E.2, we finish the proof of the first item.

- For every index $k \in [m]$, we have

$$\|\widehat{\nabla}_k f(\widetilde{W} + W_j'') - \widehat{\nabla}_k f(\widetilde{W})\|_2$$
$$= \left\| \sum_{i=1}^n \sum_{\ell=1}^L (\widetilde{u}_{i,\ell})_k \cdot \widetilde{h}_{i,\ell} \cdot \mathbf{1}_{(\widetilde{g}_{i,\ell+1})_k \geq 0} - \sum_{i=1}^n \sum_{\ell=1}^L (u_{i,\ell})_k \cdot h_{i,\ell} \cdot \mathbf{1}_{(g_{i,\ell+1})_k \geq 0} \right\|_2$$
$$\leq \sum_{i=1}^n \sum_{\ell=1}^L \|(\widetilde{u}_{i,\ell})_k \cdot \widetilde{h}_{i,\ell}\|_2 + \|(u_{i,\ell})_k \cdot h_{i,\ell}\|_2 \ .$$

Using (E.9) and $\|(\widetilde{u}_{i,\ell})_k \cdot \widetilde{h}_{i,\ell}\|_2 \leq O(\rho^2 L)$ (see the proof of Lemma E.2), we immediately have the desired bound.

$\square$

*Proof of Lemma E.8.* Letting $J \stackrel{\text{def}}{=} \left\{ k \in [m] \ \middle| \ \left|\|\widehat{\nabla}_k f(\widetilde{W} + W_j'')\|_2^2 - \|\widehat{\nabla}_k f(\widetilde{W})\|_2^2\right| \leq O\left(\frac{\rho^{11} \theta^{1/3}}{m^{1/6}}\right) \right\}$, we have $|J| \geq \left(1 - \frac{\rho^5 \theta^{2/3}}{m^{1/3}}\right) m$ according to Claim E.9.

Now, for the indices in $N$ that are randomly sampled from $[m]$, if $|N \cap J| \geq |N| - S$ for a parameter $S = \rho^2$, then we have

$$\sum_{k \in N} \left|\|\widehat{\nabla}_k f(\widetilde{W})\|_2^2 - \|\widehat{\nabla}_k f(\widetilde{W} + W_j'')\|_2^2\right| \leq |S| \cdot O(\rho^6) + |N| \cdot O\left(\frac{\rho^{11} \theta^{1/3}}{m^{1/6}}\right) \leq O\left(\rho^8 + \frac{\rho^{11} \theta^{1/3}}{m^{1/6}}|N|\right) \ .$$

Otherwise, if $|N \cap J| \leq |N| - S$, this means at least $S$ indices that are chosen from $N$ are outside $J \subseteq [m]$. This happens with probability at most $\left(\frac{\rho^5 \theta^{2/3}}{m^{1/3}}\right)^S \leq e^{-\Omega(\rho^2)}$. $\square$

### E.3.2 Core Lemma B

In this subsection, we split $W$ into three parts $W = W_2 + W_N' + W_{-N}'$ followingdef:random-decomp:N-N. Our purpose is to rewrite Lemma E.8 into the following variant:

**Lemma E.10** (Core Lemma B). *Let $\rho = nLd\log m$ and $\varrho = nLd\delta^{-1}\log(m/\varepsilon)$. Fix parameter $\theta$ from Definition E.3, vectors $\{\text{loss}_{i,\ell} \in \mathbb{R}^d\}_{i\in[n],\ell\in[L]\setminus\{1\}}$, integer $|N| \in [1, \varrho^{100}]$, parameter $t = m/|N|$. Let $N_1, \ldots, N_t$ be i.i.d. random subsets of $[m]$ with cardinality $|N_i| = |N|$. Then, with probability $\geq 1 - e^{-\Omega(\rho^2)}$ over the randomness of $W_2, A, B$ and $N_1, \cdots N_t$, the following holds:*

- *for every $N \in \{N_1, N_2, \cdots, N_t\}$, the following holds:*
  - *with probability at $1 - e^{-\Omega(\rho^2)}$ over the randomness of $W'_{-N}$, the following holds:*
    * *with probability at least $1 - e^{-\Omega(\rho^2)}$ over the randomness of $W'_N$, we have:*
      · *for all $j \in N$, and for all $W''_j = u_j v_{i^*,\ell^*}^\top$ satisfying*

$$\|u_j\|_0 = 1, \quad \|u_j\|_\infty \leq \tfrac{3\theta\rho}{\sqrt{m}}, \quad v_{i^*,\ell^*} = \tfrac{(I-U_{\ell^*-1}U_{\ell^*-1}^\top)h_{i^*,\ell}h_{i^*,\ell^*}}{\|(I-U_{\ell^*-1}U_{\ell^*-1}^\top)h_{i^*,\ell^*}\|_2}, \quad \text{(E.10)}$$

*we have*

$$\sum_{k \in N} \left| \|\widehat{\nabla}_k f(W_2 + W'_N + W'_{-N})\|_2^2 - \|\widehat{\nabla}_k f(W_2 + W'_N + W'_{-N} + W''_j)\|_2^2 \right|$$
$$\leq O(\rho^8) \cdot \max_{i,\ell} \|\text{loss}_{i,\ell}\|_2^2 \ .$$

*Proof of Lemma E.10.* Without loss of generality we assume $\max_{i,\ell} \|\text{loss}_{i,\ell}\|_2^2 = 1$ in the proof. We now first rewrite Lemma E.8 as follows (recalling $m$ is sufficiently large so we only need to keep the $O(\rho^8)$ term):

- with probability $\geq 1 - e^{-\Omega(\rho^2)}$ over random $N \subset [m]$, random $j \in N$, random $W, A, B$, and random $W'_j$, the following holds:

$$\sum_{k\in[N]} \left| \|\widehat{\nabla}_k(W)\|_2^2 - \|\widehat{\nabla}_k(W + W''_j)\|_2^2 \right| \leq O\left(\rho^8 + \frac{\rho^{11}\theta^{1/3}}{m^{1/6}}|N|\right) \leq O(\rho^8) \ . \quad \text{(E.11)}$$

We can split the randomness (see Fact A.1) and derive that
- With probability $\geq 1 - e^{-\Omega(\rho^2)}$ over $N \subseteq [m], W, A, B$, the following holds:
  - With probability $\geq 1 - e^{-\Omega(\rho^2)}$ over $j \in N$ and $W''_j$, Eq. (E.11) holds.

Applying standard $\varepsilon$-net argument, we derive that
- With probability $\geq 1 - e^{-\Omega(\rho^2)}$ over $N \subseteq [m], W, A, B$, the following holds:
  - for all $j \in N$, and for all $W''_j = u_j v_{i^*,\ell^*}^\top$ satisfying (E.10), we have Eq. (E.11) holds.

Finally, letting $W = W_2 + W'_N + W'_{-N}$, we have
- With probability $\geq 1 - e^{-\Omega(\rho^2)}$ over the randomness of $N \subseteq [m], W_2, W_N, W'_{-N}, A, B$, the following holds:
  - for all $j \in N$, and for all $W''_j = u_j v_{i^*,\ell^*}^\top$ satisfying (E.10), we have

$$\sum_{k\in N} \left| \|\widehat{\nabla}_k f(W_2 + W'_N + W'_{-N})\|_2^2 - \|\widehat{\nabla}_k f(W_2 + W'_N + W_{-N'} + W''_j)\|_2^2 \right| \leq O\left(\rho^8\right) \ .$$

Finally, splitting the randomness (using Fact A.1), and applying a union bound over multiple samples $N_1, \ldots, N_t$, we finish the proof. □

### E.4 Main Theorem: Proof of Theorem 5

*Proof of Theorem 5.* Without loss of generality, we assume that $\max_{i,\ell} \|\text{loss}_{i,\ell}\|_2^2 = 1$.

Combining Core Lemma A and B (i.e., Lemma E.7 and Lemma E.10), we know that if $|N|$ is appropriately chosen, with probability $\geq 1 - e^{-\Omega(\rho^2)}$ over the randomness of $W_2, A, B$ and $N_1, \cdots N_t$, the following holds:
- for every $N \in \{N_1, N_2, \cdots, N_t\}$, the following holds:

- with probability at $1 - e^{-\Omega(\rho^2)}$ over the randomness of $W'_{-N}$, the following boxed statement holds:

---
**the boxed statement**

- (core A)
$$\mathbb{E}_{W'_N}\left[\sum_{k\in N} \|\widehat{\nabla}_k f(W_2 + W'_N + W'_{-N})\|_2^2\right] \geq q \stackrel{\text{def}}{=} \Omega\left(\frac{\beta_- |N|}{\rho^2}\right)$$

- (core B) with probability at least $1 - e^{-\Omega(\rho^2)}$ over the randomness of $W'_N$, the following holds:
  - for all $j \in N$, and for all $W''_j = u_j v_{i^*,\ell^*}^\top$ satisfying
  $$\|u_j\|_0 = 1, \quad \|u_j\|_\infty \leq \frac{3\theta\rho}{\sqrt{m}}, \quad v_{i^*,\ell^*} = \frac{(I - U_{\ell^*-1}U_{\ell^*-1}^\top)h_{i^*,\ell}h_{i^*,\ell^*}}{\|(I - U_{\ell^*-1}U_{\ell^*-1}^\top)h_{i^*,\ell^*}\|_2},$$
  we have
  $$\sum_{k\in N}\left|\|\widehat{\nabla}_k f(W_2 + W'_N + W'_{-N})\|_2^2 - \|\widehat{\nabla}_k f(W_2 + W'_N + W'_{-N} + W''_j)\|_2^2\right| \leq p \stackrel{\text{def}}{=} O(\rho^8).$$

---

We now wish to apply our extended McDiarmid's inequality (see Lemma A.19) to the boxed statement. For this goal, recalling that $W'_N = u_N v_{i^*,\ell^*}^\top$ for a vector $u_N \in \mathbb{R}^m$ that is only supported on indices in $N$. That is, $(u_N)_k = 0$ for all $k \notin N$. Therefore, we can define function $F(u_N) \stackrel{\text{def}}{=} \sum_{k\in N} \|\widehat{\nabla}_k f(W_2 + W'_{-N} + W'_N)\|^2$ where $W'_N = u_N v_{i^*,\ell^*}^\top$. We emphasize here that, inside the boxed statement, $W_2$, $W'_{-N}$ and $v_{i^*,\ell^*}$ are already fixed, and $u_N$ is the only source of randomness.

Below, we condition on the high probability event (see Lemma E.4d) that $\|u_N\|_\infty \leq \frac{3\theta\rho}{2\sqrt{m}}$. The boxed statement tells us:

- $\mathbb{E}_{u_N}[F(u_N)] \geq q$, and
- With probability $\geq 1 - e^{-\Omega(\rho^2)}$ over $u_N$, it satisfies
$$\forall j \in [N], \forall u''_j: \ \left|F(u_{-j}, u_j) - F(u_{-j}, u''_j)\right| \leq p.$$

At the same time, we have $0 \leq F(u_N) \leq O(N\rho^8)$ owing to Lemma E.2. Therefore, scaling down $F(u_N)$ by $\Theta(N\rho^8)$ (to make sure the function value stays in $[0,1]$), we can applying extended McDiarmid's inequality (see Lemma A.19), we have

$$\Pr_{u_N}\left[F(u_N) \geq \frac{q}{2}\right] \geq 1 - N^2 \cdot e^{-\Omega(\rho^2)} - \exp\left(-\Omega\left(\frac{(q/\rho^8)^2}{N(p/\rho^8)^2 + Ne^{-\Omega(\rho^2)}}\right)\right). \quad (E.12)$$

As long as $N \geq \frac{\rho^{22}}{\beta_-^2}$, we have that the above probability is at least $1 - e^{-\Omega(\rho^2)}$.

Finally, we choose

$$\boxed{N = \frac{\rho^{22}}{\beta_-^2}} \quad (E.13)$$

in order to satisfy Lemma E.7. We replace the boxed statement with (E.12). This tells us

- with probability $\geq 1 - e^{-\Omega(\rho^2)}$ over the randomness of $W_2, A, B$ and $N_1, \cdots N_t$, the following holds:
  - for every $N \in \{N_1, N_2, \cdots, N_t\}$, the following holds:
    * with probability at $1 - e^{-\Omega(\rho^2)}$ over the randomness of $W'_{-N}$ and $W'_N$, the following holds:
    $$\sum_{k\in N} \|\widehat{\nabla}_k f(W_2 + W'_N + W'_{-N})\|_2^2 \geq \Omega\left(\frac{\beta_-|N|}{\rho^2}\right)$$

After rearranging randomness, and using $W = W_2 + W'_N + W'_{-N}$, we have

- with probability $\geq 1 - e^{-\Omega(\rho^2)}$ over the randomness of $W, A, B$, the following holds:
  - with probability $\geq 1 - e^{-\Omega(\rho^2)}$ over the randomness of $N_1, \cdots N_t$, the following holds:
    * for every $N \in \{N_1, N_2, \cdots, N_t\}$, the following holds:
    $$\sum_{k \in N} \|\widehat{\nabla}_k f(W)\|_2^2 \geq \Omega\left(\frac{\beta_-|N|}{\rho^2}\right)$$

Finally, since $N_1, \ldots, N_t$ are $t$ random subsets of $[m]$, we know that with probability at least $1 - e^{-\Omega(\rho^2)}$, for each index $k \in [m]$, it is covered by at most $\rho^2$ random subsets. Therefore,

$$\sum_{k \in N} \|\widehat{\nabla}_k f(W)\|_2^2 \geq \frac{1}{\rho^2} \times t \times \Omega\left(\frac{\beta_-|N|}{\rho^2}\right) = \Omega\left(\frac{\delta}{\rho^{14}}\right) \times m \ .$$

□

# F  Gradient Bound After Perturbation (Theorem 3)

We use the same fake gradient notion (see Definition E.1) and first derive the following result based on Lemma E.2 and Theorem 5.

**Lemma F.1.** *Let $\rho = nLd\log m$ and $\varrho = nLd\delta^{-1}\log(m/\varepsilon)$. Given fixed vectors $\{\mathsf{loss}_{i,\ell} \in \mathbb{R}^d\}_{i \in [n], \ell \in [L] \setminus \{1\}}$, with probability at least $1 - e^{-\Omega(\rho^2)}$ over $\widetilde{W}, A, B$, it satisfies for all $W' \in \mathbb{R}^{m \times m}$ with $\|W'\|_2 \le \frac{\tau_0}{\sqrt{m}}$ with $\tau_0 \le \varrho^{100}$,*

$$\|\widehat{\nabla} f(\widetilde{W} + W')\|_F^2 \ge \Omega\left(\frac{\delta}{\rho^{14}}\right) \times m \times \max_{i,\ell}\{\|\mathsf{loss}_{i,\ell}\|^2\}$$

$$\|\widehat{\nabla} f(\widetilde{W} + W')\|_F^2 \le O(\rho^{12} m) \times \max_{i,\ell}\{\|\mathsf{loss}_{i,\ell}\|^2\}$$

$$\|\widehat{\nabla} f_i(\widetilde{W} + W')\|_F^2 \le \frac{1}{n^2} O(\rho^{12} m) \times \max_{i,\ell}\{\|\mathsf{loss}_{i,\ell}\|^2\} \ .$$

*Proof of Lemma F.1.* Like before, we denote by $\widetilde{D}_{i,\ell}, \widetilde{u}_{i,\ell}, \widetilde{h}_{i,\ell}$ the corresponding matrices or vectors when the weight matrix is $\widetilde{W}$, and by $D_{i,\ell}, u_{i,\ell}, h_{i,\ell}$ the corresponding terms when when the weight matrix is $W = \widetilde{W} + W'$.

**Lower bound on $\|\widehat{\nabla} f(\widetilde{W} + W')\|_F$.**  Recall

$$\widehat{\nabla} f(\widetilde{W} + W') = \sum_{i=1}^n \sum_{\ell=1}^L D_{i,\ell+1} \left(\sum_{a=\ell}^L \mathsf{Back}_{i,\ell \to a}^\top \mathsf{loss}_{i,a}\right) h_{i,\ell}^\top = \sum_{i=1}^n \sum_{\ell=1}^L D_{i,\ell+1} u_{i,\ell} h_{i,\ell}^\top$$

In Theorem 5 of the previous section, we already have a lower bound on $\|\widehat{\nabla} f(\widetilde{W})\|_F$. We just need to upper bound $\|\widehat{\nabla} f(\widetilde{W} + W') - \widehat{\nabla} f(\widetilde{W})\|_F$.

$$\|\widehat{\nabla} f(\widetilde{W} + W') - \widehat{\nabla} f(\widetilde{W})\|_F \stackrel{①}{=} \left\|\sum_{i=1}^n \sum_{\ell=1}^L D_{i,\ell+1} u_{i,\ell} h_{i,\ell}^\top - \widetilde{D}_{i,\ell+1} \widetilde{u}_{i,\ell} \widetilde{h}_{i,\ell}^\top\right\|_F$$

$$\stackrel{②}{\le} \sum_{i=1}^n \sum_{\ell=1}^L \left\|D_{i,\ell+1} u_{i,\ell} h_{i,\ell}^\top - \widetilde{D}_{i,\ell+1} \widetilde{u}_{i,\ell} \widetilde{h}_{i,\ell}^\top\right\|_F$$

$$\stackrel{③}{\le} \sum_{i=1}^n \sum_{\ell=1}^L \|D'_{i,\ell+1} \widehat{u}_{i,\ell} \widehat{h}_{i,\ell}^\top\|_F + \|\widetilde{D}_{i,\ell+1} u'_{i,\ell} \widehat{h}_{i,\ell}^\top\|_F + \|\widetilde{D}_{i,\ell+1} \widetilde{u}_{i,\ell} h'_{i,\ell}{}^\top\|_F + o(m^{1/3}).$$

where ① follows by definition and ② and ③ follow by the triangle inequality. Note that in inequality ③ we have hidden four more higher order terms in $o(m^{1/3})$. We ignore the details for how to bound them, for the ease of presentation. We bound the three terms separately:

- Using Corollary B.18 (with $s^2 = O(L^{10/3} \tau_0^{2/3})$ from Lemma C.2b) and Lemma B.3 we have

$$\|D'_{i,\ell+1} \widehat{u}_{i,\ell} \widehat{h}_{i,\ell}^\top\|_F \le \|D'_{i,\ell+1} \widehat{u}_{i,\ell}\|_2 \cdot \|\widehat{h}_{i,\ell}\|_2$$

$$\le O(L \cdot L^4 \tau_0^{1/3} m^{1/3} \log m \cdot \|\mathsf{loss}_{i^*, \ell^*}\|_2) \cdot L$$

- Using $\|\widetilde{D}_{i,\ell+1}\|_2 \le 1$, Lemma C.9 and Lemma B.3 we have

$$\|\widetilde{D}_{i,\ell+1} u'_{i,\ell} \widehat{h}_{i,\ell}^\top\|_F \le \|\widetilde{D}_{i,\ell+1}\| \cdot \|u'_{i,\ell}\|_2 \cdot \|\widehat{h}_{i,\ell}\|_2$$

$$\le O(L \cdot L^6 \tau_0^{1/3} m^{1/3} \log m \cdot \|\mathsf{loss}_{i^*, \ell^*}\|_2) \cdot L$$

- Using $\|\widetilde{D}_{i,\ell+1}\|_2 \leq 1$, $\|\widetilde{u}_{i,\ell}\|_2 \leq O(L^4)\sqrt{m}\|\mathsf{loss}_{i^*,\ell^*}\|_2$ (implied by Lemma B.11) and Lemma C.2a, we have

$$\|\widetilde{D}_{i,\ell+1}\widetilde{u}_{i,\ell}{h'_{i,\ell}}^\top\|_F \leq \|\widetilde{D}_{i,\ell+1}\| \cdot \|\widetilde{u}_{i,\ell}\|_2 \cdot \|h'_{i,\ell}\|_2$$
$$\leq O(L^4\sqrt{m} \cdot \|\mathsf{loss}_{i^*,\ell^*}\|_2) \cdot (L^6 \tau_0^{1/2} \frac{1}{\sqrt{m}})$$

Putting it all together, we have

$$\|\widehat{\nabla} f(\widetilde{W} + W') - \widehat{\nabla} f(\widetilde{W})\|_F \leq O(\rho^8 \tau_0^{1/3} m^{1/3}) \cdot \|\mathsf{loss}_{i^*,\ell^*}\|_F \ . \tag{F.1}$$

Finally, using $(a-b)^2 \geq \frac{1}{2}a^2 - b^2$, we have

$$\|\widehat{\nabla} f(\widetilde{W} + W')\|_F^2 \geq (\|\widehat{\nabla} f(\widetilde{W})\|_F - \|\widehat{\nabla} f(\widetilde{W} + W') - \widehat{\nabla} f(\widetilde{W})\|_F)^2$$
$$\geq \frac{1}{2}\|\widehat{\nabla} f(\widetilde{W})\|_F^2 - \|\widehat{\nabla} f(\widetilde{W} + W') - \widehat{\nabla} f(\widetilde{W})\|_F^2$$
$$\geq \Omega\left(\frac{\delta}{\rho^{14}}\right) \times m \times \|\mathsf{loss}_{i^*,\ell^*}\|^2 \ .$$

where the last step follows by (F.1) (with our sufficiently large choice of $m$) and Theorem 5.

**Upper bound on $\|\widehat{\nabla} f(\widetilde{W} + W')\|_F$.** Using $(a+b)^2 \leq 2a^2 + 2b^2$, we have

$$\|\widehat{\nabla} f(\widetilde{W} + W')\|_F^2 \leq 2\|\widehat{\nabla} f(\widetilde{W})\|_F^2 + 2\|\widehat{\nabla} f(\widetilde{W} + W') - \widehat{\nabla} f(\widetilde{W})\|_F^2$$
$$\overset{\textcircled{\scriptsize 1}}{\leq} O(\rho^{12}m) \cdot \|\mathsf{loss}_{i^*,\ell^*}\|_F^2 + 2\|\widehat{\nabla} f(\widetilde{W} + W') - \widehat{\nabla} f(\widetilde{W})\|_F^2 \overset{\textcircled{\scriptsize 2}}{\leq} O(\rho^{12}m) \cdot \|\mathsf{loss}_{i^*,\ell^*}\|_F^2$$

where ① follows by Lemma E.2 and ② follows by (F.1).

**Upper bound on $\|\widehat{\nabla} f_i(\widetilde{W} + W')\|_F$.** This is completely analogous so we do not replicate the proofs here. □

By applying an $\varepsilon$-net argument over all possible $\mathsf{loss}_{i,\ell} \in \mathbb{R}^d$, we can modify Lemma F.1 from "for fixed $\mathsf{loss}$" to "for all $\mathsf{loss}$." This allows us to plug in the true loss $\mathsf{loss}_{i,\ell} = B h_{i,\ell} - y^*_{i,\ell}$ and derive that:

> **Theorem 3** (restated). *Let $\rho = nLd\log m$ and $\varrho = nLd\delta^{-1}\log(m/\varepsilon)$. With probability at least $1 - e^{-\Omega(\rho^2)}$ over $W, A, B$, it satisfies for all $W' \in \mathbb{R}^{m\times m}$ with $\|W'\|_2 \leq \frac{\varrho^{100}}{\sqrt{m}}$,*
> 
> $$\|\nabla f(\widetilde{W} + W')\|_F^2 \geq \Omega\left(\frac{\delta}{\rho^{14}}\right) \times m \times \max_{i,\ell}\{\|\mathsf{loss}_{i,\ell}\|^2\}$$
> $$\|\nabla f(\widetilde{W} + W')\|_F^2 \leq O(\rho^{12}m) \times \max_{i,\ell}\{\|\mathsf{loss}_{i,\ell}\|^2\}$$
> $$\|\nabla f_i(\widetilde{W} + W')\|_F^2 \leq \frac{1}{n^2}O(\rho^{12}m) \times \max_{i,\ell}\{\|\mathsf{loss}_{i,\ell}\|^2\} \ .$$

# G  Objective Semi-Smoothness (Theorem 4)

The purpose of this section is to prove

> **Theorem 4** (objective semi-smoothness, restated)**.** *Let $\rho = nLd\log m$, $\varrho = nLd\delta^{-1}\log(m/\varepsilon)$, $\tau_0 \in [\varrho^{-100}, \varrho^{100}]$, and $\widetilde{W}, A, B$ be at random initialization. With probability at least $1 - e^{-\Omega(\rho^2)}$ over the randomness of $\widetilde{W}, A, B$, we have for every $\check{W} \in \mathbb{R}^{m \times m}$ with $\|\check{W} - \widetilde{W}\| \le \frac{\tau_0}{\sqrt{m}}$, and for every $W' \in \mathbb{R}^{m \times m}$ with $\|W'\| \le \frac{\tau_0}{\sqrt{m}}$, and letting $(i^*, \ell^*) = \arg\max_{i,\ell}\{\|\mathsf{loss}_{i,\ell}\|_2\}$,*
> 
> $$f(\check{W} + W') \le f(\check{W}) + \langle \nabla f(\check{W}), W' \rangle + O(\rho^{11}\tau_0^{1/3}m^{1/3}) \cdot \|\mathsf{loss}_{i^*,\ell^*}\|_2 \cdot \|W'\|_2 + O(L^{18}nm)\|W'\|_2^2$$

We introduce the following notations before we go to proofs.

**Definition G.1.** *For $i \in [n]$ and $\ell \in [L]$:*

$$\widetilde{g}_{i,0} = \widetilde{h}_{i,0} = 0 \qquad \check{g}_{i,0} = \check{h}_{i,0} = 0 \qquad g_{i,0} = h_{i,0} = 0$$
$$\widetilde{g}_{i,\ell} = \widetilde{W}\widetilde{h}_{i,\ell-1} + Ax_{i,\ell} \qquad \check{g}_{i,\ell} = \check{W}\check{h}_{i,\ell-1} + Ax_{i,\ell} \qquad g_{i,\ell} = (\check{W}+W')h_{i,\ell-1} + Ax_{i,\ell}$$
$$\widetilde{h}_{i,\ell} = \phi(\widetilde{W}\widetilde{h}_{i,\ell-1} + Ax_{i,\ell}) \qquad \check{h}_{i,\ell} = \phi(\check{W}\check{h}_{i,\ell-1} + Ax_{i,\ell}) \qquad h_{i,\ell} = \phi((\check{W}+W')h_{i,\ell-1} + Ax_{i,\ell})$$
$$\mathsf{loss}_{i,\ell} = B\check{h}_{i,\ell} - y^*_{i,\ell}$$

*Define diagonal matrices $\widetilde{D}_{i,\ell}$ and $\check{D}_{i,\ell}$ respectively by letting*

$$(\widetilde{D}_{i,\ell})_{k,k} = \mathbb{1}_{(\widetilde{g}_{i,\ell})_k \ge 0} \text{ and } (\check{D}_{i,\ell})_{k,k} = \mathbb{1}_{(\check{g}_{i,\ell})_k \ge 0}.$$

The following claim gives rise to a new recursive formula to calculate $h_{i,\ell} - \check{h}_{i,\ell}$.

**Claim G.2** (c.f. (7.1))**.** *There exist diagonal matrices $D''_{i,\ell} \in \mathbb{R}^{m \times m}$ with entries in $[-1,1]$ such that,*

$$\forall i \in [n], \forall \ell \in [L]: \quad h_{i,\ell} - \check{h}_{i,\ell} = \sum_{a=1}^{\ell-1}(\check{D}_{i,\ell} + D''_{i,\ell})\check{W} \cdots \check{W}(\check{D}_{i,a+1} + D''_{i,a+1})W'h_{i,a} \qquad (G.1)$$

*Furthermore, we have $\|h_{i,\ell} - \check{h}_{i,\ell}\| \le O(L^9)\|W'\|_2$ and $\|D''_{i,\ell}\|_0 \le O(L^{10/3}\tau_0^{2/3}m^{2/3})$.*

*Proof of Claim G.2.* We ignore the subscript in $i$ for cleanness, and calculate that

$$h_\ell - \check{h}_\ell \overset{\text{①}}{=} \phi((\check{W}+W')h_{\ell-1} + Ax_\ell) - \phi(\check{W}\check{h}_{\ell-1} + Ax_\ell)$$
$$\overset{\text{②}}{=} (\check{D}_\ell + D''_\ell)\left((\check{W}+W')h_{\ell-1} - \check{W}\check{h}_{\ell-1}\right)$$
$$= (\check{D}_\ell + D''_\ell)\check{W}(h_{\ell-1} - \check{h}_{\ell-1}) + (\check{D}_\ell + D''_\ell)W'h_{\ell-1}$$
$$\overset{\text{③}}{=} \sum_{a=1}^{\ell-1}(\check{D}_\ell + D''_\ell)\check{W} \cdots \check{W}(\check{D}_{a+1} + D''_{a+1})W'h_a$$

Above, ① is by the recursive definition of $h_\ell$ and $\check{h}_\ell$; ② is by Proposition G.3 and $D''_\ell$ is defined according to Proposition G.3; and inequality ③ is by recursively computing $h_{\ell-1} - \check{h}_{\ell-1}$. As for the two properties:

- We have $\|h_\ell - \check{h}_\ell\| \le O(L^9)\|W'\|_2$. This is because we have
  - $\left\|(\check{D}_\ell + D''_\ell)\check{W} \cdots \check{W}(\check{D}_{a+1} + D''_{a+1})\right\|_2 \le O(L^7)$ by Lemma C.7;
  - $\|h_a\| \le O(L)$ (by $\|\widetilde{h}_a\| \le O(L)$ from Lemma B.3 and $\|\widetilde{h}_a - h_a\| \le o(1)$ from Lemma C.2a); and

- $\|W'h_a\| \leq \|W'\|_2 \|h_a\|$.

- We have $\|D''_\ell\|_0 \leq O(L^{10/3} \tau_0^{2/3} m^{2/3})$.

  This is because, $(D''_\ell)_{k,k}$ is non-zero only at the coordinates $k \in [m]$ where the signs of $\breve{g}_\ell$ and $g_\ell$ are opposite (by Proposition G.3). Such a coordinate $k$ must satisfy either $(\widetilde{D}_\ell)_{k,k} \neq (\breve{D}_\ell)_{k,k}$ or $(\widetilde{D}_\ell)_{k,k} \neq (D_\ell)_{k,k}$, and therefore by Lemma C.2 — with probability $\geq 1 - e^{-\Omega(L^6 \tau_0^{4/3} m^{1/3})}$ — there are at most $O(L^{10/3} \tau_0^{2/3} m^{2/3})$ such coordinates $k$.

  □

*Proof of Theorem 4.* First of all, since
$$\frac{1}{2}\|Bh_{i,\ell} - y^*_{i,\ell}\|^2 = \frac{1}{2}\|\breve{\text{loss}}_{i,\ell} + B(h_{i,\ell} - \breve{h}_{i,\ell})\|^2 = \frac{1}{2}\|\breve{\text{loss}}_{i,\ell}\|^2 + \breve{\text{loss}}_{i,\ell}^\top B(h_{i,\ell} - \breve{h}_{i,\ell}) + \frac{1}{2}\|B(h_{i,\ell} - \breve{h}_{i,\ell})\|^2 \tag{G.2}$$

we can write

$$f(\breve{W} + W') - f(W) - \langle \nabla f(W), W' \rangle$$
$$\overset{①}{=} -\langle \nabla f(\breve{W}), W' \rangle + \frac{1}{2}\sum_{i=1}^{n}\sum_{\ell=2}^{L} \|Bh_{i,\ell} - y^*_{i,\ell}\|^2 - \|B\breve{h}_{i,\ell} - y^*_{i,\ell}\|^2$$
$$\overset{②}{=} -\langle \nabla f(\breve{W}), W' \rangle + \sum_{i=1}^{n}\sum_{\ell=2}^{L} \breve{\text{loss}}_{i,\ell}^\top B(h_{i,\ell} - \breve{h}_{i,\ell}) + \frac{1}{2}\|B(h_{i,\ell} - \breve{h}_{i,\ell})\|^2$$
$$\overset{③}{=} \sum_{i=1}^{n}\sum_{\ell=2}^{L} \breve{\text{loss}}_{i,\ell}^\top B \left( (h_{i,\ell} - \breve{h}_{i,\ell}) - \sum_{a=1}^{\ell-1} \breve{D}_{i,\ell}\breve{W}\cdots\breve{W}\breve{D}_{i,a+1}W'\breve{h}_{i,a} \right) + \frac{1}{2}\|B(h_{i,\ell} - \breve{h}_{i,\ell})\|^2$$
$$\overset{④}{=} \sum_{i=1}^{n}\sum_{\ell=2}^{L} \breve{\text{loss}}_{i,\ell}^\top B \left( \sum_{a=1}^{\ell-1}(\breve{D}_{i,\ell} + D''_{i,\ell})\breve{W}\cdots\breve{W}(\breve{D}_{i,a+1} + D''_{i,a+1})W'h_{i,a} - \breve{D}_{i,\ell}\breve{W}\cdots\breve{W}\breve{D}_{i,a+1}W'\breve{h}_{i,a} \right)$$
$$+ \frac{1}{2}\|B(h_{i,\ell} - \breve{h}_{i,\ell})\|^2 \tag{G.3}$$

Above, ① is by the definition of $f(\cdot)$; ② is by (G.2); ③ is by the definition of $\nabla f(\cdot)$ (see Fact 2.5 for an explicit form of the gradient); ④ is by Claim G.2.

We next bound the RHS of (G.3). We drop subscripts in $i$ for notational simplicity. We first use (G.1) in Claim G.2 to calculate

$$\|B(h_\ell - \breve{h}_\ell)\| \leq \sum_{a=1}^{\ell-1} \|B(\breve{D}_\ell + D''_\ell)\breve{W}\cdots\breve{W}(\breve{D}_{a+1} + D''_{a+1})W'h_a\|$$
$$\leq \sum_{a=1}^{\ell-1} \|B\|_2 \|(\breve{D}_\ell + D''_\ell)\breve{W}\cdots\breve{W}(\breve{D}_{a+1} + D''_{a+1})\|_2 \|W'\|_2 \|h_a\|$$
$$\overset{①}{\leq} \sum_{a=1}^{\ell-1} O(\sqrt{m}) \cdot O(L^7) \cdot O(L) \cdot \|W'\|_2 \leq O(L^9\sqrt{m}) \cdot \|W'\|_2 . \tag{G.4}$$

In the last inequality ① above, we have used $\left\|(\breve{D}_\ell + D''_\ell)\breve{W}\cdots\breve{W}(\breve{D}_{a+1} + D''_{a+1})\right\|_2 \leq O(L^7)$ from Lemma C.7; we have used $\|B\|_2 \leq O(\sqrt{m})$ with high probability; and we have used $\|h_a\| \leq O(L)$ (by $\|\widetilde{h}_a\| \leq O(L)$ from Lemma B.3 and $\|\widetilde{h}_a - h_a\| \leq o(1)$ from Lemma C.2a).

Since for each $D''_\ell$, we can write it as $D''_\ell = D^{0/1}_\ell D''_\ell D^{0/1}_\ell$, where each $D^{0/1}_\ell$ is a diagonal matrix

satisfying

$$(D_\ell^{0/1})_{k,k} = \begin{cases} 1, & (D_\ell'')_{k,k} \neq 0; \\ 0, & (D_\ell'')_{k,k} = 0. \end{cases} \quad \text{and} \quad \|D_\ell^{0/1}\|_0 \leq O(L^{10/3}\tau_0^{2/3}m^{2/3})$$

Therefore,

$$\left|\widecheck{\mathsf{loss}}_\ell^\top B(\widecheck{D}_\ell + D_\ell'')\widecheck{W}\cdots\widecheck{W}(\widecheck{D}_{a+1} + D_{a+1}'')W'h_a - \widecheck{\mathsf{loss}}_\ell^\top B\widecheck{D}_\ell\widecheck{W}\cdots\widecheck{D}_{a+1}W'h_a\right|$$

$$\overset{①}{\leq} \|\widecheck{\mathsf{loss}}_\ell\|_2 \cdot \sum_{b=1}^{\ell-a} \binom{\ell-a}{b} \|B\widecheck{D}\widecheck{W}\cdots\widecheck{D}\widecheck{W}D^{0/1}\|_2 \cdot \|D^{0/1}\widecheck{W}\widecheck{D}\cdots\widecheck{D}\widecheck{W}D^{0/1}\|_2^{b-1} \cdot \|D^{0/1}\widecheck{W}\cdots\widecheck{W}\widecheck{D}W'h_a\|_2$$

$$\overset{②}{\leq} \|\widecheck{\mathsf{loss}}_\ell\|_2 \cdot \sum_{b=1}^{\ell-a} \binom{\ell-a}{b} \cdot O(\rho^2\tau_0^{1/3}m^{1/3}) \cdot \left(O(\frac{\rho^2}{m^{1/6}})\right)^{b-1} \cdot O(L^7) \cdot \|W'\|_2 \cdot O(L)$$

$$\overset{②}{\leq} \|\widecheck{\mathsf{loss}}_\ell\|_2 \cdot O(\rho^{11}\tau_0^{1/3}m^{1/3}) \cdot \|W'\|_2 \tag{G.5}$$

Above, ① is an abbreviation and we have dropped the subscripts for the easy of presentation; ② uses Corollary B.18, Corollary B.15, Lemma C.7 and Lemma B.3.

Finally, we also have

$$\left|\widecheck{\mathsf{loss}}_\ell^\top B\widecheck{D}_\ell\widecheck{W}\cdots\widecheck{D}_{a+1}W'(h_a - \widecheck{h}_a)\right| \overset{①}{\leq} \|\widecheck{\mathsf{loss}}_\ell\|_2 \cdot O(\sqrt{m}) \cdot O(L^7) \cdot \|W'\|_2 \cdot \|h_a - \widecheck{h}_a\|_2$$

$$\overset{②}{\leq} O(\rho^{16}\sqrt{m}) \cdot \|\widecheck{\mathsf{loss}}_\ell\|_2 \cdot \|W'\|_2^2 \tag{G.6}$$

where ① uses Lemma C.7 and ② uses Claim G.2 to bound $\|h_a - \widecheck{h}_a\|_2$.

Putting (G.4), (G.5) and (G.6) back to (G.3), and using triangle inequality, we have the desired result. □

### G.1 Tool

**Proposition G.3.** *Given vectors $a, b \in \mathbb{R}^m$ and $D \in \mathbb{R}^{m \times m}$ the diagonal matrix where $D_{k,k} = \mathbb{1}_{a_k \geq 0}$. Then, then there exists a diagonal matrix $D'' \in \mathbb{R}^{m \times m}$ with*

- $|D_{k,k}''| \leq 1$ *for every* $k \in [m]$,
- $D_{k,k}'' \neq 0$ *only when* $\mathbb{1}_{a_k \geq 0} \neq \mathbb{1}_{b_k \geq 0}$, *and*
- $\phi(a) - \phi(b) = D(a - b) + D''(a - b)$

*Proof.* We verify coordinate by coordinate for each $k \in [m]$.

- If $a_k \geq 0$ and $b_k \geq 0$, then $(\phi(a) - \phi(b))_k = a_k - b_k = \big(D(a - b)\big)_k$.
- If $a_k < 0$ and $b_k < 0$, then $(\phi(a) - \phi(b))_k = 0 - 0 = \big(D(a - b)\big)_k$.
- If $a_k \geq 0$ and $b_k < 0$, then $(\phi(a) - \phi(b))_k = a_k = (a_k - b_k) + \frac{b_k}{a_k - b_k}(a_k - b_k) = \big(D(a - b) + D''(a - b)\big)_k$, if we define $(D'')_{k,k} = \frac{b_k}{a_k - b_k} \in [-1, 0]$.
- If $a_k < 0$ and $b_k \geq 0$, then $(\phi(a) - \phi(b))_k = -b_k = 0 \cdot (a_k - b_k) - \frac{b_k}{b_k - a_k}(a_k - b_k) = \big(D(a - b) + D''(a - b)\big)_k$, if we define $(D'')_{k,k} = \frac{b_k}{b_k - b_a} \in [0, 1]$. □

# H Convergence Rate of Gradient Descent (Theorem 1)

**Theorem 1** (gradient descent, restated). *There exists some absolute constant $C > 1$ such that the following holds. Let $\rho = nLd \log m$, $\varepsilon \in (0,1]$, $\varrho = nLd\delta^{-1}\log(m/\varepsilon)$, and $\eta \stackrel{\text{def}}{=} \frac{\delta}{\rho^{44}m}$. Suppose $m \geq C\varrho^C$, and $W^{(0)}, A, B$ be at random initialization. Then, with probability at least $1 - e^{-\Omega(\rho^2)}$ over the randomness of $W^{(0)}, A, B$, suppose we start at $W^{(0)}$ and for each $t = 0, 1, \ldots, T-1$,*

$$W^{(t+1)} = W^{(t)} - \eta \nabla f(W^{(t)}) \ .$$

*Then, it satisfies*

$$f(W^{(T)}) \leq \varepsilon \quad \text{for all} \quad T \in \left[ \frac{\rho^{59}}{\delta^2} \log \frac{1}{\varepsilon}, \frac{\rho^{59}\varrho^{28}}{\delta^2} \log \frac{1}{\varepsilon} \right] \ .$$

*In other words, the training loss of the recurrent neural network drops to $\varepsilon$ in a linear convergence speed.*

To present the simplest possible result, we have not tried to tighten the polynomial dependency with respect to $n, d$ and $L$. We only tightened the dependency with respect to $\delta$ and $\varepsilon$. In fact, a more involved analysis can also get ride of the $\log(1/\varepsilon)$ dependency in $m$ [3].

*Proof of Theorem 1.* Using Lemma B.3 and the randomness of $B$, it is easy to show that $\|Bh_{i,\ell} - y^*_{i,\ell}\|^2 \leq O(\rho^2 L^2)$ with at least $1 - e^{-\Omega(\rho^2)}$ (where $h_{i,\ell}$ is defined with respect to $W^{(0)}$), and therefore

$$f(W^{(0)}) \leq O(n\rho^2 L^3) \ .$$

In the rest of the proof, we first assume that for every $t = 0, 1, \ldots, T-1$, the following holds

$$\|W^{(t)} - W^{(0)}\|_F \leq \frac{\tau_0}{\sqrt{m}} \stackrel{\text{def}}{=} \frac{\varrho^{50}}{\sqrt{m}} \ . \tag{H.1}$$

We shall prove the convergence of gradient descent assuming (H.1), so that previous statements such as Theorem 4 and Theorem 3 can be applied. At the end of the proof, we shall verify that (H.1) is satisfied throughout the gradient descent process.

For each $t = 0, 1, \ldots, T-1$, we let $\{\text{loss}^{(t)}_{i,\ell}\}_{i,\ell}$ denote the loss vectors with respect to the current point $W^{(t)}$. We denote by $(i^*, \ell^*) = \arg\max_{i,\ell}\{\|\text{loss}^{(t)}_{i,\ell}\|_2\}$ and $\nabla_t = \nabla f(W^{(t)})$.

We calculate that

$$f(W^{(t+1)}) \stackrel{①}{\leq} f(W^{(t)}) - \eta\|\nabla_t\|_F^2 + O(\rho^{11}\tau_0^{1/3}m^{1/3}) \cdot \|\text{loss}^{(t)}_{i^*,\ell^*}\|_2 \cdot \eta\|\nabla_t\|_2 + O(L^{18}nm\eta^2)\|\nabla_t\|_2^2$$

$$\stackrel{②}{\leq} f(W^{(t)}) - \eta\|\nabla_t\|_F^2 + O\left(\rho^{30}\eta^2 m^2\right) \cdot \|\text{loss}^{(t)}_{i^*,\ell^*}\|_2^2$$

$$\stackrel{③}{\leq} f(W^{(t)}) - \left(\Omega\Big(\frac{\eta\delta}{\rho^{14}}m\Big) - O(\rho^{30}\eta^2 m^2)\right) \cdot \|\text{loss}^{(t)}_{i^*,\ell^*}\|_2^2$$

$$\stackrel{④}{\leq} f(W^{(t)}) - \Omega\Big(\frac{\eta\delta}{\rho^{14}}m\Big) \cdot \|\text{loss}^{(t)}_{i^*,\ell^*}\|_2^2$$

$$\stackrel{⑤}{\leq} \left(1 - \Omega\Big(\frac{\eta\delta}{\rho^{15}}m\Big)\right) f(W^{(t)})$$

Above, ① uses Theorem 4; ② uses Theorem 3 (which gives $\|\nabla_t\|_2 \leq \|\nabla_t\|_F \leq O(\rho^6\sqrt{m}) \times \|\text{loss}^{(t)}_{i^*,\ell^*}\|$), and our choices of $\tau_0$ and $m$; ③ uses Theorem 3; ④ uses our choice of $\eta$; and ⑤ uses $f(W^{(t)}) \leq nL\|\text{loss}^{(t)}_{i^*,\ell^*}\|^2$. In other words, after $T = \Omega(\frac{\rho^{15}}{\eta\delta m})\log\frac{nL^2}{\varepsilon}$ iterations we have $f(W^{(T)}) \leq \varepsilon$.

We need to verify for each $t$, $\|W^{(t)} - W^{(0)}\|_F$ is small so that (H.1) holds. By Theorem 3,

$$\|W^{(t)} - W^{(0)}\|_F \leq \sum_{i=0}^{t-1} \|\eta \nabla f(W^{(i)})\|_F \leq O(\eta \rho^6 \sqrt{m}) \cdot \sum_{i=0}^{t-1} \sqrt{f(W^{(i)})} \leq O(\eta \rho^6 \sqrt{m}) \cdot O(T \cdot \sqrt{n\rho^2 L^3})$$

$$\leq \eta T \cdot O(\rho^{8.5} \sqrt{m}) \leq \frac{\varrho^{50}}{\sqrt{m}} \ .$$

where the last step follows by our choice of $T$. $\square$

## I Convergence Rate of Stochastic Gradient Descent (Theorem 2)

**Theorem 2** (stochastic gradient descent, stated). *There exists some absolute constant $C > 1$ such that the following holds. Let $\rho = nLd\log m$, $\varepsilon \in (0,1]$, $\varrho = nLd\delta^{-1}\log(m/\varepsilon)$, and $\eta \stackrel{\text{def}}{=} \frac{\delta}{\rho^{42} m}$. Suppose $m \geq C\varrho^C$, and $W^{(0)}, A, B$ be at random initialization. Then, with probability at least $1 - e^{-\Omega(\rho^2)}$ over the randomness of $W^{(0)}, A, B$, suppose we start at $W^{(0)}$ and for each $t = 0, 1, \ldots, T-1$,*

$$W^{(t+1)} = W^{(t)} - \eta \cdot \frac{n}{|S_t|} \sum_{i \in S_t} \nabla f(W^{(t)}) \qquad \text{(for a random subset } S_t \subseteq [n] \text{ of fixed cardinality.)}$$

*Then, it satisfies with probability at least $1 - e^{-\Omega(\rho^2)}$ over the randomness of $S_1, \ldots, S_T$:*

$$f(W^{(T)}) \leq \varepsilon \quad \text{for all} \quad T \in \Big[\frac{\rho^{57}}{\delta^2} \log \frac{1}{\varepsilon}, \frac{\rho^{57} \varrho^{28}}{\delta^2} \log \frac{1}{\varepsilon}\Big] \ .$$

(To present the simplest possible result, we have not tried to tighten the polynomial dependency with respect to $n, d$ and $L$. We only tightened the dependency with respect to $\delta$ and $\varepsilon$. In fact, a more involved analysis can also get ride of the $\log(1/\varepsilon)$ dependency in $m$.)

*Proof of Theorem 2.* The proof is almost identical to that of Theorem 1. We again have with probability at least $1 - e^{-\Omega(\rho^2)}$

$$f(W^{(0)}) \leq O(n\rho^2 L^3) \ .$$

Again, we first assume for every $t = 0, 1, \ldots, T-1$, the following holds

$$\|W^{(t)} - W^{(0)}\|_F \leq \frac{\tau_0}{\sqrt{m}} \stackrel{\text{def}}{=} \frac{\varrho^{50}}{\sqrt{m}} \ . \tag{I.1}$$

We shall prove the convergence of SGD assuming (I.1), so that previous statements such as Theorem 4 and Theorem 3 can be applied. At the end of the proof, we shall verify that (I.1) is satisfied throughout the SGD with high probability.

For each $t = 0, 1, \ldots, T-1$, using the same notation as Theorem 1, except that we choose $\nabla_t = \frac{n}{|S_t|} \sum_{i \in S_t} \nabla f_i(W^{(t)})$. We have $\mathbb{E}_{S_t}[\nabla_t] = \nabla f(W^{(t)})$ and therefore

$$\mathbb{E}_{S_t}[f(W^{(t+1)})]$$

$$\stackrel{\text{\textcircled{1}}}{\leq} f(W^{(t)}) - \eta \|\nabla f(W^{(t)})\|_F^2 + O(\rho^{11} \tau_0^{1/3} m^{1/3}) \cdot \|\text{loss}_{i^*, \ell^*}^{(t)}\|_2 \cdot \eta \mathbb{E}_{S_t}[\|\nabla_t\|_2] + O(L^{18} nm\eta^2) \mathbb{E}_{S_t}[\|\nabla_t\|_2^2]$$

$$\stackrel{\text{\textcircled{2}}}{\leq} f(W^{(t)}) - \eta \|\nabla_t\|_F^2 + O\left(\rho^{30} \eta^2 m^2\right) \cdot \|\text{loss}_{i^*, \ell^*}^{(t)}\|_2^2$$

$$\stackrel{\text{\textcircled{3}}}{\leq} \left(1 - \Omega\Big(\frac{\eta\delta}{\rho^{15}} m\Big)\right) f(W^{(t)}) \ . \tag{I.2}$$

Above, ① uses Theorem 4 and $\mathbb{E}_{S_t}[\nabla_t] = \nabla f(W^{(t)})$; ② uses Theorem 3 (which gives $\|\nabla_t\|_2^2 \leq \frac{n^2}{|S_t|^2} \sum_{i \in S_t} \|\nabla f_i(W^{(t)})\|_F^2 \leq O(\rho^{12} m) \times \|\text{loss}_{i^*, \ell^*}^{(t)}\|^2)$; ③ is identical to the proof of Theorem 1.

At the same time, we also have the following absolute value bound:

$$f(W^{(t+1)})$$

$$\overset{①}{\leq} f(W^{(t)}) + \eta \|\nabla f(W^{(t)})\|_F \cdot \|\nabla_t\|_F + O(\rho^{11} \tau_0^{1/3} m^{1/3}) \cdot \|\text{loss}_{i^*, \ell^*}^{(t)}\|_2 \cdot \eta \|\nabla_t\|_2 + O(L^{18} n m \eta^2) \|\nabla_t\|_2^2$$

$$\overset{②}{\leq} f(W^{(t)}) + O\left(\rho^{12} \eta m + \rho^{30} \eta^2 m^2\right) \cdot \|\text{loss}_{i^*, \ell^*}^{(t)}\|_2^2$$

$$\overset{③}{\leq} \left(1 + O(\rho^{12} \eta m)\right) f(W^{(t)}) \; . \tag{I.3}$$

Above, ① uses Theorem 4 and Cauchy-Shwartz $\langle A, B \rangle \leq \|A\|_F \|B\|_F$, and ② uses Theorem 3 and the derivation from (I.2).

Next, taking logarithm on both sides of (I.2) and (I.3), and using Jensen's inequality $\mathbb{E}[\log X] \leq \log \mathbb{E}[X]$, we have

$$\mathbb{E}[\log f(W^{(t+1)})] \leq \log f(W^{(t)}) - \Omega\left(\frac{\eta \delta}{\rho^{15}} m\right) \quad \text{and} \quad \log f(W^{(t+1)}) \leq \log f(W^{(t)}) + O(\rho^{12} \eta m)$$

By one-sided Azuma's inequality (a.k.a. martingale concentration), we have with probability at least $1 - e^{-\Omega(\rho^2)}$, for every $t = 1, 2, \ldots, T$:

$$\log f(W^{(t)}) - \mathbb{E}[\log f(W^{(t)})] \leq \sqrt{t} \cdot O(\rho^{12} \eta m) \cdot \rho \; .$$

This implies two things.

- On one hand, after $T = \Omega(\frac{\rho^{15} \log(nL^2/\varepsilon)}{\eta \delta m})$ iterations we have

$$\log f(W^{(T)}) \leq \sqrt{T} \cdot O(\rho^{12} \eta m) \cdot \rho + \log f(W^{(0)}) - \Omega\left(\frac{\eta \delta}{\rho^{15}} m\right) T$$

$$\leq \log f(W^{(0)}) - \Omega\left(\frac{\eta \delta}{\rho^{15}} m\right) T \leq \log O(n\rho^2 L^3) - \Omega(\log \frac{\rho^5}{\varepsilon}) \leq \log \varepsilon \; .$$

Therefore, we have $f(W^{(T)}) \leq \varepsilon$.

- On the other hand, for every $t = 1, 2, \ldots, T$, we have

$$\log f(W^{(t)}) \leq \sqrt{t} \cdot O(\rho^{12} \eta m) \cdot \rho + \log f(W^{(0)}) - \Omega\left(\frac{\eta \delta}{\rho^{15}} m\right) t$$

$$\overset{①}{=} \log f(W^{(0)}) - \left(\sqrt{\frac{\eta \delta m}{\rho^{15}}} \cdot \Omega(\sqrt{t}) - \sqrt{\frac{\rho^{15}}{\eta \delta m}} \cdot O(\rho^{13} \eta m)\right)^2 + O\left(\frac{\rho^{41} \eta m}{\delta}\right)$$

$$\overset{②}{\leq} \log f(W^{(0)}) + 1$$

where in ① we have used $2a\sqrt{t} - b^2 t = -(b\sqrt{t} - a/b)^2 + a^2/b^2$, and in ② we have used $\eta \leq O\left(\frac{\delta}{\rho^{42} m}\right)$. This implies $f(W^{(t)}) \leq O(n\rho^2 L^3)$. We can now verify for each $t$, $\|W^{(t)} - W^{(0)}\|_F$ is small so that (I.1) holds. By Theorem 3,

$$\|W^{(t)} - W^{(0)}\|_F \leq \sum_{i=0}^{t-1} \|\eta \nabla_t\|_F \leq O(\eta \rho^6 \sqrt{m}) \cdot \sum_{i=0}^{t-1} \sqrt{f(W^{(i)})} \leq O(\eta \rho^6 \sqrt{m}) \cdot O(T\sqrt{n\rho^2 L^3})$$

$$\leq \eta T \cdot O(\rho^{8.5} \sqrt{m}) \leq \frac{\varrho^{50}}{\sqrt{m}} \; .$$

where the last step follows by our choice of $T$. □